\def\TECHREPORTOVERLEAF{1}
\documentclass[10pt, logo, copyright]{nv}

\usepackage{amsmath,amsfonts,bm}

\def\eqref#1{equation~\ref{#1}}

\def\1{\bm{1}}

\DeclareMathAlphabet{\mathsfit}{\encodingdefault}{\sfdefault}{m}{sl}
\SetMathAlphabet{\mathsfit}{bold}{\encodingdefault}{\sfdefault}{bx}{n}

\usepackage[square,sort,comma,numbers]{natbib}
\usepackage{float}
\usepackage{placeins}

\newcommand{\submissionvspace}[1]{}

\makeatletter
\providecommand{\runningtitle}[1]{\gdef\@runningtitle{#1}}
\renewcommand{\abscontent}{%
  \noindent
  \parbox{\dimexpr\linewidth}{\absfont
    \textbf{Abstract:}\nobreakspace\theabstract}%
  \@ifundefined{@keywords}{}{%
    \vskip1em \noindent \keywordsfont Keywords: \@keywords}%
}
\makeatother

\renewcommand{\absfont}{\linespread{1.1}\fontsize{10}{12}\selectfont\looseness=-1\relax}
\titleformat{\subsection}
  {\normalfont\bfseries\fontsize{11}{13}\selectfont}
  {\thesubsection.}{0.5em}{#1}[]
\titleformat{name=\subsection,numberless}
  {\normalfont\bfseries\fontsize{11}{13}\selectfont}
  {}{0em}{#1}[]
\titleformat{\subsubsection}
  {\normalfont\bfseries\itshape\fontsize{10.5}{12.5}\selectfont}
  {\thesubsubsection.}{0.5em}{#1}[]

\renewcommand{\titlefont}{\color{nvidiagreen}\normalfont\bfseries\fontsize{16}{19}\selectfont}

\makeatletter
\ifdefined\TECHREPORTOVERLEAF
  \def\input@path{{./}}
\else
  \def\input@path{{./}}
\fi
\makeatother

\ifdefined\TECHREPORTOVERLEAF
  \graphicspath{{figures/}}
\else
  \graphicspath{{figures/}}
\fi

\hypersetup{
  colorlinks=true,
  linkcolor=nvidiagreen,
  citecolor=nvidiagreen,
  urlcolor=nvidiagreen,
  pdfauthor={Yunze Liu, Jing Wang, Shenrui Wu, Wenkun He, Guanqi Zhan, Xianzheng Ma, Junyu Chen, Wenhang Ge, Haozhan Tang, Yuchao Gu, Zhuoyang Zhang, Zekai Wang, Dantong Niu, Zhen Dong, Yuke Zhu, Trevor Darrell, Linxi “Jim” Fan, Song Han, Han Cai},
  pdftitle={Robota-Embed: Unified VLA Embeddings for Robot In-Context Learning and Evolution},
  pdfsubject={NVIDIA technical report}
}

\def\EXPANDEDREPORT{1}
\def\TECHREPORTPROMOTERETRIEVALEXAMPLE{1}

\newcommand{\physisname}[1]{\texorpdfstring{{\fontshapeforce{it}\selectfont #1}}{#1}}
\newcommand{\methodname}{\physisname{Embed2ICL}}
\newcommand{\embeddingmodelplain}{Robota-Embed}
\newcommand{\embeddingmodelname}{\physisname{\embeddingmodelplain}}
\newcommand{\benchmarkname}{\physisname{Robota-Embed-Bench}}

\newcommand{\detailref}[1]{Section~\ref{#1}}

\definecolor{PhysisTeal}{HTML}{187F79}
\definecolor{PhysisInk}{HTML}{182B35}
\newcommand{\tablebest}[1]{\textcolor{PhysisTeal}{\textbf{#1}}}
\newcommand{\tablegroup}[1]{\textcolor{PhysisInk}{\textbf{#1}}}

\newif\ifhasreferences
\hasreferencestrue

\newcommand{\reportprojecturl}{https://robota-ai.github.io/robota-embed/}

\title{Robota-Embed: Unified VLA Embeddings for Robot\\
In-Context Learning and Evolution}
\runningtitle{Robota-Embed: Unified VLA Embeddings for Robot In-Context Learning and Evolution}
\newcommand{\reportaffilmark}[1]{\raisebox{0.55ex}{\normalfont\scriptsize #1}}
\newcommand{\reportauthorline}[1]{{\normalfont\bfseries\upshape\fontsize{9pt}{12.5pt}\selectfont\makebox[\linewidth][l]{#1}}}
\newcommand{\reportauthorsep}{,\hspace{0.7em}}
\author{%
\vspace{-1.5em}
\makebox[\linewidth][l]{%
\begin{minipage}[t]{\dimexpr\linewidth+24pt\relax}
\raggedright
\reportauthorline{\mbox{Yunze Liu\reportaffilmark{1,*}}\reportauthorsep
\mbox{Jing Wang\reportaffilmark{1,*}}\reportauthorsep
\mbox{Shenrui Wu\reportaffilmark{3}}\reportauthorsep
\mbox{Wenkun He\reportaffilmark{2}}\reportauthorsep
\mbox{Guanqi Zhan\reportaffilmark{1}},}\\
\reportauthorline{\mbox{Xianzheng Ma\reportaffilmark{4}}\reportauthorsep
\mbox{Junyu Chen\reportaffilmark{1}}\reportauthorsep
\mbox{Wenhang Ge\reportaffilmark{1}}\reportauthorsep
\mbox{Haozhan Tang\reportaffilmark{5}}\reportauthorsep
\mbox{Yuchao Gu\reportaffilmark{1}},}\\
\reportauthorline{\mbox{Zhuoyang Zhang\reportaffilmark{2}}\reportauthorsep
\mbox{Zekai Wang\reportaffilmark{3}}\reportauthorsep
\mbox{Dantong Niu\reportaffilmark{3}}\reportauthorsep
\mbox{Zhen Dong\reportaffilmark{1}}\reportauthorsep
\mbox{Yuke Zhu\reportaffilmark{1}},}\\
\reportauthorline{\mbox{Trevor Darrell\reportaffilmark{3}}\reportauthorsep
\mbox{Linxi ``Jim'' Fan\reportaffilmark{1}}\reportauthorsep
\mbox{Song Han\reportaffilmark{2}}\reportauthorsep
\mbox{Han Cai\reportaffilmark{1,\ensuremath{\dagger}}}}
\par\vspace{2.5mm}\noindent
{\normalfont\color{black!75}\fontsize{8pt}{10.4pt}\selectfont
\reportaffilmark{1}\,NVIDIA\quad
\reportaffilmark{2}\,MIT\quad
\reportaffilmark{3}\,UCB\quad
\reportaffilmark{4}\,Oxford\quad
\reportaffilmark{5}\,CMU\\
\reportaffilmark{*}\,Equal contribution.\quad
\reportaffilmark{\ensuremath{\dagger}}\,Corresponding author.
\par\vspace{1.2mm}\noindent
\href{\reportprojecturl}{\color{nvidiagreen}\bfseries Project Page}}
\end{minipage}
}%
\vspace{-13pt}
}

\begin{abstract}
Retrieving useful context for a robot requires matching not only its task,
but also its current stage of execution and the continuation that should
follow. We introduce \embeddingmodelname{}, to our knowledge the first
vision--language--action (VLA) embedding model to jointly encode all three
modalities in both current-state queries and recorded future continuations.
Predictive retrieval training aligns these endpoints for physical-experience
retrieval, and the resulting representation retrieves historical
continuations as context at inference time. We construct a multi-source corpus
of 1,695,684 cleaned training anchors and introduce \benchmarkname{}, which
evaluates 4,000 queries against a shared 8,000-entry gallery. Our model achieves
47.33\% clip-level and 81.77\% episode-level R@1, surpassing action-free Gemini
Embedding 2 by 22.09 and 33.47 percentage points. Across controlled retrieval
studies, removing native action degrades retrieval more than removing vision,
indicating that execution information is a central part of the retrieval key.
We then study how this shared retrieval space transfers beyond retrieval
metrics through two complementary applications. \emph{Embed2ICL} uses
retrieved experience for training-free and training-based policy conditioning,
increasing success from 47.0\% to 56.3\% in DROID-Sim and from 86.4\% to 92.8\%
on the LIBERO difficult subset. Both interfaces also improve task progress in
real-robot evaluations. \emph{Embed2Evolve} uses execution feedback to grow
task-local memory and refine how retrieved context is used around a frozen
GPT-6 Astra policy. Across five ManiSkill tasks, it achieves 68\% success,
compared with 24\% for GPT-6 Astra Direct and 33\% for a harness independently
evolved with Omni-Embed-Nemotron-3B under the same protocol. Relative to the
latter, it uses approximately 79.5\% fewer policy tokens per test rollout.
Together, these results show that a unified, action-aware retrieval space can
support both robot in-context learning and task-local context evolution.
\end{abstract}

\begin{document}

\maketitle
\vspace{8pt}
\begin{figure}[!ht]
  \vspace{-10pt}
  \centering
  \includegraphics[width=\textwidth]{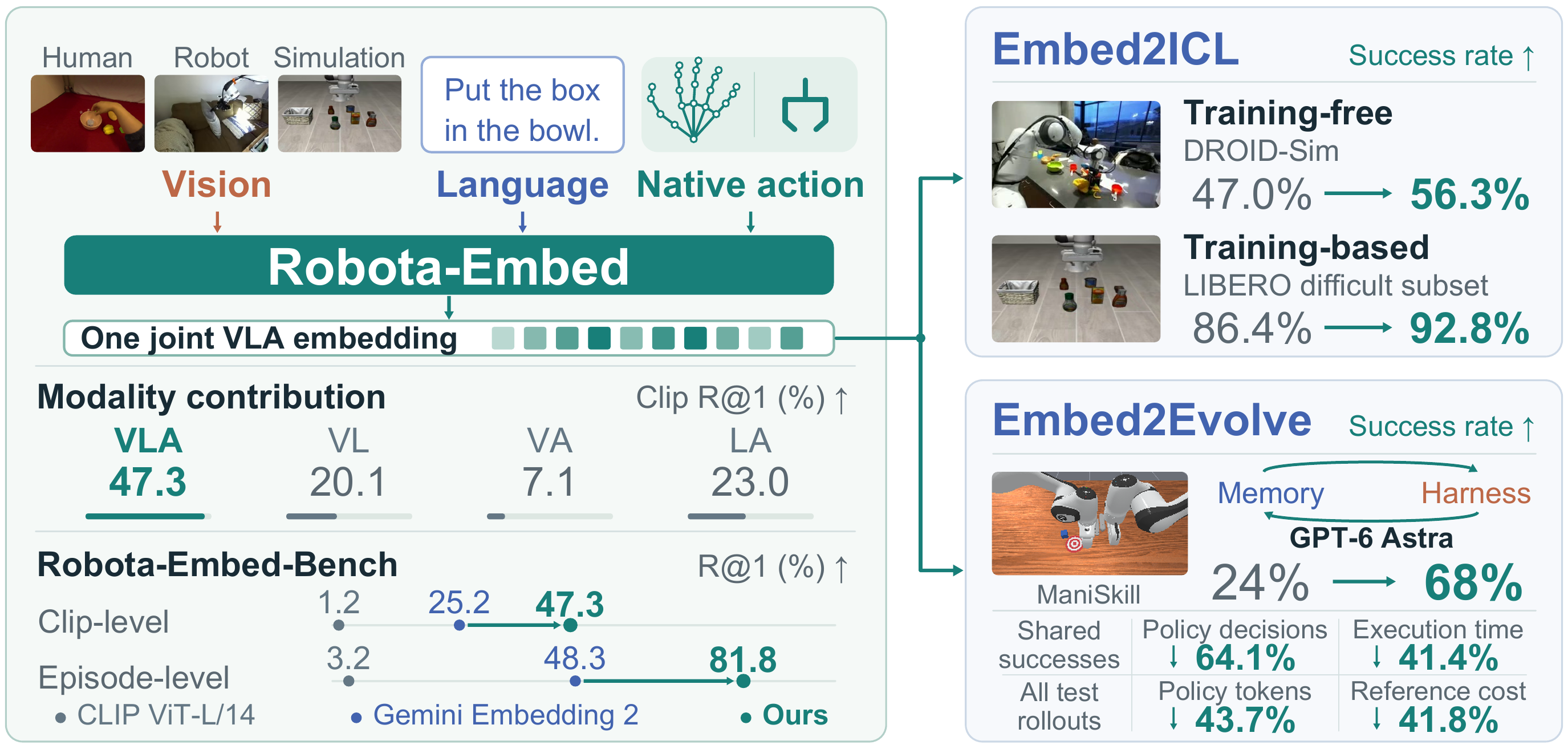}
  \par\vspace{-12pt}
  \ifdefined\EXPANDEDREPORT\captionsetup{skip=17pt}\fi
  \ifdefined\EXPANDEDREPORT
  \caption{\textbf{One joint VLA embedding, two context applications.}
  \embeddingmodelplain{} jointly encodes vision, language, and native action from
  human, robot, and simulated experience to retrieve recorded continuations.
  \textbf{Left:} retrieval accuracy and modality contributions on \benchmarkname{}.
  \textbf{Right:} retrieved demonstrations condition robot policies in Embed2ICL;
  task-local memory and context use evolve around a fixed GPT-6 Astra policy
  in Embed2Evolve. Token savings include failed final tests and exclude
  development; decision/time savings use matched successful pairs.
  These efficiency comparisons use Direct as the baseline.}
  \else
  \caption{\textbf{One joint VLA embedding, two context applications.}
  \embeddingmodelplain{} jointly encodes vision, language, and native action from human,
  robot, and simulated experience to retrieve recorded continuations.
  Action improves retrieval beyond vision and language alone.
  Embed2ICL uses retrieved demonstrations to condition robot policies.
  Embed2Evolve grows task-local memory and refines how a fixed GPT-6 Astra
  policy uses it, improving success and efficiency relative to Direct.}
  \fi
  \label{fig:teaser}
  \vspace{-13pt}
\end{figure}
\FloatBarrier
\clearpage

\section{Introduction}
\label{sec:introduction}

Robot policies increasingly use experience supplied in context.
GEN-1.5~\citep{generalist2026gen15} and S1~\citep{skild2026s1} demonstrate
Robot In-Context Learning from sensorimotor or video examples. In our experiments,
GPT-6 Astra~\citep{openaigpt6astra} generates robot actions from observations and execution
history. These capabilities make experience selection part of the control problem.
Demonstration-conditioned world action models (WAMs) need useful existing
examples; policies acting repeatedly must organize accumulated experience.
Both raise a representation question:
\emph{how should physical experience be represented for task- and stage-sensitive retrieval?}

Task similarity alone is insufficient. Reaching toward a bowl and withdrawing
after placement can share objects and an instruction yet require different
continuations. Language specifies the goal, vision the
interaction state, and native action the ongoing motion. The retrieval endpoints
have different temporal roles: the query describes the present, whereas a
candidate records what followed an earlier interaction. \emph{Predictive retrieval}
therefore requires sensitivity to task intent and execution stage: the goal is
an appropriate recorded continuation, not simply a similar scene.

General-purpose embeddings align vision and
language~\citep{radford2021learning,girdhar2023imagebind,xu2025omni}, while
action-aware metric learning connects perception or language with
behavior~\citep{sung2017deep,rana2023contrastive}.
We introduce \textbf{\embeddingmodelname{}}, a unified vision--language--action
(VLA) embedding model for physical-experience retrieval. To our knowledge, it is
the first to jointly encode all three modalities in both current-state queries
and recorded continuations. Each endpoint yields one vector, so native action
informs retrieval itself. Source-specific projectors handle
heterogeneous action formats.
\ifdefined\EXPANDEDREPORT
Retrieval training combines predictive alignment, multimodal grounding and
discrimination, and representation regularization.
\fi

\ifdefined\EXPANDEDREPORT
We construct a multi-source training corpus and introduce \benchmarkname{},
with 4,000 queries and 8,000 gallery candidates spanning human, simulated,
and real-robot experience. Nearby same-episode clips distinguish finding the
right experience from locating its recorded continuation. Across four sources,
\embeddingmodelname{} achieves 47.33\% clip-level and 81.77\% episode-level
R@1. Native action adds information beyond text and images: with backbone,
data, and training budget matched on DROID, adding action raises clip-level
R@1 from 59.8\% to 78.7\%. Complementary retraining and query-masking studies
examine which information the representation learns and uses.
\else
We construct multi-source training data and introduce \benchmarkname{}
for human, simulated, and real-robot experience. Nearby same-episode clips test
whether retrieval locates the recorded continuation or merely its episode.
Across the four sources, \embeddingmodelname{} achieves 47.33\% clip-level
and 81.77\% episode-level R@1.
Native action contributes beyond text and images: with backbone, data,
and training budget matched on DROID,
adding action raises clip-level R@1 from 59.8\% to 78.7\%.
Removing action reduces retrieval accuracy more than removing vision in both
matched retraining and inference-time query masking.
\fi

\ifdefined\EXPANDEDREPORT
Figure~\ref{fig:teaser} connects retrieval to demonstration conditioning and
memory-based interaction. \textbf{Embed2ICL} supports robot in-context learning
through two policy interfaces. For
DreamZero~\citep{ye2026dreamzero}, retrieved videos initialize the visual
cache without policy updates. For Fast-WAM~\citep{yuan2026fastwam}, we
fine-tune the policy to condition on precomputed Context Tokens from retrieved
demonstrations. Both interfaces improve simulation task success and real-robot
task progress. Random-reference controls show that useful context depends on
reference selection.

\textbf{Embed2Evolve} extends retrieval to context management for agentic
robots. Repeated attempts grow task-local memory, while execution feedback
refines retrieval, refresh, and retained-history settings around a frozen
GPT-6 Astra policy. Policy and embedding weights remain fixed. Across five
ManiSkill tasks, the selected context harness raises success from 24\% with
Direct to 68\%, using 43.7\% fewer policy tokens per test rollout. To examine
the retrieval model's role, we independently repeat evolution with
Omni-Embed-Nemotron-3B from the same initial memory under the same search
protocol. Its harness reaches 33\% success; Embed2Evolve uses approximately
79.5\% fewer tokens than this baseline. Token averages include failed tests
and exclude development.
\else
As shown in Figure~\ref{fig:teaser}, we use \embeddingmodelname{} in two applications:
learning from existing demonstrations and evolving context through repeated interaction.
\textbf{Embed2ICL} provides \emph{embedding-guided robot in-context learning}.
For DreamZero~\citep{ye2026dreamzero},
we initialize the visual cache with reference videos without policy updates;
for Fast-WAM~\citep{yuan2026fastwam}, we fine-tune the policy to condition on
retrieved Context Tokens. Both interfaces improve simulation task success and
real-robot task progress. Random references yield lower simulation success.

\textbf{Embed2Evolve} performs \emph{embedding-guided context evolution}:
repeated attempts grow task-local memory, and execution feedback refines how
a frozen GPT-6 Astra policy accesses it.
Across five ManiSkill tasks, the selected \emph{context harness} raises
success from 24\% with GPT-6 Astra Direct to 68\%. To test the retrieval model's role, we replace it with
Omni-Embed-Nemotron-3B and evolve both conditions independently
from the same initial memory under the same search protocol. Compared with
this baseline, which reaches 33\% success, Embed2Evolve uses
approximately 79.5\% fewer policy tokens per test rollout, including failures.
The retrieval representation therefore matters for both memory selection and
the final harness's success and token use.
\fi

We make four contributions.
\ifdefined\EXPANDEDREPORT
\begin{itemize}[leftmargin=1.4em,itemsep=4pt,topsep=4pt,parsep=0pt]
  \item \textbf{\embeddingmodelplain{}.} A unified embedding model that jointly
  encodes vision, language, and native action at both endpoints for predictive
  physical-experience retrieval.

  \item \textbf{Data and evaluation.} A multi-source training corpus,
  \benchmarkname{}, and controlled studies of temporal precision, representation
  learning, and native action's contribution beyond vision and language.

  \item \textbf{Embed2ICL.} Training-free and training-based interfaces that
  turn retrieved demonstrations into useful policy context, evaluated in
  simulation and on real robots.

  \item \textbf{Embed2Evolve.} Embedding-guided memory and context evolution
  around a frozen policy, improving success and token efficiency under matched
  comparisons of retrieval embedding models.
\end{itemize}
\clearpage
\else
\textbf{(1) \embeddingmodelplain{}}, a unified VLA embedding model for predictive
physical-experience retrieval.
\textbf{(2) Data and evaluation}: a multi-source training corpus,
\benchmarkname{}, and controlled studies of the representation and its inputs.
\textbf{(3) Embed2ICL}, connecting retrieved experience to training-free and
training-based Robot In-Context Learning in simulation and on real robots.
\textbf{(4) Embed2Evolve}, developing task-local memory and its use
around a frozen policy, with a matched comparison of retrieval embedding models.
\fi
\section{Related Work}
\label{sec:related-work}

\paragraph{Multimodal and action-aware embeddings.}
\label{sec:related-unified-embeddings}

CLIP aligns images with language~\citep{radford2021learning};
Perception Encoder and InternVideo2 extend vision--language representation
learning to video~\citep{bolya2025perceptionencoder,wang2024internvideo2}.
Unified vision--language embeddings support image and video retrieval in one
space~\citep{meng2025vlm2vecv2,xu2025omni,zhou2026wemmembedding,
shanbhogue2026geminiembedding2,li2026qwen3vlembedding};
\mbox{ImageBind}~\citep{girdhar2023imagebind} includes additional sensory modalities.
Deep Multimodal Embedding~\citep{sung2017deep} and
CLASP~\citep{rana2023contrastive} align language or perception with action.
VLA-R~\citep{seong2025vlar} embeds visual--language queries and action-token
candidates in a shared space, while RADAR~\citep{gao2026radar} reuses VLA-policy features for
retrieval. \embeddingmodelname{} jointly encodes vision, language, and native
action at both retrieval endpoints: the current state and the recorded future clip.

\paragraph{Physical-experience retrieval.}
\label{sec:related-robot-rag}

Action can enter the retrieval key or the content returned to the policy.
ROSER~\citep{rahman2026roser} searches proprioceptive sequences;
RAEA~\citep{zhu2024retrievalaugmented}, REGENT~\citep{sridhar2025regent}, and
RICL~\citep{sridhar2025ricl} return actions selected through action-free keys.
Zeva~\citep{chen2026zeva} indexes action--effect evidence, whereas
RADAR~\citep{gao2026radar} uses policy features.
MemER~\citep{sridhar2026memer} selects rollout keyframes;
Retrieve-Reason-Act~\citep{temiraliev2026retrievalaugmented} indexes visual manuals.
\embeddingmodelname{} includes native action in both query and candidate keys,
so motion informs the similarity comparison itself
(Table~\ref{tab:related-protocols}).

\paragraph{Robot In-Context Learning and Evolution.}
\label{sec:related-physical-icl}

GEN-1.5~\citep{generalist2026gen15} conditions on a sensorimotor example;
S1~\citep{skild2026s1}, Zero-WAM~\citep{zhou2026zerowam}, and
HOST~\citep{chen2026host} use supplied video. R+X~\citep{papagiannis2025rx}
retrieves human-video clips and converts them into keypoint and hand-trajectory
context. Embed2ICL uses \embeddingmodelname{} to select experience for
training-free and training-based WAM conditioning.
Reflexion~\citep{shinn2023reflexion} stores verbal reflections in episodic
memory, using feedback to improve later trials without updating model weights.
In robotics, Zeva~\citep{chen2026zeva} retrieves action--effect evidence across
retries within an episode, resetting memory for each new episode.
Embed2Evolve performs \emph{in-context evolution}:
it grows task-local memory across development episodes and refines retrieval,
refresh, and history settings around a frozen policy. Evaluation uses the selected
memory and context settings without further changes on fresh task initializations.
\section{\embeddingmodelname{}}
\label{sec:method}

\subsection{Problem Formulation}
\label{sec:method-overview}

\emph{Physical AI experience} combines language, vision, and source-native
motion or control from human, simulated, or robotic interaction. Given a current
VLA state, \embeddingmodelname{} retrieves the video and actions that followed
an anchor in a recorded trajectory.

For sample $i$ from source $d_i$, the query combines instruction $\ell_i$,
an interaction-region image $z_{i,t}$,
and synchronized native action $a_{i,t}$. The paired target combines the same
instruction with a four-second video clip
$V_i^+$ and aligned actions $A_i^+$ recorded strictly after $t$:
\begin{equation}
  x_i^q=(\ell_i,z_{i,t},a_{i,t},d_i),\qquad
  x_i^k=(\ell_i,V_i^+,A_i^+,d_i).
  \label{eq:paired-vla-views}
\end{equation}
\subsection{Action-Native Predictive VLA Embeddings}
\label{sec:method-embedding}

\begin{figure}[!ht]
  \submissionvspace{-9pt}
  \centering
  \ifdefined\EXPANDEDREPORT
    \includegraphics[width=\linewidth,trim=0 0 0 11bp,clip]{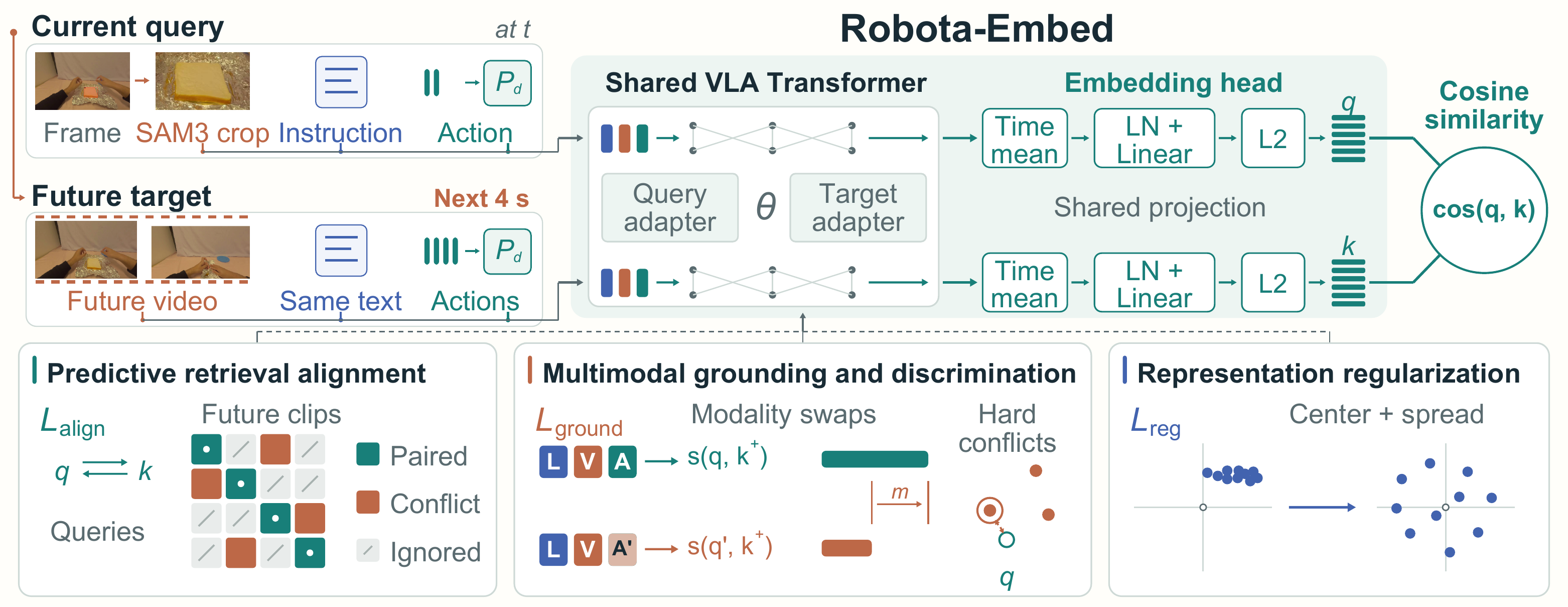}
  \else
    \includegraphics[width=\linewidth]{figures/pai-embedding-objective.pdf}
  \fi
  \par\vspace{-12pt}
  \ifdefined\EXPANDEDREPORT\captionsetup{skip=17pt}\fi
  \caption{\textbf{\embeddingmodelname{} architecture and objectives.}
  Each query is paired with the next four seconds of recorded video and actions.
  Three objective groups train a joint VLA space.}
  \label{fig:pai-embedding-objective}
  \submissionvspace{-8pt}
\end{figure}

We train \embeddingmodelname{} for current-to-future retrieval using
Cosmos3-Edge 4B~\citep{nvidia2026cosmos3} as the base model
(Figure~\ref{fig:pai-embedding-objective}). Query and target share
the Transformer, video tokenizer, and head, with separate adapters $A_q,A_k$.
Source-specific projectors $P_{d_i}$ map native actions
into the backbone. Each endpoint jointly encodes language, vision, and action
into one normalized vector.
For retrieval, we rank candidates by the cosine similarity between the
current-state query $q_i=f_\theta^q(x_i^q;P_{d_i})$ and the future-clip target
$k_i=f_\theta^k(x_i^k;P_{d_i})$.

\paragraph{Retrieval-oriented training.}
We train with predictive alignment, multimodal grounding, and representation
regularization. Structured parsing and deterministic rules supply weak labels.
\ifdefined\EXPANDEDREPORT
The supervision distinguishes exact continuation identity from broader task
compatibility; a different clip ID alone does not establish a valid negative.
Native action is learned within the fused representation, with modality swaps
encouraging sensitivity to motion without requiring action-only matching to
an exact future clip.
\fi

\begin{samepage}
Predictive retrieval alignment ($\mathcal{L}_{\mathrm{align}}$) uses bidirectional
InfoNCE with exact future-clip positives and task, object, action, or joint
conflicts as negatives. Compatible, uncertain, or potentially duplicate
non-positives are ignored. Figure~\ref{fig:pai-embedding-objective} shows the
validity mask. We optimize:
\begin{equation}
  \mathcal{L}=\mathcal{L}_{\mathrm{align}}
  +\mathcal{L}_{\mathrm{ground}}+\mathcal{L}_{\mathrm{reg}}.
  \label{eq:composite-objective}
\end{equation}
Multimodal grounding and discrimination
($\mathcal{L}_{\mathrm{ground}}$) combines task/object auxiliaries with margins
against modality swaps and hard semantic conflicts, so that similarity responds
to changes in the inputs. Representation regularization
($\mathcal{L}_{\mathrm{reg}}$) centers query and target batches and discourages
low-variance, low-rank embeddings.
\detailref{sec:rerum-implementation}--\ref{sec:pai-objective-details} detail the
architecture, native-action preprocessing, optimization schedule, and individual
training losses.
\par
\end{samepage}
\subsection{Training Data and Retrieval Benchmark}
\label{sec:method-training-corpus}

Our training corpus contains 1,695,684 filtered current--future pairs from
EgoDex~\citep{hoque2026egodex}, DROID~\citep{khazatsky2024droid},
LIBERO~\citep{liu2023libero}, and
RoboMIND--Franka-dual~\citep{wu2025robomind}. EgoDex records tracked human
motion; the other sources record robot controls. We sample sources equally and form
within-source pairs (Figure~\ref{fig:roborag-2m-curation};
\detailref{sec:training-data-curation}).

\begin{figure}[!ht]
  \submissionvspace{-5pt}
  \centering
  \ifdefined\EXPANDEDREPORT
    \includegraphics[width=\linewidth,trim=0 40bp 0 20bp,clip]{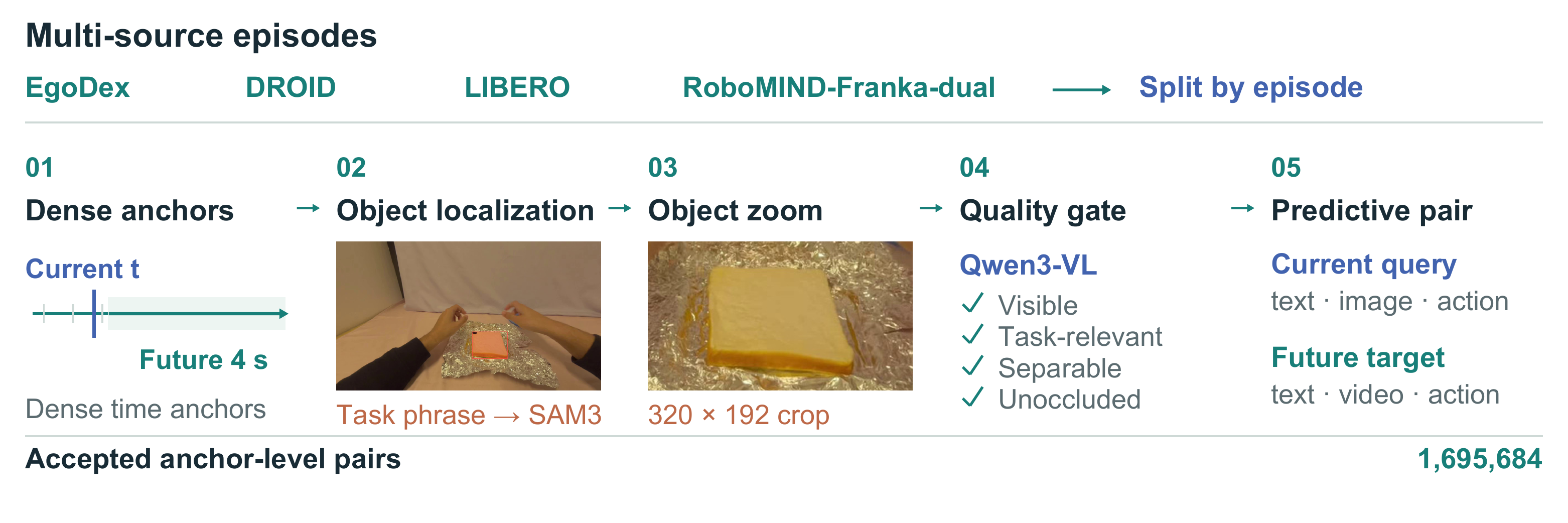}
  \else
    \includegraphics[width=\linewidth,trim=0 40bp 0 0,clip]{figures/roborag-2m-pipeline.pdf}
  \fi
  \par\vspace{-7pt}
  \ifdefined\EXPANDEDREPORT\captionsetup{skip=13pt}\fi
  \caption{\textbf{Multi-source training-data curation.} We split episodes before
  sampling anchors, localize and verify interaction crops, and pair accepted
  anchors with video and actions from the next four seconds. Query filtering
  uses only the instruction and current observations.}
  \label{fig:roborag-2m-curation}
  \submissionvspace{-10pt}
\end{figure}
\paragraph{\benchmarkname{}.}
\label{sec:method-benchmark}

All 4,000 held-out queries search one shared 8,000-clip gallery from the four
source datasets. Each source contributes 1,000 queries from 800 episodes,
1,000 paired future clips, and 1,000 nearby same-episode clips.
Queries follow Equation~\ref{eq:paired-vla-views}.
\emph{Clip-level retrieval} treats only the paired clip as positive;
\emph{episode-level retrieval} accepts all clips from the same source episode.
Downstream experiments test whether retrieved context helps control.

We split episodes, sessions, and content identities before sampling anchors;
task, object, and embodiment categories may recur. Clip-level metrics weight
sources equally. Episode-level metrics aggregate within episodes, then equally
over each source's 800 episodes and the four sources. We use cosine similarity without reranking
(\detailref{sec:benchmark-protocol-details}).
\FloatBarrier
\section{Embedding Evaluation}
\label{sec:experiments}

\begin{table*}[!ht]
  \submissionvspace{-7pt}
  \centering
  \caption{\textbf{Retrieval on \benchmarkname{}.} Four-source average recall
  and MRR (\%) on the shared 8,000-clip gallery. Bold marks column-wise best
  results.}
  \label{tab:roborag-bench-main}
  \small
  \setlength{\tabcolsep}{5.5pt}
  \renewcommand{\arraystretch}{0.92}
  \begin{tabular*}{\linewidth}{@{\extracolsep{\fill}}lrrrr@{\hspace{9pt}}rrrr@{}}
  \toprule
  & \multicolumn{4}{c}{\tablegroup{Clip-level}} &
    \multicolumn{4}{c}{\tablegroup{Episode-level}} \\
  \cmidrule(lr){2-5}\cmidrule(l){6-9}
  \tablegroup{Model} & R@1 & R@5 & R@10 & MRR & R@1 & R@5 & R@10 & MRR \\
  \midrule
  \multicolumn{9}{@{}l}{\tablegroup{CLIP-style text--video embeddings}} \\
  \cmidrule{1-9}
  CLIP ViT-L/14
    & 1.2 & 5.8 & 9.8 & 4.06 & 3.2 & 7.7 & 11.6 & 5.98 \\
  Perception Encoder B16
    & 2.0 & 8.1 & 12.8 & 5.52 & 5.0 & 10.1 & 15.2 & 8.01 \\
  InternVideo2
    & 1.6 & 7.3 & 11.4 & 4.82 & 4.1 & 9.1 & 13.2 & 6.94 \\
  \midrule
  \multicolumn{9}{@{}l}{\tablegroup{Unified VL embeddings}} \\
  \cmidrule{1-9}
  Qwen3-VL-Embedding-2B
    & 23.3 & 51.6 & 57.3 & 36.20 & 45.1 & 55.9 & 61.0 & 50.71 \\
  VLM2Vec-V2
    & 24.1 & 51.9 & 57.6 & 36.67 & 47.3 & 58.1 & 63.3 & 52.88 \\
  WeMM-Embedding-2B
    & 24.1 & 52.2 & 58.2 & 36.78 & 45.1 & 57.3 & 63.2 & 51.08 \\
  Omni-Embed-Nemotron-3B
    & 24.5 & 54.5 & 60.1 & 38.10 & 48.6 & 59.1 & 64.4 & 53.84 \\
  Gemini Embedding 2
    & 25.2 & 52.3 & 57.5 & 37.68 & 48.3 & 58.2 & 64.7 & 53.69 \\
  \midrule
  \multicolumn{9}{@{}l}{\tablegroup{Unified VLA embedding}} \\
  \cmidrule{1-9}
  \tablegroup{\embeddingmodelname{} (ours)}
    & \tablebest{47.3} & \tablebest{91.9} & \tablebest{97.0}
    & \tablebest{67.05} & \tablebest{81.8} & \tablebest{94.9}
    & \tablebest{98.1} & \tablebest{86.96} \\
  \bottomrule
\end{tabular*}

  \submissionvspace{-10pt}
\end{table*}
\subsection{Unified VLA Retrieval on \benchmarkname{}}
\label{sec:exp-retrieval}

\paragraph{Protocol and models.}
Eight action-free baselines search the shared gallery
(Section~\ref{sec:method-benchmark}). CLIP-style models retrieve video from
text; unified VL models also use visual queries
(Table~\ref{tab:roborag-bench-main}). We use each model's supported interface.
The retraining studies in Section~\ref{sec:analysis-ablation} use the same DROID
data splits and training budgets for the compared model variants.

\begin{samepage}
\paragraph{Main results.}
\embeddingmodelname{} achieves 47.33\% clip-level and 81.77\% episode-level
R@1. The strongest baselines, Gemini Embedding 2 and Omni-Embed-Nemotron-3B,
reach 25.24\% and 48.59\%, respectively. \embeddingmodelname{} improves on
them by 22.09 and 33.18 percentage points.
\par
\end{samepage}

Gains over unified VL baselines are largest on LIBERO and RoboMIND-FD
(\detailref{sec:benchmark-per-domain-results}).
Episode retrieval does not resolve temporal localization: on EgoDex,
episode/clip R@1 is 87.06/36.70\%, versus 99.88/74.10\% on DROID.
Finding a relevant experience is easier than locating its paired continuation,
the target of predictive retrieval.

\paragraph{Benchmark validity.}
Masking candidate text before fusion and recomputing embeddings
lowers clip-level R@1 from 47.33\% to 12.78\%, confirming that the model uses
language. Restricting candidates to the same instruction still gives 47.62\%
R@1, so the model also distinguishes continuations that share an instruction.
Within same-instruction galleries, \embeddingmodelname{} achieves 38.60\%
and 40.10\% R@1 on LIBERO and RoboMIND-FD, compared with random-ranking
rates of 1.90\% and 0.24\%, respectively. High-$k$ recall approaches
saturation in the smaller EgoDex and DROID pools.
\ifdefined\EXPANDEDREPORT
Tables~\ref{tab:benchmark-validity-summary} and~\ref{tab:same-instruction-difficulty}
give the full aggregate results, per-source candidate counts, and random baselines.
\else
\detailref{sec:benchmark-validity-audit} gives per-source results, candidate
counts, and random baselines.
\fi

\paragraph{Index-size scaling.}
Expanding the precomputed gallery from 1M to 15M entries increases query
encoding plus retrieval latency from 160.36 to 195.17 ms on one NVIDIA A100
(Table~\ref{tab:gallery-size-scaling}). At 15M, Top-10 search alone takes
37.47 ms. These timings exclude time spent on SAM3 preprocessing, reference
reranking, and downstream policy execution (\detailref{sec:gallery-size-scaling}).

\subsection{What Makes Predictive Retrieval Work?}
\label{sec:report-controlled-analysis}

The controlled studies separate representation adaptation from input use.
On the shared four-source benchmark, a frozen Cosmos3-Edge readout reaches
only 0.65\% clip-level R@1, whereas retrieval training reaches 47.33\%
(Table~\ref{tab:pai-frozen-readout}). On the matched DROID split, a learned
head over frozen features reaches 13.8\%, compared with 78.7\% when the
backbone and head are adapted (Table~\ref{tab:droid-trainable-scope}).
Predictive alignment already places the paired continuation near the top at
75.6\% R@1 and 100.0\% R@10; representation regularization raises R@1 to
77.0\%, and multimodal grounding and discrimination further raise it to
78.7\% (Table~\ref{tab:droid-objective-groups}). The added objectives therefore
refine top-rank ordering rather than merely broad recall.

Action is the most consequential non-language input in both controlled views.
Retraining without action lowers R@1 by 18.9 percentage points, versus 1.9
points without vision (Table~\ref{tab:droid-training-modalities}); with the
final model and gallery fixed, masking query action or vision lowers clip R@1
by 27.25 or 24.33 points
(Table~\ref{tab:roborag-bench-modality-availability}). These interventions
answer different questions: retraining measures what the remaining modalities
can relearn, whereas masking measures reliance under an input shift. Thus the
small no-vision retraining gap does not imply that the final model ignores
vision. SAM3 interaction crops provide a further, modest localization gain
over full frames, 78.7\% versus 77.9\% R@1
(Table~\ref{tab:droid-visual-input}). Appendix~\ref{app:roborag-bench} reports
each intervention as a separate table and conclusion.

\subsection{Benchmark Retrieval Examples}
\label{sec:benchmark-retrieval-examples}

Figure~\ref{fig:pai-retrieval-diagnostics} contrasts a paired-clip hit with a
temporal near-miss. The latter retrieves a nearby window from the correct
episode rather than the paired continuation, illustrating why episode matching
and clip localization require separate metrics.

\begin{figure}[!ht]
  \submissionvspace{-3pt}
  \centering
  \includegraphics[width=\linewidth]{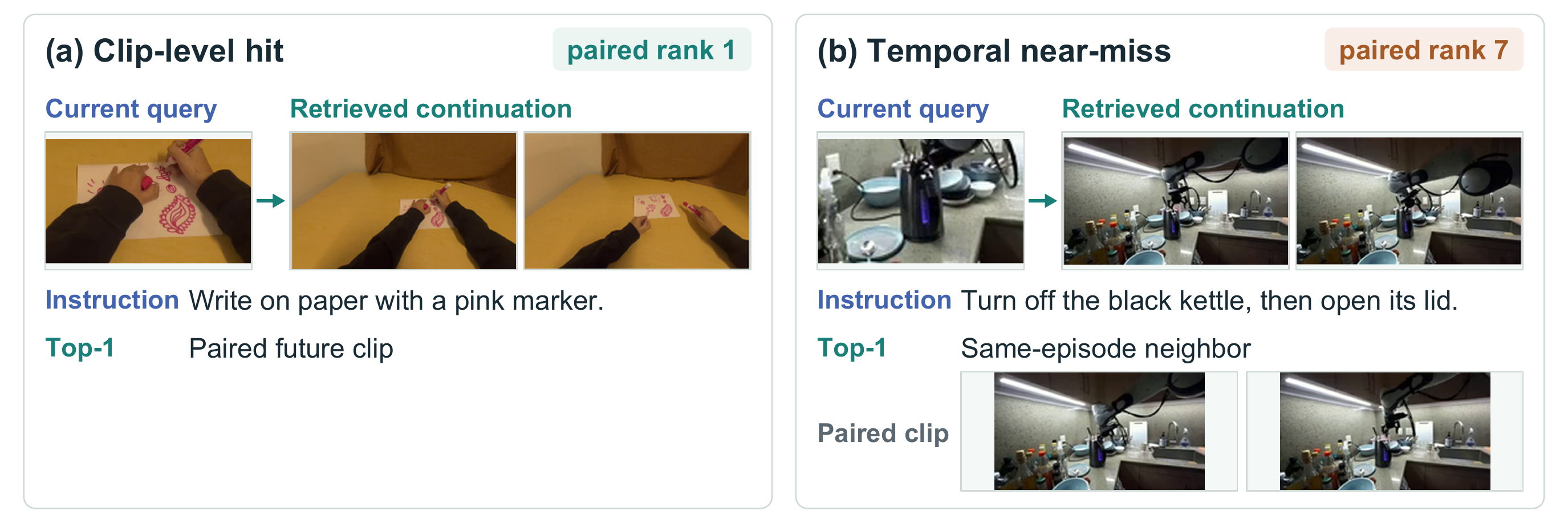}
  \ifdefined\EXPANDEDREPORT
    \vspace{-18pt}
  \else
    \vspace{-4pt}
  \fi
  \caption{\textbf{Selected retrieval diagnostics.}
  (a) The paired future clip ranks first. (b) A nearby clip from the same episode
  ranks first, while the paired clip ranks seventh. Both top-1 returns are
  episode-level matches; only (a) is also a clip-level match.}
  \label{fig:pai-retrieval-diagnostics}
  \submissionvspace{-4pt}
\end{figure}
\FloatBarrier
\section{From Embeddings to Robot Context}
\label{sec:applications}

\subsection{Embed2ICL: Embedding-Guided Robot In-Context Learning}
\label{sec:exp-physical-icl}
\label{sec:method-downstream-icl}
\label{sec:method-zero-shot}
\label{sec:method-fastwam}

\begin{figure}[!ht]
  \submissionvspace{-8pt}
  \centering
  \includegraphics[width=\linewidth]{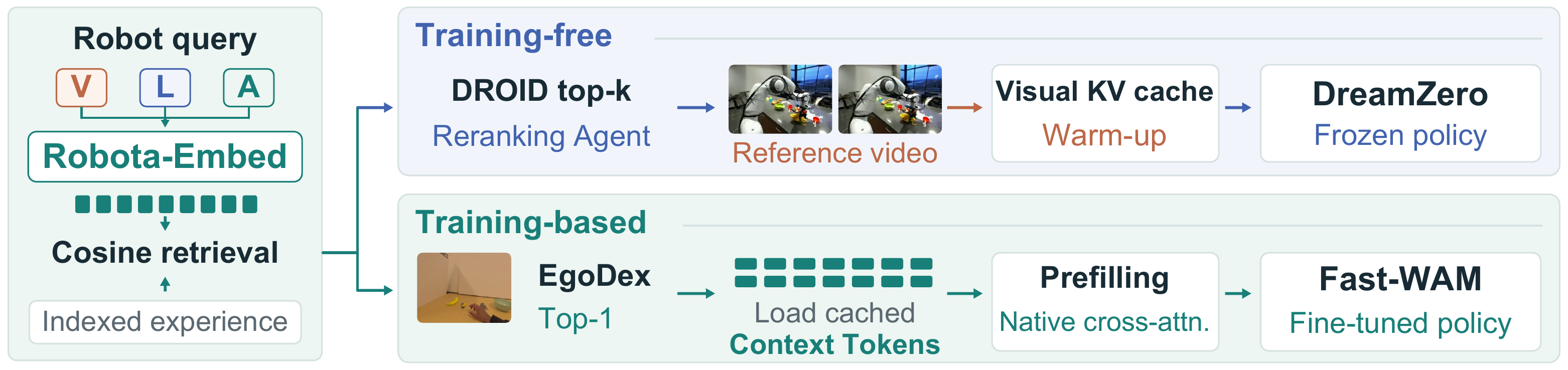}
  \par\vspace{-7pt}
  \ifdefined\EXPANDEDREPORT\captionsetup{skip=13pt}\fi
  \caption{\textbf{Embed2ICL: one retriever, two policy interfaces.}
  Reranked DROID videos warm up DreamZero's visual cache;
  top-1 EgoDex matches supply Context Tokens for Fast-WAM's native cross-attention.
  Retrieval vectors select experience; videos or tokens condition the policy.}
  \label{fig:physis-icl-framework}
  \submissionvspace{-7pt}
\end{figure}

\paragraph{Training-free ICL with DreamZero.}
\label{sec:exp-zero-shot-physical-icl}

Embed2ICL initializes the visual cache of frozen DreamZero-DROID~\citep{ye2026dreamzero}
with a retrieved video.
Table~\ref{tab:dreamzero-zero-shot-results} compares top-1 and reranked retrieval
on three DROID-Sim tasks. Both retrievers use the same GPT-5.6 sol Reranking Agent
to select task-relevant four-second references with completion evidence.
All six conditions use 100 trials per task, including the random-reference control.
Embed2ICL uses one fixed reference video per task across all evaluation trials
(\detailref{sec:dreamzero-warmup-details}).

\begin{table}[!ht]
  \ifdefined\EXPANDEDREPORT
    \submissionvspace{-2pt}
  \else
    \submissionvspace{-14pt}
  \fi
  \caption{\textbf{Training-free DreamZero-DROID in DROID-Sim.}
  Success rates (\%); parentheses show percentage-point changes from no
  warm-up. Reranked conditions share the same agent.}
  \label{tab:dreamzero-zero-shot-results}
  \centering
  \begingroup
\small
\setlength{\tabcolsep}{2.8pt}
\renewcommand{\arraystretch}{1.05}
\begin{tabular*}{\linewidth}{@{\extracolsep{\fill}}lr@{\,\extracolsep{0pt}}l@{\hspace{6pt}\extracolsep{\fill}}r@{\,\extracolsep{0pt}}l@{\hspace{6pt}\extracolsep{\fill}}r@{\,\extracolsep{0pt}}l@{\hspace{6pt}\extracolsep{\fill}}r@{\,\extracolsep{0pt}}l@{}}
\toprule
\tablegroup{Reference condition}
  & \multicolumn{2}{c}{\tablegroup{Cube $\rightarrow$ Bowl}}
  & \multicolumn{2}{c}{\tablegroup{Can $\rightarrow$ Mug}}
  & \multicolumn{2}{c}{\tablegroup{Banana $\rightarrow$ Bin}}
  & \multicolumn{2}{c}{\tablegroup{Average}} \\
\midrule
No warm-up                 & 61 & & 13 & & 67 & & 47.0 & \\
Random reference           & 50 & ($-11$) & 7 & ($-6$) & 51 & ($-16$) & 36.0 & ($-11.0$) \\
\addlinespace[3pt]
\multicolumn{9}{@{}l}{Omni-Embed-Nemotron-3B} \\
\quad Top-1                & 57 & ($-4$) & 5 & ($-8$) & 68 & ($+1$) & 43.3 & ($-3.7$) \\
\quad + reranking          & 62 & ($+1$) & 19 & ($+6$) & 72 & ($+5$) & 51.0 & ($+4.0$) \\
\addlinespace[3pt]
\embeddingmodelname{} Top-1 & 63 & ($+2$) & 16 & ($+3$) & 72 & ($+5$) & 50.3 & ($+3.3$) \\
\tablegroup{\methodname{}}      & \tablebest{65} & ($+4$) & \tablebest{24} & ($+11$)
  & \tablebest{80} & ($+13$) & \tablebest{56.3} & ($+9.3$) \\
\bottomrule
\end{tabular*}
\endgroup

  \submissionvspace{-8pt}
\end{table}

Embed2ICL performs best on all three tasks, averaging 56.3\% success versus
47.0\% without warm-up and 36.0\% with random references.
Reranking improves both retrievers.
\embeddingmodelname{} exceeds Omni-Embed-Nemotron-3B before reranking
(50.3\% versus 43.3\%) and with the same reranker
(56.3\% versus 51.0\%; \detailref{sec:dreamzero-reference-diagnostics}).
Random references underperform no warm-up: adding a demonstration is not
sufficient. Selected references preserve placement relations despite changes
in objects or receptacles, suggesting that useful context need not reproduce
the target scene exactly.

On YAM, Embed2ICL raises Banana-to-Bowl mean task progress from 21.67 to 40.00
(Table~\ref{tab:physis-icl-real-robot-results}). Both conditions use the same
DreamZero-AgiBot policy adapted to YAM, with no further policy updates for reference warm-up
(\detailref{sec:dreamzero-real-robot-details}).

\paragraph{Training-based ICL with Fast-WAM.}
\label{sec:exp-training-based-physical-icl}

\begin{figure}[!htb]
  \submissionvspace{-5pt}
  \centering
  \includegraphics[width=\linewidth]{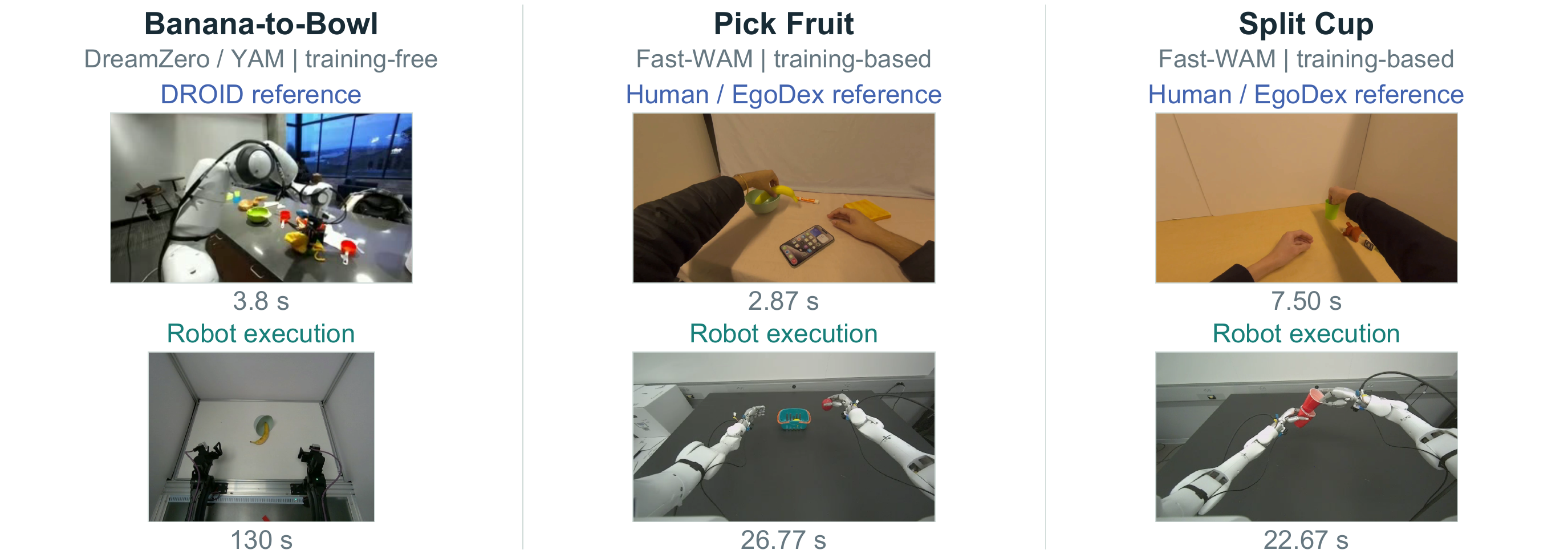}
  \par\vspace{-9pt}
  \ifdefined\EXPANDEDREPORT\captionsetup{skip=16pt}\fi
  \caption{\textbf{Retrieved references and real-robot execution.}
  Each task pairs a retrieved reference (top) with real-robot execution (bottom).
  References come from DROID for DreamZero/YAM and EgoDex for Fast-WAM.
  Timestamps are relative to each clip's start.}
  \label{fig:fastwam-real-robot-storyboard}
  \submissionvspace{-12pt}
\end{figure}

For Fast-WAM~\citep{yuan2026fastwam}, \embeddingmodelname{} retrieves the
top-ranked EgoDex demonstration. Cached Context Tokens from its multimodal
temporal states condition the policy (Figure~\ref{fig:physis-icl-framework}).
We train the policy to use this context while keeping the offline embedding
encoder fixed (\detailref{sec:fastwam-injection-details}).
\ifdefined\TECHREPORTPROMOTEFASTWAM
\ifdefined\TECHREPORTPROMOTEFASTWAM
  \subsubsection{Fast-WAM Context Interface}
\else
  \subsection{Fast-WAM Integration Details}
\fi
\label{sec:fastwam-injection-details}

\paragraph{Context Tokens from target states.}
The target encoder jointly encodes task text, future video, and native
actions. From valid action slots after layer 28's generation RMSNorm,
we extract fused states $H_i^k\in\mathbb{R}^{T_i\times2048}$ used by the
global retrieval branch (Eq.~\ref{eq:rerum-readout}). Adaptive temporal average
pooling retains at most 32 positions; standard windows from all four
sources yield 32. Pooled states pass through shared LayerNorm,
$2048\!\rightarrow\!1024$ projection $W$, and target residual adapter $A_k$:
\begin{equation}
    \begin{aligned}
      \widetilde H_i&=\operatorname{Pool}_{32}(H_i^k),\qquad
      u_{i,j}=A_k\!\left(W\operatorname{LN}(\widetilde H_{i,j})\right),\\
      C_{i,j}&=u_{i,j}/\|u_{i,j}\|_2,\qquad
      C_i\in\mathbb{R}^{32\times1024}.
    \end{aligned}
    \label{eq:fastwam-context-tokens}
\end{equation}
Standard windows yield 32 independently L2-normalized tokens from temporally
pooled states, not copies of the global retrieval vector. Cosine search uses
$k_i$; Fast-WAM consumes $C_i$.

\paragraph{Retrieval-trained parameters.}
Context Tokens reuse the Transformer, shared projection, and endpoint adapters
learned during retrieval training. The video VAE remains frozen. Token
extraction adds no separate head or token-level alignment or reconstruction loss.

\begin{figure}[!ht]
  \submissionvspace{-3pt}
  \centering
  \includegraphics[width=\linewidth,trim=0 30bp 0 0,clip]{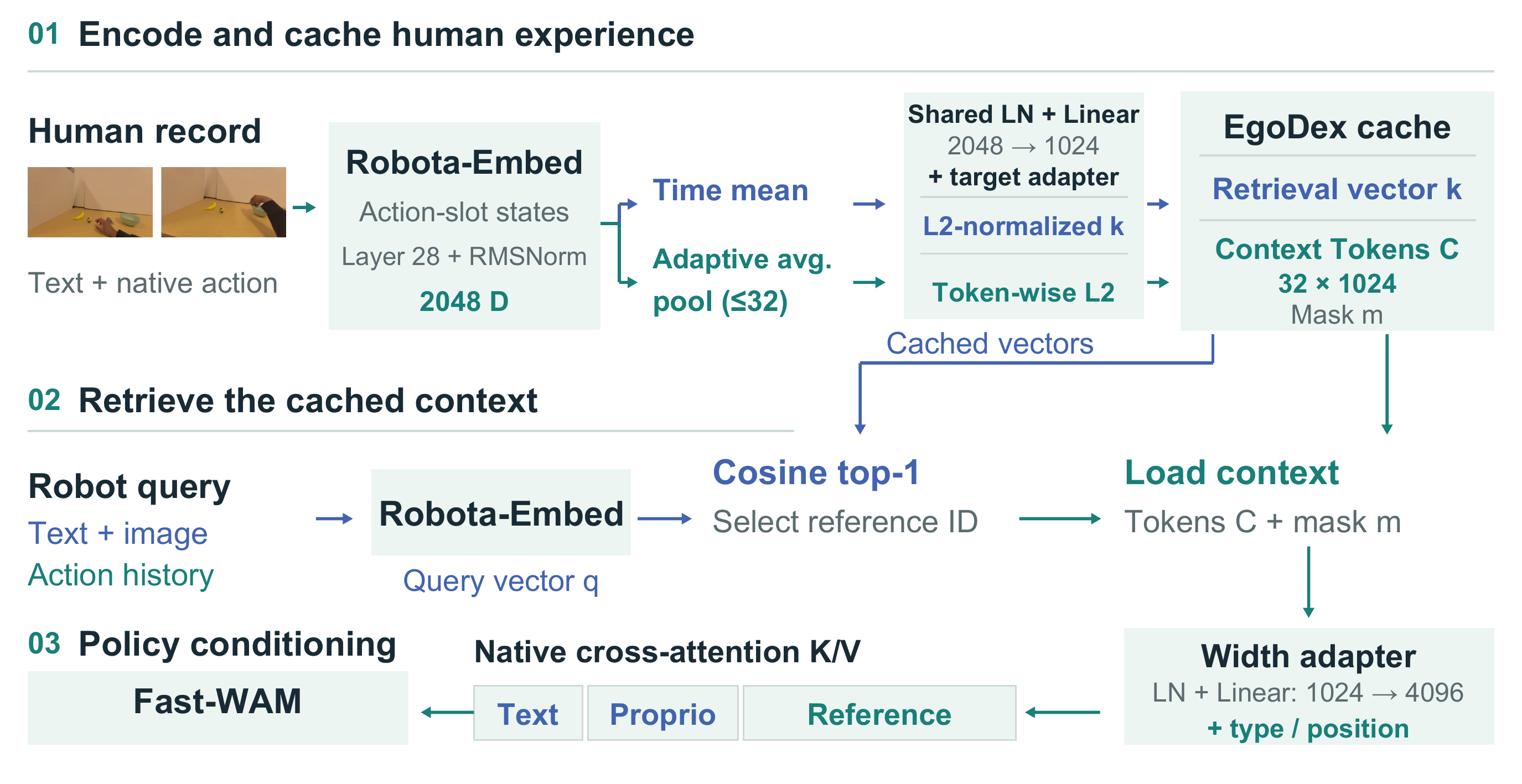}
  \caption{\textbf{Training-based Embed2ICL.}
  Offline encoding caches each candidate's retrieval vector, Context Tokens,
  and context mask. Online cosine top-1 search selects a reference and loads
  its tokens and mask. The prefilling adapter projects the tokens to Fast-WAM's
  context width and appends them to the native text and proprioceptive
  conditioning for cross-attention.}
  \label{fig:fastwam-injection}
  \submissionvspace{-4pt}
\end{figure}

\ifdefined\EXPANDEDREPORT\else\begin{samepage}\fi
\submissionvspace{-3pt}
\paragraph{Offline cache and online lookup.}
Offline, we cache each candidate's retrieval vector, Context Tokens, and
context mask $m_i$. Online, we encode the robot query as $q$, find
$i^*=\arg\max_i q^\top k_i$, and load $(C_{i^*},m_{i^*})$ without re-encoding
the candidate. Each rollout starts without reference context.
After the first successful retrieval, that reference stays active until the
rollout ends. Task-progress metadata is excluded from conditioning.
\par
\ifdefined\EXPANDEDREPORT\else\end{samepage}\fi

\ifdefined\EXPANDEDREPORT\else\begin{samepage}\fi
\submissionvspace{-3pt}
\paragraph{Prefilling.}
Embed2ICL projects $C=C_{i^*}$ to Fast-WAM's conditioning width:
\begin{equation}
    R
    = \operatorname{Linear}_{1024\rightarrow4096}
      \!\left(\operatorname{LN}(C)\right)
      + E_{\mathrm{type}} + E_{\mathrm{pos}},
    \quad R\in\mathbb{R}^{32\times4096}.
    \label{eq:fastwam-prefill-adapter}
\end{equation}
The projected sequence extends native conditioning
$[H_{\mathrm{text}};h_{\mathrm{proprio}}]$; $E_{\mathrm{type}}$ and
$E_{\mathrm{pos}}$ are learned type and position embeddings. Native cross-attention reads it
in all 30 video-expert and 30 action-expert layers, at every
action-diffusion denoising step, and during video KV-cache prefilling.
\par
\ifdefined\EXPANDEDREPORT\else\end{samepage}\fi

\ifdefined\EXPANDEDREPORT\else\begin{samepage}\fi
\paragraph{Policy-adaptation scope.}
Video and action flow-matching losses update approximately 5.00B video-expert
parameters, 1.02B action-expert parameters, the proprioception encoder, and a
4.34M-parameter projection adapter: approximately 6.02B trainable parameters
in total. The text encoder and VAE remain frozen. The context adapter inserts
cached tokens into native conditioning, with no gradient propagation to the
offline \embeddingmodelname{} encoder.
\par
\ifdefined\EXPANDEDREPORT\else\end{samepage}\fi

\paragraph{LIBERO training schedule.}
The retrieved-context condition starts from the official Fast-WAM LIBERO
checkpoint and trains on 1,712 trajectories comprising 277,713 samples
across the four LIBERO suites. Five epochs yield 21,700 parameter updates
with an effective batch size of 64. We use AdamW with a learning rate of
$10^{-4}$, a 5\% warm-up, and cosine decay to $10^{-6}$.
The random-reference condition evaluates this same policy checkpoint.
A separate no-ICL control uses the same 21,700 training updates
(\detailref{sec:fastwam-libero-details}).

\paragraph{Real-robot evaluation.}
Pick Fruit and Split Cup each use 12 rollouts per task and
condition. Following T-Rex's
partial-credit approach~\citep{niu2026trex}, we report
$100\,\frac{1}{12}\sum_{i=1}^{12}s_i$, where $s_i\in[0,1]$ is a rollout's
normalized task-progress score.

\paragraph{Additive scoring.}
Pick Fruit and Split Cup award 0.25 independently for each completed sub-step,
up to a total of 1. For \emph{Pick Fruit}, the four sub-steps are: the left
hand grasps a fruit from the left side; the left hand places that fruit in
the center basket; the right hand grasps a fruit from the right side; and
the right hand places that fruit in the center basket.
For \emph{Split Cup}, they are: the left hand grasps and stabilizes the cup
stack; the right hand separates exactly one cup; the right hand grasps the
separated cup; and the separated cup is placed upright on the table.

Table~\ref{tab:physis-icl-real-robot-results} reports mean task progress and
gains computed before rounding. Figure~\ref{fig:fastwam-real-robot-storyboard}
pairs retrieved references with illustrative executions for these two
tasks and DreamZero/YAM.
\fi

\paragraph{LIBERO difficult subset.}
Before ICL evaluation, an independent pilot selected five tasks on which the
official release succeeded in at most 46 of 50 trials. In 50 trials per task
and condition, Embed2ICL improves every task and raises average success from 86.4\% to 92.8\%
(Table~\ref{tab:fastwam-libero-summary}).

\begin{table}[!ht]
  \submissionvspace{-7pt}
  \caption{\textbf{Training-based ICL on the LIBERO difficult subset.}
  Successes out of 50 per task; last row: average success (\%).
  Parentheses give percentage-point gains over the official release.}
  \label{tab:fastwam-libero-summary}
  \centering
  \begingroup
\small
\setlength{\tabcolsep}{3pt}
\renewcommand{\arraystretch}{1.12}
\begin{tabular*}{\linewidth}{@{\extracolsep{\fill}}llrr@{\,\extracolsep{0pt}}l@{}}
\toprule
\tablegroup{Suite / ID} & \tablegroup{Abbreviated task} & \shortstack{\tablegroup{Official}\\\tablegroup{release}}
  & \multicolumn{2}{c}{\tablegroup{Embed2ICL}} \\
\midrule
Spatial / S4 & Black bowl $\rightarrow$ plate & 46 & \tablebest{49} & ($+6$) \\
Goal / G3 & Bowl $\rightarrow$ top drawer & 42 & \tablebest{50} & ($+16$) \\
Goal / G9 & Wine bottle $\rightarrow$ rack & 45 & \tablebest{47} & ($+4$) \\
LIBERO-10 / L4 & Two mugs $\rightarrow$ plates & 43 & \tablebest{44} & ($+2$) \\
LIBERO-10 / L6 & Mug--pudding arrangement & 40 & \tablebest{42} & ($+4$) \\
\midrule
\multicolumn{2}{l}{\tablegroup{Five-task average (\%)}} & 86.4
  & \tablebest{92.8} & ($+6.4$) \\
\bottomrule
\end{tabular*}
\endgroup

  \ifdefined\EXPANDEDREPORT
    \par\vspace{2pt}
  \else
    \par\vspace{-6pt}
  \fi
  \normalsize
  \caption{\textbf{LIBERO controls.} Five-task average success (\%);
  50 trials per task and condition.}
  \label{tab:fastwam-libero-context-controls}
  \label{tab:fastwam-libero-random-controls}
  \centering
  \begingroup
\small
\setlength{\tabcolsep}{3pt}
\renewcommand{\arraystretch}{1.10}
\begin{tabular*}{\linewidth}{@{\extracolsep{\fill}}ccc@{}}
\toprule
\shortstack{Official release\\(no ICL)} & \shortstack{Random\\Context Tokens} & \shortstack{\tablegroup{Embed2ICL}\\\tablegroup{(retrieved)}} \\
\midrule
86.4 & 88.0 & \tablebest{92.8} \\
\bottomrule
\end{tabular*}
\endgroup

  \ifdefined\EXPANDEDREPORT
    \vspace{3pt}
  \else
    \submissionvspace{-10pt}
  \fi
\end{table}

With the same policy checkpoint, retrieved Context Tokens outperform tokens from random
references by 4.8 percentage points (Table~\ref{tab:fastwam-libero-context-controls}).
After the same 21,700 training updates as Embed2ICL, the no-ICL continuation
still achieves 86.4\% (\detailref{sec:fastwam-libero-details}), versus 92.8\% with retrieved context.
Figure~\ref{fig:fastwam-libero-analysis} illustrates a retrieved human reference and robot execution.
\ifdefined\EXPANDEDREPORT\else
\paragraph{Real-robot evaluation.}
\fi
Both interfaces improve task progress on real robots
(Table~\ref{tab:physis-icl-real-robot-results}). With partial-credit scoring
inspired by T-Rex~\citep{niu2026trex}, Fast-WAM gains more on Split Cup than on Pick Fruit.
Split Cup requires one hand to stabilize a stack while the other separates a cup;
Pick Fruit scores each arm's grasp and placement into the center basket.

\begin{table}[!htb]
  \ifdefined\EXPANDEDREPORT
    \vspace{3pt}
  \else
    \submissionvspace{-12pt}
  \fi
  \caption{\textbf{Embed2ICL on real robots.} Mean task progress on
  task-specific 0--100 scales over 12 rollouts per task and condition.
  Parentheses show gains computed before rounding.}
  \label{tab:physis-icl-real-robot-results}
  \label{tab:fastwam-real-robot-results}
  \centering
  \normalsize
  \renewcommand{\arraystretch}{1.10}
  \setlength{\tabcolsep}{5.0pt}
  \begin{tabular*}{\linewidth}{@{\extracolsep{\fill}}lrr@{\,\extracolsep{0pt}}l@{}}
  \toprule
  \tablegroup{Task} & \tablegroup{No ICL}
    & \multicolumn{2}{c}{\tablegroup{Embed2ICL}} \\
  \midrule
  \multicolumn{4}{@{}l}{\tablegroup{Training-free: DreamZero}} \\
  Banana $\rightarrow$ Bowl & 21.67 & \tablebest{40.00} & ($+18.33$) \\
  \midrule
  \multicolumn{4}{@{}l}{\tablegroup{Training-based: Fast-WAM}} \\
  Pick Fruit & 75.00 & \tablebest{79.17} & ($+4.17$) \\
  Split Cup & 66.67 & \tablebest{87.50} & ($+20.83$) \\
  \bottomrule
\end{tabular*}

  \submissionvspace{-8pt}
\end{table}

\ifdefined\EXPANDEDREPORT
These progress gains complement binary LIBERO success rates but are not
directly comparable across tasks: each real-robot score reflects task-specific
completion stages. The references span robot and human experience, while each
policy retains its own conditioning interface. Retrieval quality and the form
of policy context therefore need to be evaluated together.
\fi
\ifdefined\TECHREPORTPROMOTEYAM
  \FloatBarrier
\subsection{Training-Free Embed2ICL on the YAM Robot}
\label{sec:dreamzero-real-robot-details}

\paragraph{Policy and reference.}
We adapt DreamZero-AgiBot~\citep{ye2026dreamzero} to YAM for 20 epochs on
55 play trajectories, comprising approximately 30 minutes of interaction.
Both conditions share this adapted policy; Embed2ICL adds retrieved video
through visual KV-cache warm-up without further policy updates. It reuses
the banana-placement reference selected by
\embeddingmodelname{} and the Reranking Agent for the DROID-Sim
Banana-to-Bin task. This DROID reference depicts placement into a bowl;
the real-robot task is Banana-to-Bowl.

\paragraph{Evaluation.}
We evaluate each condition over 12 real-robot rollouts. Success requires
the banana to remain inside the bowl while the bowl stays upright.
We also report mean cumulative task progress on a 0--100 scale, awarding
20 points for each completed stage: (1) gripper contact with the banana;
(2) grasping and lifting it from the table; (3) transporting it above the
bowl; (4) releasing it into the bowl; and (5) stable placement with the bowl
upright. Embed2ICL raises mean progress from 21.67 to 40.00
(Table~\ref{tab:physis-icl-real-robot-results}); success increases from
0.0\% to 16.7\%.

\ifdefined\TECHREPORTPROMOTEYAM
\else
\ifdefined\EXPANDEDREPORT
\else
\begin{figure}[!ht]
  \centering
  \relax
  \par\vspace{-4pt}
  \caption{\textbf{Retrieved reference and DreamZero execution on YAM.}
  Simulation and YAM share the DROID banana-placement reference.
  Frames show a selected real-robot rollout's initial state, transport,
  bowl interaction, and final visible state. Times are relative to each
  video's start.}
  \label{fig:dreamzero-real-robot-storyboard}
  \submissionvspace{-4pt}
\end{figure}
\fi
\fi
\fi
\ifdefined\EXPANDEDREPORT\FloatBarrier\fi
\subsection{Embed2Evolve: Embedding-Guided Context Evolution}
\label{sec:physis-evolve}

\begin{figure}[!ht]
  \submissionvspace{-9pt}
  \centering
  \includegraphics[width=\linewidth]{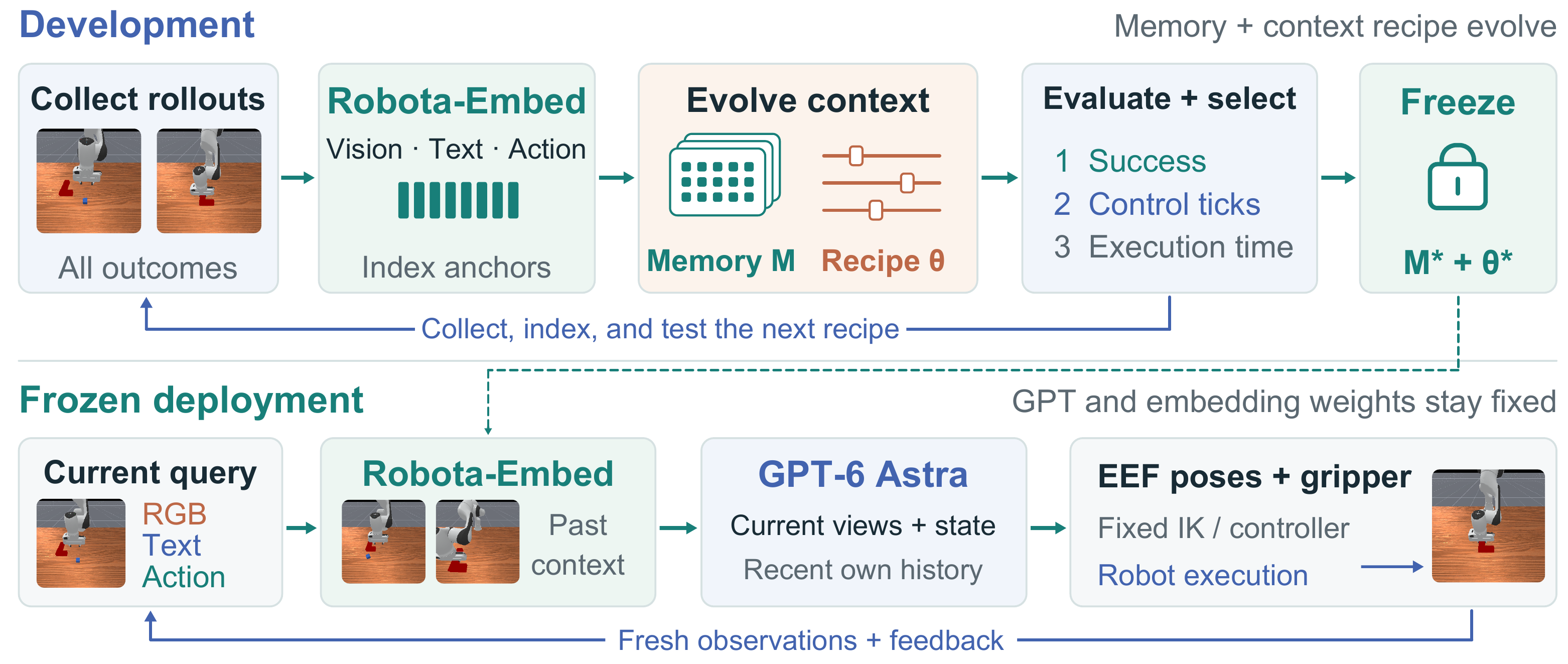}
  \par\vspace{-10pt}
  \ifdefined\EXPANDEDREPORT\captionsetup{skip=14pt}\fi
  \caption{\textbf{Embed2Evolve.} Development grows memory and optimizes its
  use; testing fixes both. GPT-6 Astra proposes actions for a fixed controller.
  The model comparison replaces \embeddingmodelplain{} in both phases.
  Images show the PullCubeTool task.}
  \label{fig:physis-evolve-framework}
  \submissionvspace{-10pt}
\end{figure}

Embed2Evolve builds task-local memory and refines how it is used through
repeated attempts. This memory and context recipe form a \emph{harness}; GPT-6 Astra and
\embeddingmodelname{} remain fixed (Figure~\ref{fig:physis-evolve-framework}).

\paragraph{Memory and context search.}
\ifdefined\EXPANDEDREPORT
Completed attempts, successful or failed, add indexed continuations of up to
four seconds. Deterministic search varies retrieval ranking and count, refresh
cadence, and retained history, prioritizing success before successful-execution
efficiency. The selected memory and recipe are frozen for fresh-test
initializations (Figure~\ref{fig:physis-evolve-harness};
\detailref{sec:evolve-harnesses}). Along PullCubeTool's
R4$\rightarrow$R5$\rightarrow$R8 path, memory grows from 938 to 1,667 anchors
while retained history shrinks from eight to four pairs. The current VLA query
retrieves historical RGB, actions, and feedback to guide GPT-6 Astra within an
unchanged action interface (\detailref{sec:evolve-interface}).
\else
Completed attempts, successful or failed, add indexed continuations of up to
four seconds to memory. Deterministic search varies memory ranking and count,
refresh cadence, and retained history. Development prioritizes
success, using successful-execution efficiency to break ties.
The selected memory and recipe are then fixed for testing on fresh task initializations
(Figure~\ref{fig:physis-evolve-harness}; \detailref{sec:evolve-harnesses}).
Along PullCubeTool's R4$\rightarrow$R5$\rightarrow$R8 development path, memory
grows from 938 to 1,667 anchors while retained history shrinks from eight
to four pairs. A larger store of experience need not produce a longer
context for the policy.

\paragraph{Context-guided control.}
The current VLA query retrieves historical RGB, actions, and feedback.
Context guides GPT-6 Astra's commands within an unchanged action interface
(\detailref{sec:evolve-interface}).
\fi

\paragraph{Evaluation.}
Five development-selected ManiSkill tasks~\citep{tao2025maniskill3} use
20 fresh matched initializations per condition. GPT-6 Astra Direct (Direct)
uses within-episode history; GPT-6 Astra + random history adds a fixed random
same-task RGB clip. These conditions use the same policy, current observations,
controller, and per-rollout control budgets.

\paragraph{Embedding-model comparison.}
We replace only \embeddingmodelname{} with Omni-Embed-Nemotron-3B~\citep{xu2025omni}
and independently repeat evolution. Both conditions share the policy, prompt,
initial memory, seeds, proposal and stopping rules, and eight-round cap per task.
Both retrieve actions and feedback, but their resulting memory banks and
context recipes may differ.

\begin{table}[!ht]
  \submissionvspace{-9pt}
  \centering
  \caption{\textbf{Embed2Evolve in ManiSkill simulation.} Success (\%) over
  20 trials per task and condition; mean policy tokens per test rollout, including failures.}
  \label{tab:physis-evolve-main}
  \begingroup
\small
\setlength{\tabcolsep}{3pt}
\renewcommand{\arraystretch}{1.10}
\sbox0{\shortstack[c]{\tablegroup{GPT-6 Astra}\\\tablegroup{+ random}\\\tablegroup{history}}}
\newcommand{\evolveheader}[1]{%
  \sbox2{\shortstack[c]{#1}}%
  \raisebox{.5\dimexpr\ht0-\dp0-\ht2+\dp2\relax}[\ht0][\dp0]{\usebox2}}
\begin{tabular*}{\linewidth}{@{\extracolsep{\fill}}lcccc@{}}
\toprule
\evolveheader{\tablegroup{Task}}
& \evolveheader{\tablegroup{GPT-6 Astra}\\\tablegroup{Direct}}
& \evolveheader{\tablegroup{GPT-6 Astra}\\\tablegroup{+ random}\\\tablegroup{history}}
& \evolveheader{\tablegroup{Evolve +}\\\tablegroup{Omni-Embed-}\\\tablegroup{Nemotron-3B}}
& \evolveheader{\tablegroup{Embed2Evolve}} \\
\midrule
Poke Cube & 45 & 15 & 45 & \tablebest{80} \\
Stack Cube & 10 & 20 & 20 & \tablebest{60} \\
Lift Peg Upright & 30 & 40 & 55 & \tablebest{65} \\
Pull Cube with Tool & 10 & 10 & 20 & \tablebest{50} \\
Two-Robot Stack Cube & 25 & 15 & 25 & \tablebest{85} \\
\midrule
\tablegroup{Average (\%)} & 24.0 & 20.0 & 33.0 & \tablebest{68.0} \\
\tablegroup{Tokens/rollout (M)} & 0.921 & -- & $\approx$2.531 & \tablebest{0.518} \\
\bottomrule
\end{tabular*}
\endgroup

  \submissionvspace{-9.1pt}
\end{table}

\FloatBarrier
The Omni-Embed-Nemotron-3B harness raises success from Direct's 24\% to 33\%,
while mean policy tokens rise from 0.921M to approximately 2.531M per test
rollout (Table~\ref{tab:physis-evolve-main}). Embed2Evolve reaches 68\% at
0.518M tokens, improving success on every task while using approximately 79.5\% fewer
tokens than this independently evolved baseline.
Under matched evolution settings, the embedding model affects the resulting
harness's success and token use.

\paragraph{Tool use and dual-arm coordination.}
Within the same action interface, retrieved continuations may narrow
GPT-6 Astra's effective decision space through task- and stage-specific
examples of tool contact, motion direction, and arm coordination.
Against Omni-Embed-Nemotron-3B, success rises from 20\% to 50\% on Pull Cube
with Tool and from 25\% to 85\% on Two-Robot Stack Cube.

\paragraph{Test-time efficiency.}
\ifdefined\EXPANDEDREPORT
Compared with Direct, Embed2Evolve uses 43.7\% fewer tokens across all test
rollouts and reduces reference API cost by 41.8\%, from \$9.9641 to \$5.7980
(\detailref{sec:evolve-token-cost}). Most savings come from inputs.
On 19 shared-success pairs, policy decisions fall by 64.1\% and observed
client execution time by 41.4\%. Both reductions average task-wise percentages
equally across the five tasks; the mean task-wise Direct-to-Embed2Evolve time ratio is
$1.90\times$. Execution time includes API waits and is affected by actual
caller and scheduling differences (\detailref{sec:evolve-test-protocol});
the time savings are observational, not an isolated method speedup.
Fewer policy calls need not reduce physical control ticks
(Table~\ref{tab:physis-evolve-control-ticks}).
\else
Compared with Direct, Embed2Evolve uses 43.7\% fewer tokens across all test
rollouts and reduces reference API cost by 41.8\%, from \$9.9641 to \$5.7980
(\detailref{sec:evolve-token-cost}).
Most savings come from inputs. On the 19 shared-success pairs, policy
decisions fall by 64.1\% and client execution time, including API waits,
falls by 41.4\%. Both reductions average task-wise percentages equally
across the five tasks. Averaging the task-specific Direct-to-Embed2Evolve
time ratios gives a $1.90\times$ mean speedup
(\detailref{sec:evolve-test-protocol}). These results are consistent with
context focusing the policy on relevant next actions.
Fewer policy calls do not necessarily reduce the number of physical control ticks
(Table~\ref{tab:physis-evolve-control-ticks}).
\fi
\ifdefined\TECHREPORTPROMOTEEVOLVE
  \relax
\fi
\section{Discussion}
\label{sec:report-discussion}

\paragraph{Retrieval quality has two levels.}
Episode recall measures experience identification; clip recall measures
continuation localization. Their gap shows that matching a task and embodiment
need not identify the execution phase. Same-instruction evaluation and
candidate-text masking probe temporal discrimination and target-language use
beyond episode identification.

\paragraph{The modalities play complementary roles.}
Language specifies the task, action supplies motion and phase, and vision
localizes the interaction. Action adds information beyond vision and language.
However, retraining permits adaptation to remaining inputs, whereas masking
perturbs learned fusion. The no-vision retraining/masking gap supports a joint
VLA key, not a universal modality ranking.
Accordingly, masking characterizes the trained model's use of each input,
not the attainable performance of separately trained unimodal systems.

\paragraph{Retrieved context is useful only through a compatible interface.}
Random references hurt DreamZero relative to no warm-up (36.0\% versus
47.0\%), whereas Robota-Embed's raw retrieval and reranked Embed2ICL reach
50.3\% and 56.3\%. Fast-WAM gains from retrieved rather than random Context
Tokens under fixed policy weights (92.8\% versus 88.0\%). More context is not
inherently better: reference selection, temporal content, and conditioning
jointly matter, without isolating each component's causal contribution.

\paragraph{Memory size and policy context are different resources.}
Embed2Evolve can grow memory while retaining less history and refreshing less
often. Independent evolution with Omni-Embed-Nemotron-3B compares discovered
memories and harnesses, not a fixed-harness encoder swap. Token and reference-cost metrics
cover all test rollouts; decision/time comparisons use shared successes.
Fewer policy calls need not mean fewer control ticks or shorter robot motion.

\paragraph{Future work.}
Future work will extend Robota-Embed to more robot embodiments with diverse
kinematics, native action spaces, and control interfaces. We will also explore
variable-length retrieval that adapts the temporal extent of recorded
continuations to the task and execution stage, rather than using a fixed
four-second window. The goal is to retrieve only the duration of experience
needed for the current interaction.
\FloatBarrier
\section{Conclusion}
\label{sec:conclusion}
\label{sec:discussion-limitations}

\ifdefined\EXPANDEDREPORT
  \embeddingmodelname{} jointly encodes task intent, visual state, and native
  action for physical-experience retrieval; action improves retrieval beyond
  language and vision. This shared space supports Embed2ICL conditioning and
  Embed2Evolve memory evolution. Under matched settings, Embed2Evolve's success
  and efficiency gains show that representation shapes both retrieved context
  and the harness around a frozen policy.
\else
  \embeddingmodelname{} retrieves recorded continuations by jointly encoding
  task intent, visual state, and native action. Action improves retrieval beyond
  language and vision, supporting retrieved-context conditioning in Embed2ICL
  and memory-based context evolution in Embed2Evolve.
  Under matched evolution settings, Embed2Evolve achieves higher
  success with fewer policy tokens than its Omni-Embed-Nemotron-3B counterpart.
  Experience representation affects both the context retrieved for a decision
  and the harness developed around a frozen policy.
\fi

\FloatBarrier
\typeout{TECHREPORT_MAIN_END_PAGE=\thepage}

\ifhasreferences
  \clearpage
  \typeout{TECHREPORT_REFERENCES_START_PAGE=\thepage}
  {\small
    \setlength{\bibsep}{5.5pt plus 0.5pt minus 0.5pt}

  }
\fi

\clearpage
\appendix
\def\TECHREPORTAPPENDIX{1}
\typeout{TECHREPORT_APPENDIX_START_PAGE=\thepage}
\clearpage
\section*{Appendix}
\ifdefined\TECHREPORTPROMOTEMETHOD
  \section{Training Data and Curation}
\else
  \section{\embeddingmodelname{} Method Details}
\fi
\label{app:pai-objectives}

\ifdefined\TECHREPORTPROMOTEMETHOD
\else
\subsection{Architecture and Optimization}
\label{sec:rerum-implementation}

\paragraph{Paired inputs.}
Queries use instructions, synchronized current actions, and SAM3~\citep{carion2026sam3}
interaction crops. Targets pair instructions with 16 frames and native-rate
actions from strictly future four-second windows. We favor less-occluded
third-person views; EgoDex is egocentric.

\paragraph{Shared backbone and endpoint adapters.}
\embeddingmodelname{} uses Cosmos3-Edge 4B, with 28 Transformer layers
and hidden width 2,048. The Transformer, shared retrieval head, endpoint
adapters, and temperature are initialized from its pretrained checkpoint.
Query and target share the Transformer, video tokenizer, and embedding
projection head. Separate bias-free residual linear adapters specialize the
current-state and future-clip endpoints. Each source retains
its native action projector, trained jointly with the Transformer.
For endpoint $r\in\{q,k\}$, let
$H^r\in\mathbb{R}^{T_r\times2048}$ denote the multimodally fused hidden states
at native action-token positions, after the generation RMSNorm following
layer 28. These states jointly encode the endpoint's language, vision,
and action. Let $g_r$ denote the endpoint readout, comprising the shared
LayerNorm and projection $W\in\mathbb{R}^{1024\times2048}$ together with
the endpoint-specific residual adapter. Temporal pooling and normalization give
\begin{equation}
  \bar h^r=\frac{1}{T_r}\sum_{t=1}^{T_r}H^r_t,\qquad
  z^r=\frac{g_r(\bar h^r)}{\|g_r(\bar h^r)\|_2}.
  \label{eq:rerum-readout}
\end{equation}
The shared projection is bias-free. Benchmark retrieval uses cosine
similarity; only training-free Embed2ICL reranks. Temporal pooling of the
target states, shared projection, and per-token normalization yield
Embed2ICL's cached Context Tokens
(\detailref{sec:fastwam-injection-details}).

\paragraph{Native actions.}
Table~\ref{tab:rerum-action-domains} specifies the native action interfaces.
For each source, we compute per-coordinate quantiles $q_{01}$ and $q_{99}$
using its training partition and normalize actions as
\begin{equation}
  \hat a=2\frac{a-q_{01}}{\max(q_{99}-q_{01},10^{-8})}-1.
  \label{eq:rerum-action-normalization}
\end{equation}
Values remain unclipped, and zero-padding to 64 dimensions standardizes tensor
shape. The source identifier selects the native projector; future actions
retain their native sampling rate.

\begin{table}[!ht]
  \submissionvspace{-3pt}
  \centering
  \caption{\textbf{Native action interfaces.} Dimensions and rates describe
  native actions before normalization and padding. Domain IDs route actions
  to their Cosmos3-Edge projectors.}
  \label{tab:rerum-action-domains}
  \begin{tabular*}{\linewidth}{@{\extracolsep{\fill}}lrrr@{}}
  \toprule
  \tablegroup{Source} & \tablegroup{Domain ID} & \tablegroup{Action dimensions} & \tablegroup{Rate (Hz)} \\
  \midrule
  EgoDex & 3 & 57 & 30 \\
  DROID & 8 & 10 & 15 \\
  LIBERO & 5 & 10 & 10 \\
  RoboMIND--Franka-dual & 12 & 20 & 10 \\
  \bottomrule
\end{tabular*}

  \submissionvspace{-4pt}
\end{table}

\paragraph{Optimization.}
We use AdamW with $\beta=(0.9,0.95)$ and gradient clipping at $1.0$.
The Transformer learning rate is $3\times10^{-7}$, and the
projection head and endpoint adapters use $10^{-4}$. Training runs for 26,275
optimizer steps, with 1,314 steps of linear warm-up followed by cosine decay,
global batch size 512, and equal sampling of EgoDex,
DROID, LIBERO, and RoboMIND--Franka-dual ($1{:}1{:}1{:}1$).
The optimizer and scheduler are reinitialized for retrieval training.
Both parameter groups use weight decay $0.05$; the video VAE is frozen.
We use the final-step checkpoint (26,275).
\subsection{\embeddingmodelname{} Objective}
\label{sec:pai-objective-details}

The objective (Eq.~\ref{eq:composite-objective}) combines predictive
retrieval alignment, multimodal grounding and discrimination, and
representation regularization. Table~\ref{tab:pai-objectives} lists term
coefficients and ranking margins.

\begin{table}[ht]
  \submissionvspace{-3pt}
  \centering
  \caption{\textbf{\embeddingmodelname{} training objective.} Loss coefficients
  and ranking margins for each term, organized into the three objective groups
  in Eq.~\ref{eq:composite-objective}.}
  \label{tab:pai-objectives}
  {\small\renewcommand{\arraystretch}{0.96}%
  \begin{tabular}{@{}>{\raggedright\arraybackslash}p{0.17\linewidth}
  >{\raggedright\arraybackslash}p{0.54\linewidth}
  >{\raggedleft\arraybackslash}p{0.08\linewidth}
  >{\raggedleft\arraybackslash}p{0.08\linewidth}@{}}
\toprule
\tablegroup{Objective} & \tablegroup{Constraint or positive relation} & \tablegroup{Weight} & \tablegroup{Margin} \\
\midrule
\multicolumn{4}{@{}l}{\tablegroup{Predictive retrieval alignment}} \\
Clip-level retrieval & Exact future-clip identity; bidirectional InfoNCE & 1.00 & -- \\
\addlinespace[2pt]
\multicolumn{4}{@{}l}{\tablegroup{Multimodal grounding and discrimination}} \\
Text auxiliary & Text-only query; matched task & 0.15 & -- \\
Image auxiliary & Image-only query; matched interaction object & 0.20 & -- \\
Action auxiliary & Disabled; future identity uses multimodal context & 0.00 & -- \\
Text swap & Replacing query text must reduce similarity & 0.25 & 0.08 \\
Image swap & Replacing the query zoom must reduce similarity & 0.35 & 0.10 \\
Action swap & Replacing current action must reduce similarity & 0.20 & 0.06 \\
Hard task & Same object roles; operation/relation/direction conflict & 0.25 & 0.08 \\
Hard object & Same task relation; interaction-object category conflict & 0.35 & 0.10 \\
Hard action & Same task/objects; action-primitive/phase conflict & 0.20 & 0.06 \\
Fused dominance & Fused query must outperform every unimodal query & 0.20 & 0.02 \\
\addlinespace[2pt]
\multicolumn{4}{@{}l}{\tablegroup{Representation regularization}} \\
Isotropy & Variance, participation-ratio rank, and centering & 0.05 & -- \\
Query center & Direct query-centering penalty & 0.10 & -- \\
Target center & Direct target-centering penalty & 0.10 & -- \\
\bottomrule
\end{tabular}
}
  \submissionvspace{-4pt}
\end{table}

\paragraph{Positive, negative, and ignored pairs.}
Positive pairs share the exact future-clip ID. For query $q_i$, $k_i$ is
the paired positive ($k^+$ in Figure~\ref{fig:pai-embedding-objective}).
Each pair links a current state to its recorded future clip within the same source.
Let $s(q,k)$ be cosine similarity and
$\mathcal{N}(i)$ contain only candidates with a different clip ID and an
explicit semantic conflict. The query-to-target term is
\begin{equation}
  \mathcal{L}_{q\rightarrow k}=-\frac{1}{|\mathcal{B}|}
  \sum_{i\in\mathcal{B}}\log
  \frac{\exp(s(q_i,k_i)/\tau)}
       {\exp(s(q_i,k_i)/\tau)+
        \sum_{j\in\mathcal{N}(i)}\exp(s(q_i,k_j)/\tau)}.
  \label{eq:masked-info-nce}
\end{equation}
We independently build the reverse validity mask and average the directions:
$\mathcal{L}_{\mathrm{align}}=\tfrac12
(\mathcal{L}_{q\rightarrow k}+\mathcal{L}_{k\rightarrow q})$.
The global pool contains 512 entries before masking; each denominator retains
only the positive and valid negatives. We optimize the inherited
Cosmos3-Edge temperature $\tau$ within $[0.01,0.2]$.

\paragraph{Semantic conflict rules.}
We generate weak labels through structured parsing and deterministic rules.
Qwen3-VL-8B-Instruct~\citep{bai2025qwen3vl} parses task instructions into operation names and object
roles using non-sampling decoding. From this output, we construct signatures
for task relations, object identities, and object categories. The resulting
labels are rule-driven weak supervision, not human annotations.
Before similarity-based mining, we determine eligibility for
$\mathcal{N}(i)$ using four pairwise conflicts:
\emph{Task conflict} preserves object roles but changes the operation,
relation, or direction. \emph{Object conflict} preserves the task relation
but changes a valid interaction-object category. \emph{Action conflict} shares
task and objects but differs in the current-action primitive or future-action
phase. \emph{Joint conflict} changes both task and objects.

For same-episode action negatives, future windows with temporal
intersection-over-union greater than $0.25$ are treated as near-duplicates
and excluded. Cross-episode action negatives must additionally differ in
future gripper phase. These checks filter the action-conflict class;
semantically compatible, uncertain, or potentially duplicate candidates
are masked out of the negative set. A different clip ID alone is therefore
insufficient to make a pair negative.

Current-action primitives are obtained by scale-standardizing the
domain-normalized action and discretizing each coordinate at thresholds
$-0.5$ and $+0.5$. Future-action phase combines the dominant motion direction,
the timing of peak motion, and changes in gripper state. To compare future
actions, we sample 16 time points and concatenate the per-coordinate means,
standard deviations, and first-to-last differences. We compare these
descriptors using scale-normalized RMS distances.

\paragraph{Auxiliary alignment.}
Text-only and image-only auxiliary queries align with targets sharing task and
interaction-object identity, respectively. We disable standalone action-only
alignment to exact future clips: a single current action does not uniquely
determine a future trajectory. Action remains an input to fused retrieval
alignment, with action swaps discouraging the model from ignoring it.

\paragraph{Modality comparisons.}
Each swapped query replaces exactly one modality while keeping its paired
target fixed. Text donors share object identity but differ in task relation;
image donors share the task relation but differ in valid object category.
Action donors share the action domain, task, and object identity, but differ
in both current-action primitive and future-action phase. The standardized
RMS distances between their current actions and between their future
descriptors must each be at least $0.75$. These donor conditions are stricter
than the action-conflict eligibility rules.

Let $q'$ encode a valid swapped query. We use the cosine hinge loss
\begin{equation}
  \ell_{\mathrm{swap}}(q,q',k^+)
  =\max\!\left\{0,\,m-s(q,k^+)+s(q',k^+)\right\},
  \label{eq:modality-swap-hinge}
\end{equation}
with $m=0.08$ (text), $0.10$ (image), and $0.06$ (action).
We skip only the corresponding swap loss when no valid donor exists. A $0.02$
margin favors the fused query over each unimodal query.

\paragraph{Hard-negative constraints.}
The contrastive denominator uses all valid negatives. For the hard-task,
hard-object, and hard-action terms, we select the highest-similarity candidate
from the corresponding nonempty conflict class $c$ after validity masking:
$j_c^*(i)=\arg\max_{j\in\mathcal{N}_c(i)}s(q_i,k_j)$.
Their margins are $0.08$, $0.10$, and $0.06$, respectively.
Conditional hinge losses are averaged over valid samples within each rank,
then averaged across ranks.

\paragraph{Representation regularization.}
We encourage near-zero batch centers, a per-dimension
standard-deviation floor near $0.02$, and participation-ratio effective ranks
of at least 16 for query batches and 18 for target batches.
Center losses weighted at $0.10$ per
side reinforce isotropy-based centering.
\fi
\ifdefined\TECHREPORTPROMOTEMETHOD
\else
  \subsection{Training Data and Curation}
\fi
\label{sec:training-data-curation}

Our training dataset contains 1,695,684 filtered anchor-level examples from
EgoDex~\citep{hoque2026egodex}, DROID~\citep{khazatsky2024droid},
LIBERO~\citep{liu2023libero}, and
RoboMIND--Franka-dual~\citep{wu2025robomind}. Table~\ref{tab:pai-data} gives the
source composition. EgoDex contributes tracked human hand and body pose trajectories,
encoded by its source-specific native-action projector.
Before constructing dense anchors, we assign complete episodes, sessions, and
content identities to disjoint training, validation, and original test partitions.

\begin{table}[H]
  \submissionvspace{-3pt}
  \centering
  \caption{\textbf{Training-data composition.} Filtered anchor-level training
  examples by source.}
  \label{tab:pai-data}
  \begin{tabular}{@{}lr@{}}
\toprule
\tablegroup{Source domain} & \tablegroup{Training rows} \\
\midrule
EgoDex~\citep{hoque2026egodex} & 1,345,160 \\
DROID~\citep{khazatsky2024droid} & 26,563 \\
LIBERO~\citep{liu2023libero} & 136,671 \\
RoboMIND--Franka-dual~\citep{wu2025robomind} & 187,290 \\
\midrule
\tablegroup{Total} & \tablegroup{1,695,684} \\
\bottomrule
\end{tabular}

  \submissionvspace{-4pt}
\end{table}

\ifdefined\EXPANDEDREPORT
\ifdefined\TECHREPORTAPPENDIX
\else
\begin{figure}[!htbp]
  \submissionvspace{-3pt}
  \centering
  \includegraphics[width=\linewidth,trim=0 40bp 0 0,clip]{figures/roborag-2m-pipeline.pdf}
  \par\vspace{-4pt}
  \caption{\textbf{Multi-source training-data curation.} Episodes are split
  before dense anchor sampling. SAM3 localizes interaction objects; recomputed
  crops undergo visual quality checks. Accepted anchors are paired with
  video and actions recorded over the next four seconds.}
  \label{fig:roborag-2m-curation}
  \submissionvspace{-4pt}
\end{figure}
\fi
\fi

\paragraph{Anchor enumeration and object localization.}
For a source sequence with $N$ frames, let $a$ be a frame index and $h$ the
future horizon in frames. We enumerate dense anchors satisfying \mbox{$a+h+1<N$}.
We prompt SAM3 with phrases naming the instruction's physical interaction
objects, using a proposal threshold of 0.35. We retain detections with a final
score of at least 0.40 that cover at
least 0.08\% of the source image, and permit a zoom ratio of at least 1.10.
We independently recompute each accepted crop at $320\times192$.

\paragraph{Visual quality gate.}
Qwen3-VL-8B-Instruct~\citep{bai2025qwen3vl} verifies each crop deterministically. A crop passes only
if the model confirms with high confidence that the interaction object is
visible, matches the instruction, is separable from other instances, and is
unoccluded.
If verification fails, we resample an earlier neighboring timestamp, updating
the image, synchronized action, and temporal anchor together. Selection uses
only the instruction and current observations.

\paragraph{Future target construction.}
We pair each accepted anchor with the strictly future four-second action
sequence and 16 video frames from that interval. The window starts after the
accepted timestamp, including when resampling selects an earlier one.
Figure~\ref{fig:roborag-2m-curation} shows the pipeline.
\FloatBarrier
\section{\benchmarkname{} Evaluation and Diagnostics}
\label{app:roborag-bench}
\label{sec:appendix-roborag-bench}

\subsection{\benchmarkname{}: Composition and Evaluation}
\label{sec:benchmark-protocol-details}

\ifdefined\EXPANDEDREPORT
EgoDex, DROID, LIBERO, and RoboMIND--Franka-dual each contribute 1,000 held-out
queries from 800 unique episodes.
\else
EgoDex, DROID, LIBERO, and RoboMIND--Franka-dual each contribute 1,000 held-out
queries from 800 unique episodes (Table~\ref{tab:roborag-bench-composition}).
\fi
The shared 8,000-item gallery contains
2,000 candidates per source: 1,000 paired future clips and 1,000 nearby
same-episode clips. Nearby clips often share the positive's instruction,
objects, and visual content, requiring precise temporal localization.

\ifdefined\EXPANDEDREPORT
\else
\begin{table}[ht]
  \submissionvspace{-3pt}
  \centering
  \caption{\textbf{Composition of \benchmarkname{}.} All queries search the
  shared 8,000-clip gallery; task labels are counted within each source.}
  \label{tab:roborag-bench-composition}
  \small
  \setlength{\tabcolsep}{3.2pt}
  \relax
  \submissionvspace{-4pt}
\end{table}
\fi

The four sources contain 591, 800, 39, and 5 distinct task-instruction strings,
respectively, covering human experience, simulation, and single- and dual-arm
robots. DROID and LIBERO also support the downstream ICL studies, evaluated
separately from retrieval.

Clip-level retrieval assigns one positive to each query: its uniquely paired
future four-second window. Episode-level retrieval treats any indexed window
from the same source-scoped episode as relevant. Episode-level metrics average
query outcomes within each episode, then equally over the 800 episodes per
source and the four sources, giving each episode equal weight regardless of
its query count. In total, the benchmark contains 4,000 queries and 3,200
episodes. Training, validation, and original test
partitions are separated by complete episode, session, and content identity
before dense anchoring. Benchmark episodes are excluded from training;
task, object, and embodiment categories may overlap across partitions.
Every target is strictly later than its paired query anchor.
Ranking uses cosine similarity on L2-normalized embeddings, without reranking.
We evaluate the final \embeddingmodelname{} checkpoint at training step
26,275, the last step of retrieval training.

\paragraph{Baseline input interfaces.}
\embeddingmodelname{} queries combine instruction, a SAM3 object zoom, and
synchronized action; gallery entries combine instruction, future video, and
future action. Omni-Embed-Nemotron-3B and Qwen3-VL-Embedding-2B use instruction
and the uncropped current frame as the query, and instruction and the 16-frame
future clip as the candidate. CLIP, Perception Encoder, and InternVideo2 use
the instruction to retrieve the future video clip.
Gemini Embedding 2 and VLM2Vec-V2 likewise use instruction plus the original,
uncropped current image as the query, without action input, and instruction
plus the future four-second video as the gallery entry. For Gemini
Embedding 2, we upload a video assembled from 16 frames and obtain normalized
768-dimensional vectors. VLM2Vec-V2 receives the 16 frames directly and uses
last-token pooling followed by L2 normalization, producing 1,536-dimensional
vectors. Both are evaluated on all 4,000 queries and 8,000 gallery entries.
The released WeMM-Embedding-2B model is evaluated
zero-shot with instruction and the uncropped current frame as the query,
instruction and the 16-frame future clip as the candidate, and L2-normalized
2,048-dimensional outputs. Each external baseline is evaluated through its
supported input interface. The controlled DROID study isolates modality choices
with matched training data, backbone architecture, and optimization budget.

\paragraph{Frozen Cosmos3-Edge feature baseline.}
The frozen Cosmos3-Edge control uses the same query and target fields and native
action projectors as \embeddingmodelname{}. We extract dense action-token
features after the final RMSNorm of layer 28, before the action-output projection.
Temporal averaging and L2 normalization produce a 2,048-dimensional
VLA-conditioned backbone readout.
\ifdefined\TECHREPORTPROMOTERETRIEVALEXAMPLE
\else
\subsection{Benchmark Retrieval Examples}
\label{sec:benchmark-retrieval-examples}

Figure~\ref{fig:pai-retrieval-diagnostics} contrasts a paired-clip hit with a
temporal near-miss. The latter retrieves a nearby window from the correct
episode rather than the paired continuation, illustrating why episode matching
and clip localization require separate metrics.

\begin{figure}[!ht]
  \submissionvspace{-3pt}
  \centering
  \includegraphics[width=\linewidth]{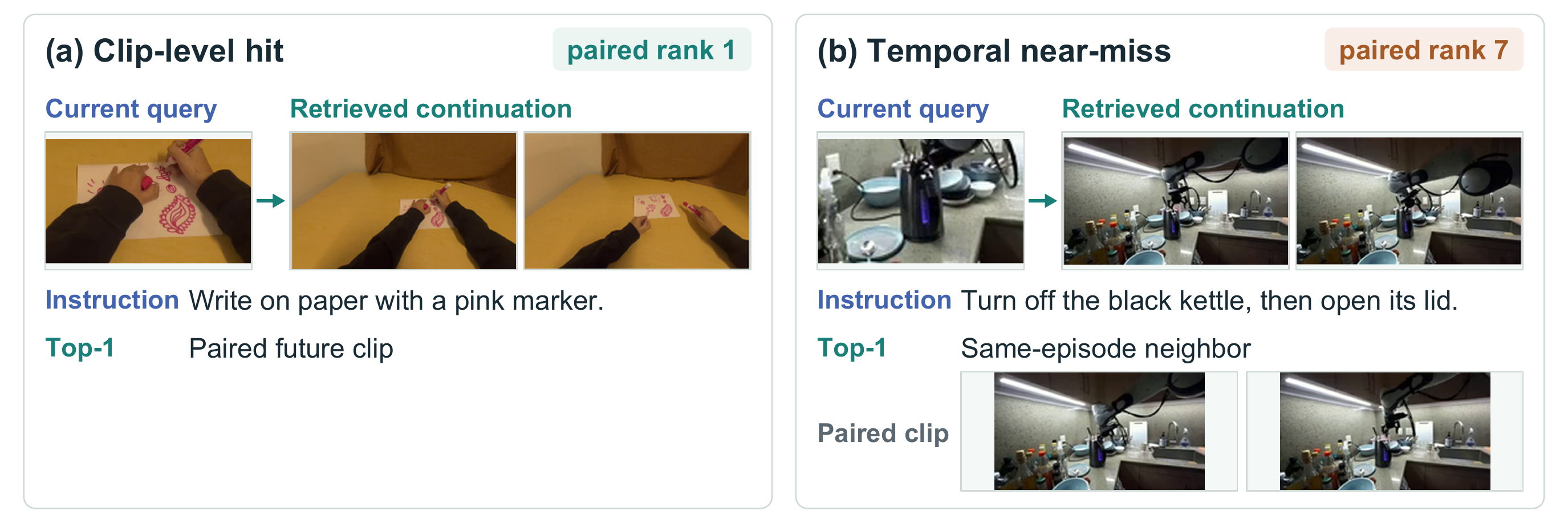}
  \ifdefined\EXPANDEDREPORT
    \vspace{-18pt}
  \else
    \vspace{-4pt}
  \fi
  \caption{\textbf{Selected retrieval diagnostics.}
  (a) The paired future clip ranks first. (b) A nearby clip from the same episode
  ranks first, while the paired clip ranks seventh. Both top-1 returns are
  episode-level matches; only (a) is also a clip-level match.}
  \label{fig:pai-retrieval-diagnostics}
  \submissionvspace{-4pt}
\end{figure}
\fi
\FloatBarrier
\ifdefined\EXPANDEDREPORT
\Needspace{0.48\textheight}
\fi
\subsection{Per-Domain Results}
\label{sec:benchmark-per-domain-results}

\ifdefined\EXPANDEDREPORT
Tables~\ref{tab:roborag-bench-domain-summary} and~\ref{tab:vlm2vec-v2-domain}
\else
Tables~\ref{tab:roborag-bench-domain-summary}--\ref{tab:wemm-embedding-2b-domain}
\fi
report retrieval on the shared 8,000-item gallery. Recall and MRR are percentages;
lower ranks are better. \emph{Average} weights sources equally. Episode-level
metrics first average within episodes, then over 800 episodes per source.

For \embeddingmodelname{}, EgoDex episode-level R@1 reaches 87.06\%, while
clip-level R@1 is 36.70\%; the corresponding DROID scores are 99.88\% and
74.10\%. The larger gap on EgoDex shows that identifying the episode often
leaves the paired window unresolved, as in the near-miss in
Figure~\ref{fig:pai-retrieval-diagnostics}. VLM2Vec-V2 and WeMM-Embedding-2B both perform best on DROID and
EgoDex, with markedly lower recall on LIBERO and RoboMIND-FD.

\ifdefined\EXPANDEDREPORT
\begin{table}[H]
  \centering
  \caption{\textbf{\embeddingmodelname{} retrieval by source.}
  Recall and MRR are percentages; lower mean ranks are better. Overall median
  ranks are 2 for clip retrieval and 1 for episode retrieval.}
  \label{tab:roborag-bench-domain-summary}
  \label{tab:roborag-bench-exact-domain}
  \label{tab:roborag-bench-episode-domain}
  \label{tab:roborag-bench-rank-summary}
  \small
  \setlength{\tabcolsep}{4.2pt}
  \begin{tabular*}{\linewidth}{@{\extracolsep{\fill}}lrrrrrr@{}}
  \toprule
  \tablegroup{Domain} & \tablegroup{R@1} & \tablegroup{R@5} & \tablegroup{R@10}
    & \tablegroup{MRR} & \tablegroup{Median rank} & \tablegroup{Mean rank} \\
  \midrule
  \multicolumn{7}{@{}l}{\tablegroup{Clip-level retrieval}} \\
  \midrule
  EgoDex & 36.70 & 96.60 & 99.30 & 62.22 & 2 & 2.68 \\
  DROID & 74.10 & 99.00 & 100.00 & 85.27 & 1 & 1.41 \\
  LIBERO & 38.50 & 87.00 & 96.60 & 60.33 & 2 & 2.92 \\
  RoboMIND-FD & 40.00 & 85.00 & 91.90 & 60.39 & 2 & 3.87 \\
  \cmidrule(l){2-7}
  \tablegroup{Average} & \tablegroup{47.33} & \tablegroup{91.90} & \tablegroup{96.95}
    & \tablegroup{67.05} & -- & \tablegroup{2.72} \\
  \midrule
  \multicolumn{7}{@{}l}{\tablegroup{Episode-level retrieval}} \\
  \midrule
  EgoDex & 87.06 & 98.94 & 99.44 & 90.98 & 1 & 1.54 \\
  DROID & 99.88 & 100.00 & 100.00 & 99.94 & 1 & 1.00 \\
  LIBERO & 67.62 & 91.81 & 98.19 & 77.42 & 1 & 2.17 \\
  RoboMIND-FD & 72.50 & 88.88 & 94.62 & 79.50 & 1 & 2.80 \\
  \cmidrule(l){2-7}
  \tablegroup{Average} & \tablegroup{81.77} & \tablegroup{94.91} & \tablegroup{98.06}
    & \tablegroup{86.96} & -- & \tablegroup{1.88} \\
  \bottomrule
\end{tabular*}

\end{table}

\FloatBarrier
\begin{table}[H]
  \centering
  \caption{\textbf{Action-free VL baselines by source.}
  Recall and MRR are percentages on the shared gallery. Average weights
  sources equally; episode-level metrics first average within each episode.}
  \label{tab:vlm2vec-v2-domain}
  \label{tab:wemm-embedding-2b-domain}
  \small
  \setlength{\tabcolsep}{3.2pt}
  \begin{tabular*}{\linewidth}{@{\extracolsep{\fill}}lrrrrrrrr@{}}
\toprule
& \multicolumn{4}{c}{\tablegroup{Clip-level}} & \multicolumn{4}{c}{\tablegroup{Episode-level}} \\
\cmidrule(lr){2-5}\cmidrule(lr){6-9}
\tablegroup{Domain} & R@1 & R@5 & R@10 & MRR & R@1 & R@5 & R@10 & MRR \\
\midrule
\multicolumn{9}{@{}l}{\tablegroup{VLM2Vec-V2}} \\
\midrule
EgoDex & 32.40 & 92.00 & 99.20 & 56.90 & 78.13 & 97.81 & 99.88 & 86.04 \\
DROID & 58.70 & 97.80 & 99.90 & 75.15 & 99.19 & 100.00 & 100.00 & 99.51 \\
LIBERO & 4.70 & 16.20 & 28.10 & 12.68 & 10.75 & 30.44 & 46.25 & 22.01 \\
RoboMIND-FD & 0.40 & 1.40 & 3.00 & 1.96 & 1.19 & 3.94 & 6.94 & 3.95 \\
\tablegroup{Average} & \tablegroup{24.05} & \tablegroup{51.85} & \tablegroup{57.55} & \tablegroup{36.67} & \tablegroup{47.31} & \tablegroup{58.05} & \tablegroup{63.27} & \tablegroup{52.88} \\
\midrule
\multicolumn{9}{@{}l}{\tablegroup{WeMM-Embedding-2B}} \\
\midrule
EgoDex & 30.80 & 90.20 & 98.50 & 54.54 & 70.31 & 96.25 & 99.63 & 80.22 \\
DROID & 61.10 & 98.40 & 100.00 & 77.18 & 99.69 & 100.00 & 100.00 & 99.84 \\
LIBERO & 4.30 & 18.30 & 30.30 & 13.31 & 9.00 & 28.56 & 44.31 & 20.05 \\
RoboMIND-FD & 0.30 & 1.70 & 3.80 & 2.09 & 1.25 & 4.44 & 8.75 & 4.20 \\
\tablegroup{Average} & \tablegroup{24.13} & \tablegroup{52.15} & \tablegroup{58.15} & \tablegroup{36.78} & \tablegroup{45.06} & \tablegroup{57.31} & \tablegroup{63.17} & \tablegroup{51.08} \\
\bottomrule
\end{tabular*}

\end{table}
\else
\begin{table}[!ht]
  \submissionvspace{-3pt}
  \centering
  \caption{\textbf{\embeddingmodelname{} retrieval by source.}
  Overall median ranks: clip 2, episode 1.}
  \label{tab:roborag-bench-domain-summary}
  \label{tab:roborag-bench-exact-domain}
  \label{tab:roborag-bench-episode-domain}
  \label{tab:roborag-bench-rank-summary}
  \small
  \setlength{\tabcolsep}{4.2pt}
  \begin{tabular*}{\linewidth}{@{\extracolsep{\fill}}lrrrrrr@{}}
  \toprule
  \tablegroup{Domain} & \tablegroup{R@1} & \tablegroup{R@5} & \tablegroup{R@10}
    & \tablegroup{MRR} & \tablegroup{Median rank} & \tablegroup{Mean rank} \\
  \midrule
  \multicolumn{7}{@{}l}{\tablegroup{Clip-level retrieval}} \\
  \midrule
  EgoDex & 36.70 & 96.60 & 99.30 & 62.22 & 2 & 2.68 \\
  DROID & 74.10 & 99.00 & 100.00 & 85.27 & 1 & 1.41 \\
  LIBERO & 38.50 & 87.00 & 96.60 & 60.33 & 2 & 2.92 \\
  RoboMIND-FD & 40.00 & 85.00 & 91.90 & 60.39 & 2 & 3.87 \\
  \cmidrule(l){2-7}
  \tablegroup{Average} & \tablegroup{47.33} & \tablegroup{91.90} & \tablegroup{96.95}
    & \tablegroup{67.05} & -- & \tablegroup{2.72} \\
  \midrule
  \multicolumn{7}{@{}l}{\tablegroup{Episode-level retrieval}} \\
  \midrule
  EgoDex & 87.06 & 98.94 & 99.44 & 90.98 & 1 & 1.54 \\
  DROID & 99.88 & 100.00 & 100.00 & 99.94 & 1 & 1.00 \\
  LIBERO & 67.62 & 91.81 & 98.19 & 77.42 & 1 & 2.17 \\
  RoboMIND-FD & 72.50 & 88.88 & 94.62 & 79.50 & 1 & 2.80 \\
  \cmidrule(l){2-7}
  \tablegroup{Average} & \tablegroup{81.77} & \tablegroup{94.91} & \tablegroup{98.06}
    & \tablegroup{86.96} & -- & \tablegroup{1.88} \\
  \bottomrule
\end{tabular*}

  \par\vspace{-2pt}
  \normalsize
  \centering
  \caption{\textbf{VLM2Vec-V2 retrieval by source.}
  Overall median ranks: clip 5, episode 2.}
  \label{tab:vlm2vec-v2-domain}
  \small
  \setlength{\tabcolsep}{4.5pt}
  \relax
  \par\vspace{-2pt}
  \normalsize
  \centering
  \caption{\textbf{WeMM-Embedding-2B retrieval by source.}
  Overall median ranks: clip 5, episode 3.}
  \label{tab:wemm-embedding-2b-domain}
  \small
  \setlength{\tabcolsep}{4.5pt}
  \relax
  \submissionvspace{-4pt}
\end{table}
\fi
\FloatBarrier
\subsection{Candidate Language and Gallery Controls}
\label{sec:benchmark-validity-audit}

Two evaluation-time controls on \benchmarkname{} test reliance on candidate
language and discrimination among clips with the same instruction.
Queries and targets come from held-out episodes excluded
from training, but retain within-episode visual continuity.

Both controls use the same fixed \embeddingmodelname{} model.
\emph{Candidate text masking} replaces each gallery instruction with an empty
prompt before multimodal fusion, then recomputes embeddings while preserving
candidate video and action. \emph{Same-instruction gallery} restricts each query
to candidates with the same instruction, testing within-instruction ranking.
The original protocol and candidate-text masking both retain the
full shared 8,000-item cross-source gallery.

\begin{table*}[!ht]
  \submissionvspace{-3pt}
  \centering
  \caption{\textbf{Aggregate benchmark-validity controls.} Candidate-text
  masking and same-instruction retrieval use a fixed model. Recall and MRR
  (\%) are averaged equally across four sources.}
  \label{tab:benchmark-validity-summary}
  \begingroup
\footnotesize
\setlength{\tabcolsep}{4.2pt}
\renewcommand{\arraystretch}{1.04}
\begin{tabular*}{\linewidth}{@{\extracolsep{\fill}}lrrrrrrrr@{}}
\toprule
& \multicolumn{4}{c}{\tablegroup{Clip-level}} & \multicolumn{4}{c}{\tablegroup{Episode-level}} \\
\cmidrule(lr){2-5}\cmidrule(lr){6-9}
\tablegroup{Setting} & R@1 & R@5 & R@10 & MRR & R@1 & R@5 & R@10 & MRR \\
\midrule
Original & 47.33 & 91.90 & 96.95 & 67.05 & 81.77 & 94.91 & 98.06 & 86.96 \\
Candidate text masked & 12.78 & 33.65 & 43.10 & 23.06 & 23.17 & 36.83 & 46.69 & 30.32 \\
Same-instruction gallery & 47.62 & 92.45 & 97.25 & 67.46 & 82.53 & 95.23 & 98.20 & 87.55 \\
\bottomrule
\end{tabular*}
\endgroup

  \submissionvspace{-4pt}
\end{table*}

Candidate text masking lowers clip-level R@1 from 47.33\% to 12.78\% and
episode-level R@1 from 81.77\% to 23.17\%. The model relies on candidate
language for current-to-future alignment: masked and original candidate
embeddings have a mean cosine similarity of 0.506.

Restricting the gallery to the same instruction yields 47.62\% clip-level R@1,
approximately 0.3 percentage points above the original protocol after removing
different-instruction distractors. The retained candidates share the instruction,
so ranking them depends on their visual and action content. Nearby same-episode
distractors further require precise localization of the paired future clip.

\begin{table*}[!ht]
  \submissionvspace{-3pt}
  \centering
  \caption{\textbf{Same-instruction gallery difficulty.} Candidate counts and
  clip-level recall (\%) within query-specific galleries. Random recall uses
  the same restricted candidates. The final row averages source means and
  recalls equally; its candidate median is the cross-source aggregate.
  RoboMIND-FD denotes RoboMIND--Franka-dual.}
  \label{tab:same-instruction-difficulty}
  \begingroup
\footnotesize
\setlength{\tabcolsep}{5.0pt}
\renewcommand{\arraystretch}{1.04}
\begin{tabular*}{\linewidth}{@{\extracolsep{\fill}}lrrrrrr@{}}
\toprule
& \multicolumn{2}{c}{\tablegroup{Candidates}} & \multicolumn{2}{c}{\tablegroup{R@1}} & \multicolumn{2}{c}{\tablegroup{R@5}} \\
\cmidrule(lr){2-3}\cmidrule(lr){4-5}\cmidrule(lr){6-7}
\tablegroup{Domain} & Mean & Median & Model & Random & Model & Random \\
\midrule
EgoDex & 4.06 & 4 & \tablebest{37.60} & 29.76 & \tablebest{98.40} & 95.09 \\
DROID & 2.71 & 2 & \tablebest{74.20} & 54.64 & \tablebest{99.20} & 97.46 \\
LIBERO & 54.51 & 50 & \tablebest{38.60} & 1.90 & \tablebest{87.00} & 9.50 \\
RoboMIND-FD & 473.59 & 342 & \tablebest{40.10} & 0.24 & \tablebest{85.20} & 1.19 \\
\midrule
\tablegroup{All domains} & 133.72 & 25 & \tablebest{47.62} & 21.63 & \tablebest{92.45} & 50.81 \\
\bottomrule
\end{tabular*}
\endgroup

  \submissionvspace{-6pt}
\end{table*}

Same-instruction pools are small in EgoDex and DROID, so R@5 approaches
saturation. LIBERO and RoboMIND-FD have larger pools, averaging 54.51 and
473.59 candidates. Their R@1 scores of 38.60\% and 40.10\% exceed the
corresponding random rates of 1.90\% and 0.24\%.

\ifdefined\EXPANDEDREPORT
\begin{table}[!ht]
  \centering
  \caption{\textbf{Benchmark-validity controls by source.}
  Recall and MRR are percentages. Candidate-text masking preserves video and
  action; same-instruction retrieval restricts each query's gallery.
  Original-protocol results are in Table~\ref{tab:roborag-bench-domain-summary}.
  Episode-level metrics use the within-episode and per-source reducer from
  Section~\ref{sec:method-benchmark}.}
  \label{tab:benchmark-validity-exact-by-domain}
  \label{tab:benchmark-validity-episode-by-domain}
  \small
  \setlength{\tabcolsep}{3.2pt}
  \begin{tabular*}{\linewidth}{@{\extracolsep{\fill}}lrrrrrrrr@{}}
\toprule
& \multicolumn{4}{c}{\tablegroup{Clip-level}} & \multicolumn{4}{c}{\tablegroup{Episode-level}} \\
\cmidrule(lr){2-5}\cmidrule(lr){6-9}
\tablegroup{Domain} & R@1 & R@5 & R@10 & MRR & R@1 & R@5 & R@10 & MRR \\
\midrule
\multicolumn{9}{@{}l}{\tablegroup{Candidate text masked}} \\
\midrule
EgoDex & 4.00 & 14.30 & 21.40 & 9.73 & 9.00 & 16.81 & 24.75 & 13.62 \\
DROID & 4.40 & 13.30 & 21.20 & 9.69 & 6.50 & 16.81 & 25.25 & 12.57 \\
LIBERO & 13.40 & 36.40 & 49.30 & 25.03 & 23.19 & 39.56 & 53.87 & 32.09 \\
RoboMIND-FD & 29.30 & 70.60 & 80.50 & 47.77 & 54.00 & 74.12 & 82.88 & 63.01 \\
\midrule
\multicolumn{9}{@{}l}{\tablegroup{Same-instruction gallery}} \\
\midrule
EgoDex & 37.60 & 98.40 & 100.00 & 63.59 & 89.69 & 100.00 & 100.00 & 92.99 \\
DROID & 74.20 & 99.20 & 100.00 & 85.34 & 100.00 & 100.00 & 100.00 & 100.00 \\
LIBERO & 38.60 & 87.00 & 96.80 & 60.41 & 67.69 & 91.81 & 98.19 & 77.47 \\
RoboMIND-FD & 40.10 & 85.20 & 92.20 & 60.51 & 72.75 & 89.12 & 94.62 & 79.73 \\
\bottomrule
\end{tabular*}

\end{table}
\else
\begin{table}[!ht]
  \submissionvspace{-3pt}
  \centering
  \caption{\textbf{Clip-level validity controls by source.} Candidate text
  masking leaves video and action unchanged; same-instruction galleries contain
  candidates sharing the query instruction. Recall and MRR are percentages;
  lower ranks are better.}
  \label{tab:benchmark-validity-exact-by-domain}
  \relax
  \submissionvspace{-4pt}
\end{table}

\begin{table}[!ht]
  \submissionvspace{-3pt}
  \centering
  \caption{\textbf{Episode-level validity controls by source.} Recall and MRR (\%)
  are averaged over queries within each episode, then equally over the 800
  episodes per source (Section~\ref{sec:method-benchmark}).}
  \label{tab:benchmark-validity-episode-by-domain}
  \relax
  \submissionvspace{-4pt}
\end{table}
\fi
\FloatBarrier
\ifdefined\TECHREPORTAPPENDIX
\subsection{Controlled \embeddingmodelname{} Ablations}
\label{sec:analysis-ablation}
\label{sec:exp-ablations}

Retraining tests how well each input set supports retrieval; query masking tests
a fixed model's reliance on its inputs. Retraining uses matched DROID splits
and budgets, with clip-level metrics (\%).
Tables~\ref{tab:pai-frozen-readout} and~\ref{tab:roborag-bench-modality-availability}
report four-source averages for frozen-feature adaptation and final-model masking.

\ifdefined\EXPANDEDREPORT

\Needspace{13\baselineskip}
\subsubsection{Retrieval Adaptation}

\begin{table}[H]
  \centering
  \caption{\textbf{Retrieval adaptation.} Four-source average retrieval
  (\%) on the shared \benchmarkname{} gallery. The frozen readout and
  \embeddingmodelname{} use the same VLA query/target fields.}
  \label{tab:pai-frozen-readout}
  \begingroup
\footnotesize
\setlength{\tabcolsep}{3pt}
\begin{tabular*}{\linewidth}{@{\extracolsep{\fill}}lrr@{}}
\toprule
\tablegroup{Setting} & \tablegroup{Clip R@1} & \tablegroup{Episode R@1} \\
\midrule
\ifdefined\EXPANDEDREPORT
Cosmos3-Edge 4B (frozen readout) & 0.65 & 1.17 \\
\else
Frozen readout & 0.65 & 1.17 \\
\fi
\tablegroup{\embeddingmodelname{}}
  & \tablebest{47.33} & \tablebest{81.77} \\
\bottomrule
\end{tabular*}
\endgroup

\end{table}

Frozen Cosmos3-Edge features provide 0.65\% clip-level R@1 and 1.17\%
episode-level R@1. Retrieval training raises these scores to 47.33\% and
81.77\%, respectively. The pretrained features contain all three modalities,
but retrieval training is needed to compare present states with recorded
continuations effectively.

\FloatBarrier
\Needspace{12\baselineskip}
\subsubsection{Trainable Scope}

\begin{table}[H]
  \centering
  \caption{\textbf{Trainable scope.} Clip-level retrieval on the controlled
  DROID split (\%). Both rows use a learned retrieval head; the second also
  updates the backbone.}
  \label{tab:droid-trainable-scope}
  \begingroup
\footnotesize
\setlength{\tabcolsep}{3pt}
\begin{tabular*}{\linewidth}{@{\extracolsep{\fill}}lrr@{}}
\toprule
\tablegroup{Trainable scope} & \tablegroup{R@1} & \tablegroup{MRR} \\
\midrule
Head only & 13.8 & 24.00 \\
\tablegroup{Backbone + head} & \tablebest{78.7} & \tablebest{88.00} \\
\bottomrule
\end{tabular*}
\endgroup

\end{table}

Training a retrieval head over the frozen backbone reaches 13.8\% R@1,
whereas training the backbone and head reaches 78.7\%. Under matched DROID training,
adapting the shared representation is much more effective than learning a
readout from fixed features.

\FloatBarrier
\Needspace{15\baselineskip}
\subsubsection{Objective Groups}

\begin{table}[H]
  \centering
  \caption{\textbf{Objective groups.} Clip-level retrieval on the controlled
  DROID split (\%). Each row adds an objective group from
  Equation~\ref{eq:composite-objective}.}
  \label{tab:droid-objective-groups}
  \begingroup
\footnotesize
\setlength{\tabcolsep}{3pt}
\begin{tabular*}{\linewidth}{@{\extracolsep{\fill}}lrr@{}}
\toprule
\tablegroup{Objective} (Eq.~\ref{eq:composite-objective}) & \tablegroup{R@1} & \tablegroup{MRR} \\
\midrule
$\mathcal{L}_{\mathrm{align}}$ & 75.6 & 86.20 \\
$\mathcal{L}_{\mathrm{align}}+\mathcal{L}_{\mathrm{reg}}$ & 77.0 & 87.10 \\
$\mathcal{L}_{\mathrm{align}}+\mathcal{L}_{\mathrm{ground}}+\mathcal{L}_{\mathrm{reg}}$
  & \tablebest{78.7} & \tablebest{88.00} \\
\bottomrule
\end{tabular*}
\endgroup

\end{table}

Predictive retrieval alignment alone reaches 75.6\% R@1. Adding representation
regularization raises R@1 to 77.0\%, and adding multimodal grounding and
discrimination reaches 78.7\%. R@10 is 100.0\% throughout these three runs.
Predictive alignment already places the correct clip near the top; the added
objective groups improve its rank among those closest candidates.

\FloatBarrier
\Needspace{15\baselineskip}
\subsubsection{Visual Input}

\begin{table}[H]
  \centering
  \caption{\textbf{Visual input.} Clip-level retrieval on the controlled
  DROID split (\%). All rows are retrained for their stated visual interface.}
  \label{tab:droid-visual-input}
  \begingroup
\footnotesize
\setlength{\tabcolsep}{3pt}
\begin{tabular*}{\linewidth}{@{\extracolsep{\fill}}lrr@{}}
\toprule
\tablegroup{Visual input} & \tablegroup{R@1} & \tablegroup{MRR} \\
\midrule
w/o vision (retrained) & 76.8 & 86.69 \\
Full frame & 77.9 & 87.40 \\
\tablegroup{SAM3 object zoom} & \tablebest{78.7} & \tablebest{88.00} \\
\bottomrule
\end{tabular*}
\endgroup

\end{table}

The no-vision, full-frame, and SAM3 object-zoom settings reach 76.8\%, 77.9\%,
and 78.7\% R@1, respectively. Their MRR values are 86.69\%, 87.40\%, and
88.00\%. SAM3 object zoom gives the highest R@1 and MRR
among these matched DROID retrieval variants, with a modest 0.8-point R@1 gain
over the full frame. Cropping to the interaction region gives a small
improvement in clip localization.

\FloatBarrier
\Needspace{15\baselineskip}
\subsubsection{Training-Time Modalities}

\begin{table}[H]
  \centering
  \caption{\textbf{Training-time modalities.} Matched DROID retraining after
  removing one modality (\%). V/L/A denote vision, language, and action.}
  \label{tab:droid-training-modalities}
  \begingroup
\small
\setlength{\tabcolsep}{2pt}
\begin{tabular*}{\linewidth}{@{\extracolsep{\fill}}lcrr@{}}
\toprule
\tablegroup{Setting} & \tablegroup{Input} & \shortstack{\tablegroup{Clip}\\\tablegroup{R@1}} & \shortstack{$\Delta$ clip\\R@1} \\
\midrule
\tablegroup{Full VLA} & VLA & \tablebest{78.7} & --- \\
No action & VL & 59.8 & $-18.9$ \\
No vision & LA & 76.8 & $-1.9$ \\
No language & VA & 20.2 & $-58.5$ \\
\bottomrule
\end{tabular*}
\endgroup

\end{table}

Removing action lowers R@1 from 78.7\% to 59.8\%, a drop of 18.9 percentage
points. Removing language has the largest effect, while retraining without
vision lowers R@1 by 1.9 points. Native action supplies
retrieval signal beyond vision and language in this controlled DROID study.

\FloatBarrier
\Needspace{22\baselineskip}
\subsubsection{Inference-Time Modality Availability}

\begin{table}[H]
  \centering
  \caption{\textbf{Inference-time modality availability.} Four-source average
  retrieval (\%) with the final checkpoint and gallery fixed. Only the query
  inputs are changed. V/L/A denote vision/language/action; $\Delta$ gives the
  percentage-point change in clip-level R@1 from Full VLA.}
  \label{tab:roborag-bench-modality-availability}
  \label{tab:single-modality-queries}
  \small
  \begin{tabular*}{\linewidth}{@{\extracolsep{\fill}}lrrr@{}}
\toprule
\tablegroup{Query input} & \tablegroup{Clip R@1} & \tablegroup{Episode R@1} & $\Delta$ clip R@1 \\
\midrule
\tablegroup{Full VLA} & \tablebest{47.33} & \tablebest{81.77} & --- \\
VL (no action) & 20.08 & 37.36 & $-27.25$ \\
LA (no vision) & 23.00 & 39.50 & $-24.33$ \\
VA (no language) & 7.13 & 14.73 & $-40.20$ \\
L (text only) & 16.23 & 31.09 & $-31.10$ \\
V (image only) & 1.68 & 3.66 & $-45.65$ \\
A (action only) & 0.53 & 1.58 & $-46.80$ \\
\bottomrule
\end{tabular*}

\end{table}

Withholding query action, image, or text lowers clip-level R@1 by 27.25,
24.33, or 40.20 percentage points, respectively. Full VLA ranks highest at
both clip and episode levels. The final model uses all three query modalities;
the drops also reflect the input distribution shift caused by masking.
Text alone reaches 16.23\% clip-level R@1; image-only and action-only queries
yield 1.68\% and 0.53\%. No single query modality reproduces the fused VLA
representation under this fixed-model protocol.

\else

\begin{table}[!ht]
  \submissionvspace{-9pt}
  \begin{minipage}[t]{0.48\linewidth}
    \centering
    \caption{\textbf{Retrieval adaptation.}}
    \label{tab:pai-frozen-readout}
    \begingroup
\footnotesize
\setlength{\tabcolsep}{3pt}
\begin{tabular*}{\linewidth}{@{\extracolsep{\fill}}lrr@{}}
\toprule
\tablegroup{Setting} & \tablegroup{Clip R@1} & \tablegroup{Episode R@1} \\
\midrule
\ifdefined\EXPANDEDREPORT
Cosmos3-Edge 4B (frozen readout) & 0.65 & 1.17 \\
\else
Frozen readout & 0.65 & 1.17 \\
\fi
\tablegroup{\embeddingmodelname{}}
  & \tablebest{47.33} & \tablebest{81.77} \\
\bottomrule
\end{tabular*}
\endgroup

  \end{minipage}\hfill
  \begin{minipage}[t]{0.48\linewidth}
    \centering
    \caption{\textbf{Trainable scope.}}
    \label{tab:droid-trainable-scope}
    \begingroup
\footnotesize
\setlength{\tabcolsep}{3pt}
\begin{tabular*}{\linewidth}{@{\extracolsep{\fill}}lrr@{}}
\toprule
\tablegroup{Trainable scope} & \tablegroup{R@1} & \tablegroup{MRR} \\
\midrule
Head only & 13.8 & 24.00 \\
\tablegroup{Backbone + head} & \tablebest{78.7} & \tablebest{88.00} \\
\bottomrule
\end{tabular*}
\endgroup

  \end{minipage}
  \submissionvspace{-10pt}
\end{table}

\submissionvspace{-3pt}
\paragraph{Retrieval adaptation and trainable scope.}
Across four sources, training raises clip R@1 from 0.65\% with frozen features
to 47.33\% (Table~\ref{tab:pai-frozen-readout}). On matched DROID, head-only training
reaches 13.8\%, versus 78.7\% with backbone adaptation (Table~\ref{tab:droid-trainable-scope}).
Current-to-future matching benefits from adapting the backbone itself,
rather than learning only a readout from its pretrained features.

\begin{table}[!ht]
  \submissionvspace{-9pt}
  \begin{minipage}[t]{0.48\linewidth}
    \centering
    \caption{\textbf{Objective groups.}}
    \label{tab:droid-objective-groups}
    \begingroup
\footnotesize
\setlength{\tabcolsep}{3pt}
\begin{tabular*}{\linewidth}{@{\extracolsep{\fill}}lrr@{}}
\toprule
\tablegroup{Objective} (Eq.~\ref{eq:composite-objective}) & \tablegroup{R@1} & \tablegroup{MRR} \\
\midrule
$\mathcal{L}_{\mathrm{align}}$ & 75.6 & 86.20 \\
$\mathcal{L}_{\mathrm{align}}+\mathcal{L}_{\mathrm{reg}}$ & 77.0 & 87.10 \\
$\mathcal{L}_{\mathrm{align}}+\mathcal{L}_{\mathrm{ground}}+\mathcal{L}_{\mathrm{reg}}$
  & \tablebest{78.7} & \tablebest{88.00} \\
\bottomrule
\end{tabular*}
\endgroup

  \end{minipage}\hfill
  \begin{minipage}[t]{0.48\linewidth}
    \centering
    \caption{\textbf{Visual input.}}
    \label{tab:droid-visual-input}
    \begingroup
\footnotesize
\setlength{\tabcolsep}{3pt}
\begin{tabular*}{\linewidth}{@{\extracolsep{\fill}}lrr@{}}
\toprule
\tablegroup{Visual input} & \tablegroup{R@1} & \tablegroup{MRR} \\
\midrule
w/o vision (retrained) & 76.8 & 86.69 \\
Full frame & 77.9 & 87.40 \\
\tablegroup{SAM3 object zoom} & \tablebest{78.7} & \tablebest{88.00} \\
\bottomrule
\end{tabular*}
\endgroup

  \end{minipage}
  \submissionvspace{-10pt}
\end{table}

\submissionvspace{-3pt}
\paragraph{Objective groups.}
Predictive alignment reaches 75.6\% R@1; adding $\mathcal{L}_{\mathrm{reg}}$ gives
77.0\%, and $\mathcal{L}_{\mathrm{ground}}$ brings it to 78.7\%
(Table~\ref{tab:droid-objective-groups}). R@10 remains 100.0\% while top-rank
ordering improves.
\submissionvspace{-3pt}
\paragraph{Visual input.}
SAM3 interaction crops reach 78.7\% R@1, versus 77.9\% for full frames and
76.8\% without vision, and yield the highest MRR (Table~\ref{tab:droid-visual-input}).

\begin{table}[!ht]
  \submissionvspace{-9pt}
  \begin{minipage}[t]{0.48\linewidth}
    \centering
    \caption{\strut\textbf{Training-time modalities.}}
    \label{tab:droid-training-modalities}
    \begingroup
\small
\setlength{\tabcolsep}{2pt}
\begin{tabular*}{\linewidth}{@{\extracolsep{\fill}}lcrr@{}}
\toprule
\tablegroup{Setting} & \tablegroup{Input} & \shortstack{\tablegroup{Clip}\\\tablegroup{R@1}} & \shortstack{$\Delta$ clip\\R@1} \\
\midrule
\tablegroup{Full VLA} & VLA & \tablebest{78.7} & --- \\
No action & VL & 59.8 & $-18.9$ \\
No vision & LA & 76.8 & $-1.9$ \\
No language & VA & 20.2 & $-58.5$ \\
\bottomrule
\end{tabular*}
\endgroup

  \end{minipage}\hfill
  \begin{minipage}[t]{0.48\linewidth}
    \centering
    \caption{\strut\textbf{Inference-time modalities.}}
    \label{tab:roborag-bench-modality-availability}
    \relax
  \end{minipage}
  \par\vspace{3pt}
  {\footnotesize Left: matched DROID retraining; right: four-source query masking
  with the checkpoint and gallery fixed. V/L/A denote vision/language/action;
  $\Delta$ is the percentage-point change from Full VLA.\par}
  \submissionvspace{-10pt}
\end{table}

\submissionvspace{-3pt}
\paragraph{Training-time modalities.}
Removing action during retraining lowers R@1 by 18.9 percentage points
(Table~\ref{tab:droid-training-modalities}). Native action contributes retrieval
information beyond vision and language. Removing language has the largest effect;
removing vision lowers R@1 by 1.9 points.

\submissionvspace{-3pt}
\paragraph{Inference-time modalities.}
With the final checkpoint and gallery fixed, withholding query action, image, or text lowers
clip R@1 by 27.25, 24.33, or 40.20 percentage points (Table~\ref{tab:roborag-bench-modality-availability}).
Full VLA leads at both clip and episode levels; the trained model uses
all three query modalities.

\fi

\newcommand{\modalitycomparisonsummary}{%
Removing action hurts retrieval more than removing vision in both studies,
whether the model is retrained or its query inputs are masked
(\detailref{sec:droid-query-masking-details}).
The DROID no-vision model reaches 76.8\% R@1 after retraining, but masking
vision in the full model gives 60.9\%. Retraining permits adaptation to the
remaining inputs; masking tests the learned fusion under an input shift.
The small retraining gap therefore does not mean the full model ignores vision.%
}
\ifdefined\EXPANDEDREPORT\else
\modalitycomparisonsummary
\fi
  \FloatBarrier
\fi
\FloatBarrier
\ifdefined\EXPANDEDREPORT
\Needspace{22\baselineskip}
\subsection{Controlled DROID Query Masking}
\else
\subsection{Additional Query Masking}
\fi
\label{sec:droid-query-masking-details}
\label{app:ablation-studies}

\ifdefined\EXPANDEDREPORT
\else
\paragraph{Single-modality queries.}
Table~\ref{tab:single-modality-queries} extends final-model masking
(Table~\ref{tab:roborag-bench-modality-availability}) to single-modality queries,
keeping model weights and gallery embeddings fixed. Text alone reaches
16.23\% clip-level R@1; vision and action alone yield lower recall, and all three
fall below the fused VLA query.
\par\nopagebreak[4]

\begin{table}[H]
  \centering
  \caption{\textbf{Single-modality queries.} Four-source average recall (\%)
  with the model and gallery embeddings fixed. $\Delta$ gives the
  percentage-point change in clip-level R@1 from the full VLA query.}
  \label{tab:single-modality-queries}
  \small
  \relax
  \submissionvspace{-4pt}
\end{table}
\fi

\ifdefined\EXPANDEDREPORT\else
\paragraph{Controlled DROID masking.}
\fi
We mask inputs to the fixed DROID Full VLA model and compare its response with
the retrained variants in Table~\ref{tab:droid-training-modalities}. This test
uses the controlled DROID checkpoint and split, whereas
Table~\ref{tab:roborag-bench-modality-availability} uses the final checkpoint
across four sources.
\par\nopagebreak[4]

\begin{table}[H]
  \submissionvspace{-3pt}
  \centering
  \caption{\textbf{DROID query-only masking.} Clip-level recall and MRR (\%)
  with the model and gallery embeddings fixed. $\Delta$ R@1 gives the
  percentage-point change from the full VLA query.}
  \label{tab:droid-query-masking}
  \small
  \begin{tabular*}{\linewidth}{@{\extracolsep{\fill}}lcrrrrr@{}}
\toprule
\tablegroup{Query setting} & \tablegroup{Input} & \tablegroup{R@1} & \tablegroup{R@5} & \tablegroup{R@10} & \tablegroup{MRR} & $\Delta$ R@1 \\
\midrule
\tablegroup{Full query}
  & VLA & \tablebest{78.7} & \tablebest{99.4} & \tablebest{100.0}
  & \tablebest{88.00} & $\mathbf{0.0}$ \\
Action masked
  & VL & 50.7 & 90.3 & 94.9 & 67.60 & $-28.0$ \\
Image masked
  & LA & 60.9 & 91.8 & 96.1 & 74.17 & $-17.8$ \\
Text masked
  & VA & 1.9 & 7.2 & 12.0 & 5.41 & $-76.8$ \\
\bottomrule
\end{tabular*}

  \submissionvspace{-4pt}
\end{table}

\ifdefined\EXPANDEDREPORT
\textbf{Conclusion.} Masking action, image, or text lowers clip-level R@1 from
78.7\% to 50.7\%, 60.9\%, or 1.9\%, respectively. Removing text has the largest
effect, followed by action and image. The larger effect of removing action
than vision is shared by the retraining and masking studies. No-vision
retraining reaches 76.8\% R@1, whereas masking vision in the full model yields
60.9\%. Retraining permits adaptation to the remaining inputs; masking tests
the fixed model's reliance on its inputs, including sensitivity to distribution
shift. The small no-vision retraining gap therefore does not imply that the
full model ignores vision.
\else
Masking action, image, or text lowers clip-level R@1 from 78.7\% to 50.7\%,
60.9\%, or 1.9\%, respectively. Removing text has the largest effect, followed
by action and image. No-vision retraining reaches 76.8\% R@1, whereas masking
vision in the full model yields 60.9\%. Retraining measures performance after
adapting to the available modalities; masking measures the fixed model's
reliance on its inputs, including sensitivity to distribution shift.
\fi
\FloatBarrier
\ifdefined\EXPANDEDREPORT
  \Needspace{23\baselineskip}
\fi
\subsection{Index-Size Scaling and Retrieval Latency}
\label{sec:gallery-size-scaling}

Table~\ref{tab:gallery-size-scaling} measures query encoding and Top-10 search
over precomputed galleries on one NVIDIA A100. Timings exclude SAM3
preprocessing, reranking, policy inference, and robot execution. Throughput is
the reciprocal of total latency; GPU memory is estimated. Encoding takes
157.70 ms at every scale, while expanding the gallery from one million to
15 million entries adds 34.81 ms to search latency. Retrieval quality is evaluated
separately on the 8,000-entry benchmark.

\ifdefined\EXPANDEDREPORT
  \begin{table}[H]
\else
  \begin{table}[!ht]
\fi
  \centering
  \caption{\textbf{Retrieval scaling on one NVIDIA A100.} Measured latencies
  cover Top-10 search alone and query encoding plus search. Throughput is
  derived from the latter; GPU memory is estimated.}
  \label{tab:gallery-size-scaling}
  \begin{tabular*}{\columnwidth}{@{\extracolsep{\fill}}lrrrr@{}}
  \toprule
  \tablegroup{Gallery} & \multicolumn{1}{r}{\shortstack[r]{\tablegroup{Top-10}\\search (ms)}}
          & \multicolumn{1}{r}{\shortstack[r]{\tablegroup{Encode + retrieve}\\latency (ms)}}
          & \multicolumn{1}{r}{\shortstack[r]{\tablegroup{Single-stream}\\QPS}}
          & \multicolumn{1}{r}{\shortstack[r]{\tablegroup{Est. GPU}\\memory (GB)}} \\
  \midrule
  1M  &  2.66 & 160.36 & 6.24 & 11.29 \\
  2M  &  5.14 & 162.84 & 6.14 & 15.39 \\
  5M  & 12.60 & 170.30 & 5.87 & 27.69 \\
  10M & 25.04 & 182.74 & 5.47 & 48.19 \\
  15M & 37.47 & 195.17 & 5.12 & 68.69 \\
  \bottomrule
\end{tabular*}

\end{table}
\subsection{Qualitative Cross-Embodiment Analysis}
\label{sec:cross-embodiment-analysis}

\ifdefined\EXPANDEDREPORT
\begin{figure}[!htbp]
\else
\begin{figure}[H]
\fi
  \submissionvspace{-14pt}
  \centering
  \includegraphics[width=\linewidth]{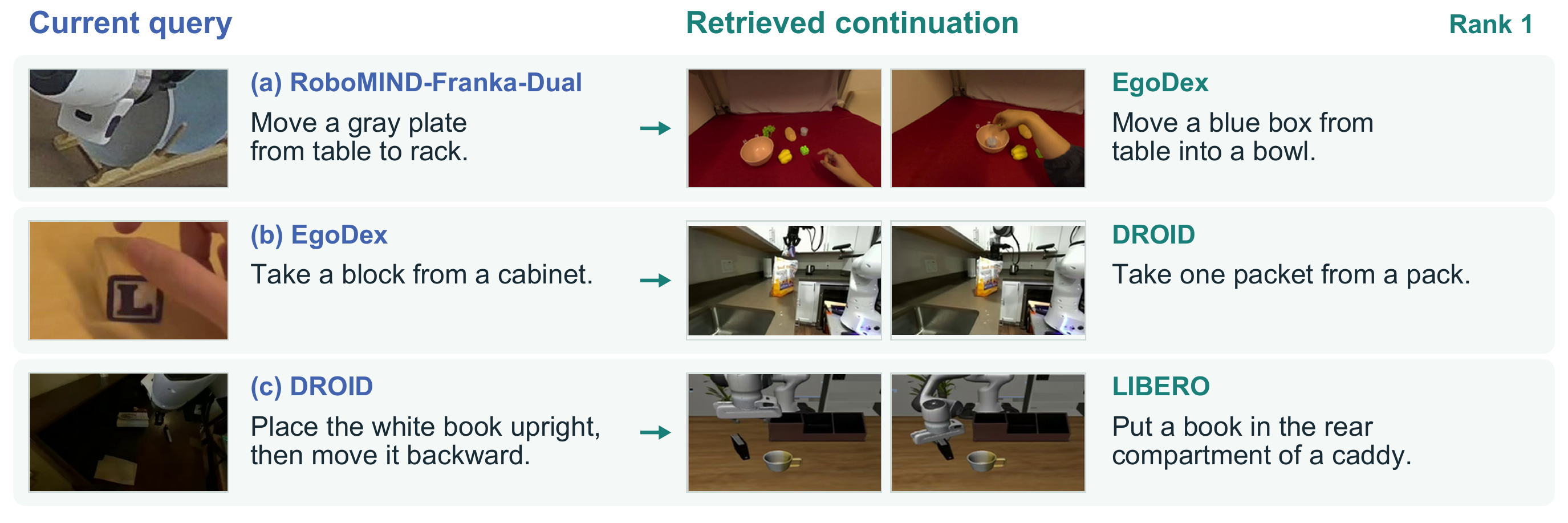}
  \par
  \ifdefined\EXPANDEDREPORT
    \vspace{-7pt}
  \else
    \vspace{-14pt}
  \fi
  \caption{\textbf{Selected qualitative evidence across embodiments.}
  Each query searches only the illustrated target source. The top-ranked clips
  share placement or removal roles with their queries.}
  \label{fig:pai-cross-embodiment-examples}
  \submissionvspace{-14pt}
\end{figure}

Figure~\ref{fig:pai-cross-embodiment-examples} presents three selected cases
from a diagnostic checkpoint. Each query searches only the illustrated target
source, excluding its own. The top-ranked clips share placement or removal
roles with their queries across human, robot, and simulated experience.
\ifdefined\EXPANDEDREPORT
The three cases provide selected qualitative evidence from a separate
diagnostic checkpoint. Aggregate cross-source accuracy, prevalence,
robustness, and source invariance are outside this analysis.
\fi
\FloatBarrier
\section{Embed2ICL Implementation and Supplementary Results}
\label{app:roborag-supplement}

This section covers DreamZero's training-free video-cache warm-up and
Fast-WAM's training-based Context Tokens, followed by integration controls.

\subsection{DreamZero-DROID Reference Warm-Up}
\label{sec:dreamzero-warmup-details}

Embed2ICL initializes frozen DreamZero-DROID's cache with a retrieved and
reranked video (Figure~\ref{fig:zero-shot-reference-warmup}). Both retrievers use
the same GPT-5.6 sol Reranking Agent. All reference conditions use four-second
videos. Each Embed2ICL reference contains three synchronized streams at 15 Hz for
four seconds (60 frames per view): 45 frames before the detected final gripper
release, the release frame, and 14 frames after it. All three streams serve
as video-only policy input. Embed2ICL performs one warm-up per applicable
task phase and issues eight reference-conditioned policy requests. Live
observations then update the rolling cache. The source episode and temporal
window remain fixed across all trials for a task, so the experiment evaluates
conditioning on a selected task reference. We evaluate the three tasks in
Table~\ref{tab:dreamzero-zero-shot-results}. No warm-up, random reference,
both retrievers' top-1, and both reranked conditions each use
100 trials per task, totaling 300 per condition. All share the released
DreamZero-DROID policy; only the test-time video reference changes.
The top-1 conditions select an episode and recut a four-second reference
from its successful stage.

\begin{figure}[!ht]
  \submissionvspace{-3pt}
  \centering
  \includegraphics[width=\linewidth,trim=0 20bp 0 0,clip]{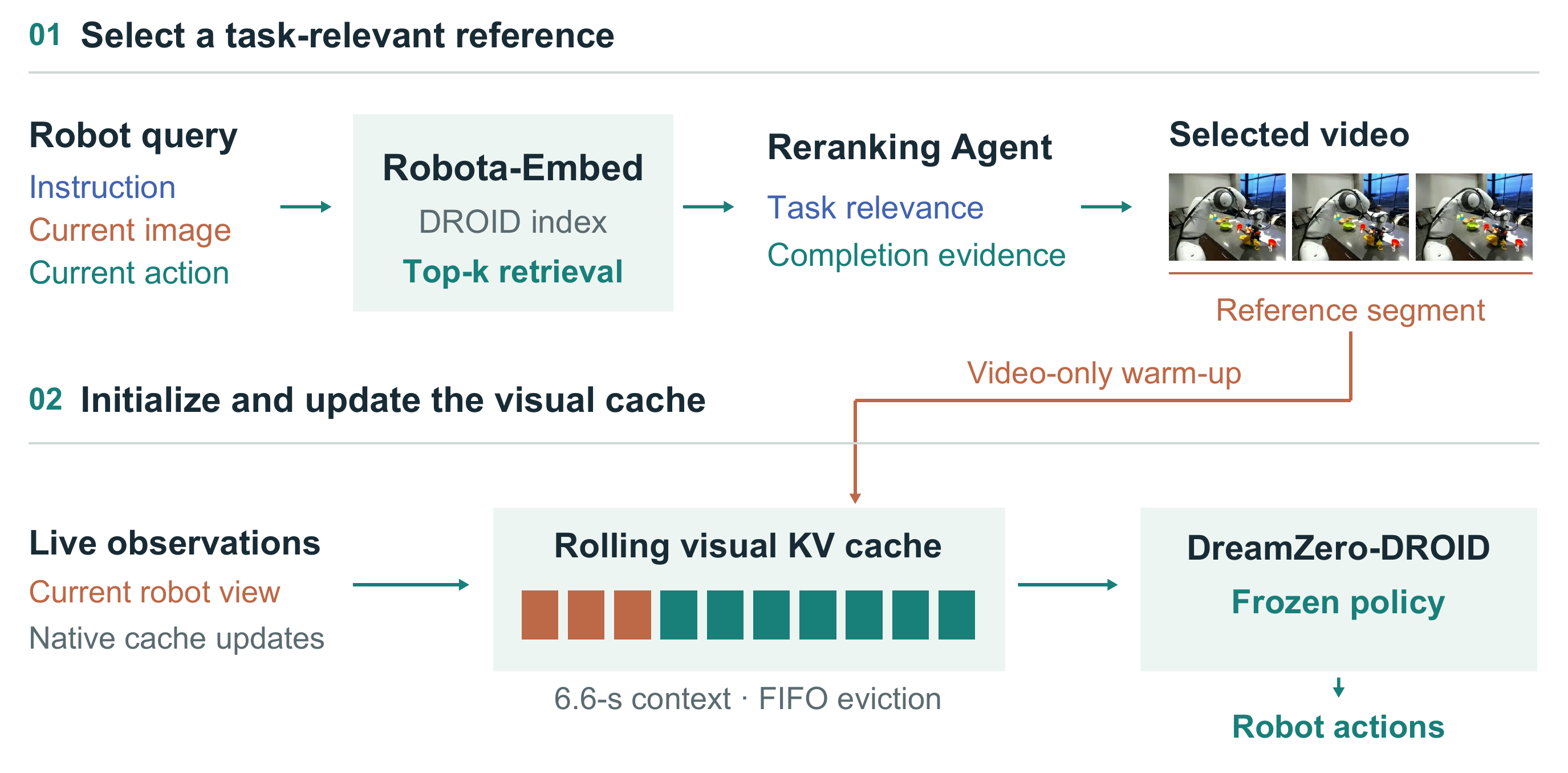}
  \par\vspace{-4pt}
  \caption{\textbf{Training-free Embed2ICL.}
  \embeddingmodelname{} retrieves candidate VLA clips, and the Reranking Agent
  selects a task-relevant reference containing completion evidence. Its video
  initializes DreamZero-DROID's visual key--value cache. The policy remains
  frozen while live observations update the cache during execution.}
  \label{fig:zero-shot-reference-warmup}
  \submissionvspace{-4pt}
\end{figure}
\ifdefined\EXPANDEDREPORT
\else
  \FloatBarrier
\fi
\subsection{DreamZero Reference and Interface Analysis}
\label{sec:dreamzero-reference-diagnostics}
\label{app:dreamzero-supplement}

Figure~\ref{fig:zero-shot-reference-clips} shows the task-specific DROID
warm-up references for full \methodname{}
(Table~\ref{tab:dreamzero-zero-shot-results}). All three preserve the
placement relation while changing the object or receptacle:
Cube$\rightarrow$Bowl uses a block-to-bowl example;
Can$\rightarrow$Mug uses a cap-to-container analogue; and
Banana$\rightarrow$Bin preserves the manipulated object but changes the receptacle.
\ifdefined\EXPANDEDREPORT
Reranking selects references using completion evidence among other criteria;
Table~\ref{tab:dreamzero-zero-shot-results} evaluates their usefulness as
policy context.
\else
\embeddingmodelname{} outperforms Omni-Embed-Nemotron-3B with both raw top-1
selection and reranking (Table~\ref{tab:dreamzero-zero-shot-results}).
Reranking improves the reported rates for both retrievers. It selects references
using completion evidence among other criteria; these comparisons measure
their usefulness as policy context.
\fi

\begin{figure}[!ht]
  \submissionvspace{-3pt}
  \centering
  \includegraphics[width=\linewidth]{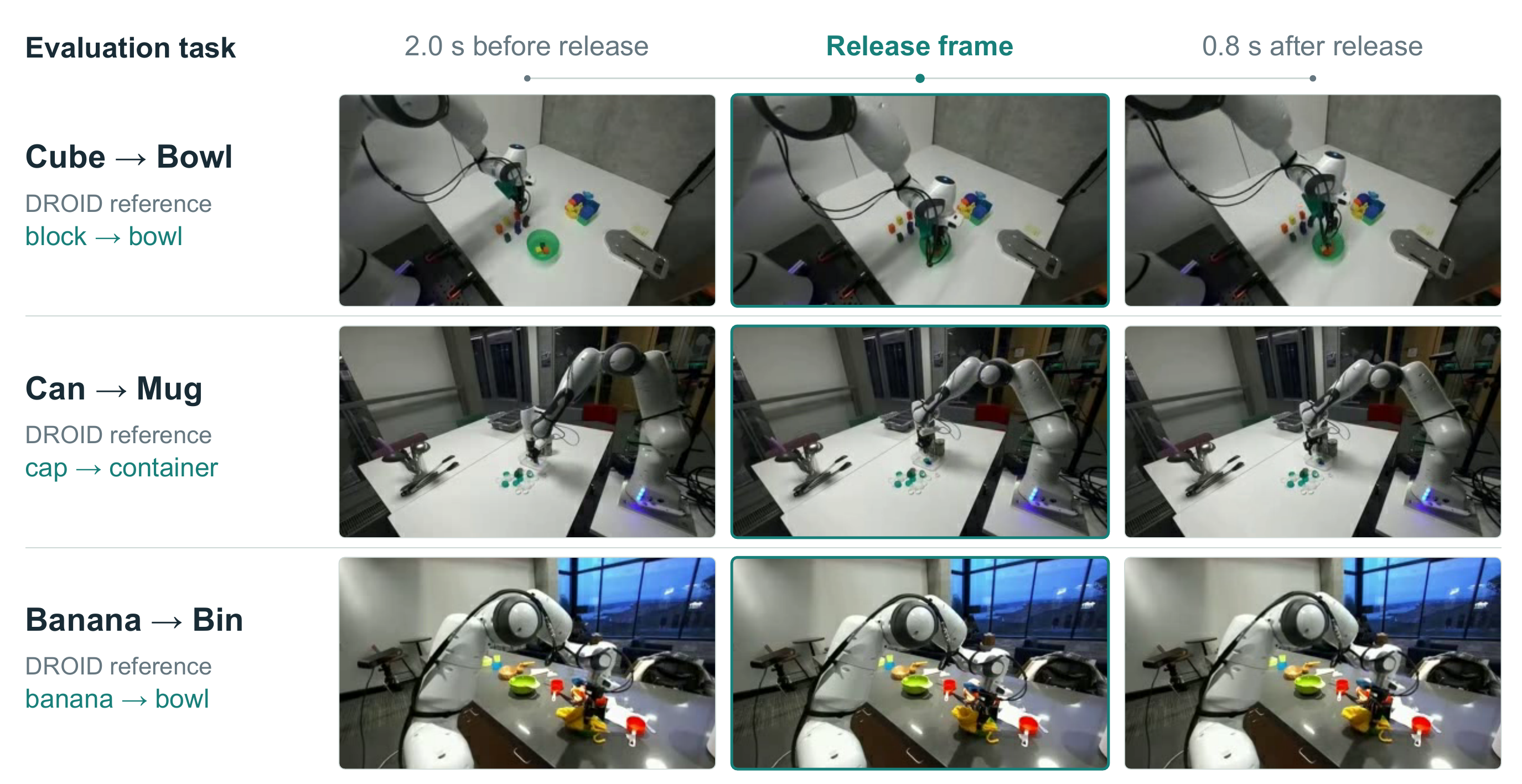}
  \caption{\textbf{DROID references for training-free Embed2ICL.}
  Each row shows one task's fixed reference from a single camera, before, at,
  and after the detected final gripper release. DreamZero-DROID receives all
  three synchronized four-second video streams.}
  \label{fig:zero-shot-reference-clips}
  \submissionvspace{-4pt}
\end{figure}

\ifdefined\EXPANDEDREPORT
\paragraph{Reference-cache compatibility.}
DreamZero-DROID~\citep{ye2026dreamzero} supports video warm-up through its
rolling 6.6-second visual KV cache and stateless VAE.
LingBot-VA~\citep{li2026lingbotva} (CauVA in the proceedings) instead requires
synchronized initialization of video/action positions and streaming VAE state.
Fast-WAM~\citep{yuan2026fastwam} resets visual KV state at each policy call;
Embed2ICL therefore supplies its trained Context-Token interface
(\detailref{sec:fastwam-injection-details}).
\else
  \begin{table}[H]
  \submissionvspace{-3pt}
  \centering
  \caption{\textbf{Reference-cache warm-up compatibility.}
  Native test-time state and compatibility with initialization from a
  retrieved video.}
  \label{tab:zero-shot-backbone-compatibility}
  \small
  \relax
  \submissionvspace{-4pt}
\end{table}
Reference warm-up depends on the policy's test-time state.
DreamZero-DROID combines a rolling visual cache with a stateless VAE.
LingBot-VA (CauVA in the proceedings) requires synchronized initialization of
video and action positions and streaming VAE state. Fast-WAM instead uses Embed2ICL's trained Context-Token
interface (\detailref{sec:fastwam-injection-details}).
\fi

\ifdefined\EXPANDEDREPORT
  \FloatBarrier
\fi
\ifdefined\TECHREPORTPROMOTEYAM
\else
\subsection{Training-Free Embed2ICL on the YAM Robot}
\label{sec:dreamzero-real-robot-details}

\paragraph{Policy and reference.}
We adapt DreamZero-AgiBot~\citep{ye2026dreamzero} to YAM for 20 epochs on
55 play trajectories, comprising approximately 30 minutes of interaction.
Both conditions share this adapted policy; Embed2ICL adds retrieved video
through visual KV-cache warm-up without further policy updates. It reuses
the banana-placement reference selected by
\embeddingmodelname{} and the Reranking Agent for the DROID-Sim
Banana-to-Bin task. This DROID reference depicts placement into a bowl;
the real-robot task is Banana-to-Bowl.

\paragraph{Evaluation.}
We evaluate each condition over 12 real-robot rollouts. Success requires
the banana to remain inside the bowl while the bowl stays upright.
We also report mean cumulative task progress on a 0--100 scale, awarding
20 points for each completed stage: (1) gripper contact with the banana;
(2) grasping and lifting it from the table; (3) transporting it above the
bowl; (4) releasing it into the bowl; and (5) stable placement with the bowl
upright. Embed2ICL raises mean progress from 21.67 to 40.00
(Table~\ref{tab:physis-icl-real-robot-results}); success increases from
0.0\% to 16.7\%.

\ifdefined\TECHREPORTPROMOTEYAM
\else
\ifdefined\EXPANDEDREPORT
\else
\begin{figure}[!ht]
  \centering
  \relax
  \par\vspace{-4pt}
  \caption{\textbf{Retrieved reference and DreamZero execution on YAM.}
  Simulation and YAM share the DROID banana-placement reference.
  Frames show a selected real-robot rollout's initial state, transport,
  bowl interaction, and final visible state. Times are relative to each
  video's start.}
  \label{fig:dreamzero-real-robot-storyboard}
  \submissionvspace{-4pt}
\end{figure}
\fi
\fi
\fi
\ifdefined\TECHREPORTPROMOTEFASTWAM
\else
\ifdefined\TECHREPORTPROMOTEFASTWAM
  \subsubsection{Fast-WAM Context Interface}
\else
  \subsection{Fast-WAM Integration Details}
\fi
\label{sec:fastwam-injection-details}

\paragraph{Context Tokens from target states.}
The target encoder jointly encodes task text, future video, and native
actions. From valid action slots after layer 28's generation RMSNorm,
we extract fused states $H_i^k\in\mathbb{R}^{T_i\times2048}$ used by the
global retrieval branch (Eq.~\ref{eq:rerum-readout}). Adaptive temporal average
pooling retains at most 32 positions; standard windows from all four
sources yield 32. Pooled states pass through shared LayerNorm,
$2048\!\rightarrow\!1024$ projection $W$, and target residual adapter $A_k$:
\begin{equation}
    \begin{aligned}
      \widetilde H_i&=\operatorname{Pool}_{32}(H_i^k),\qquad
      u_{i,j}=A_k\!\left(W\operatorname{LN}(\widetilde H_{i,j})\right),\\
      C_{i,j}&=u_{i,j}/\|u_{i,j}\|_2,\qquad
      C_i\in\mathbb{R}^{32\times1024}.
    \end{aligned}
    \label{eq:fastwam-context-tokens}
\end{equation}
Standard windows yield 32 independently L2-normalized tokens from temporally
pooled states, not copies of the global retrieval vector. Cosine search uses
$k_i$; Fast-WAM consumes $C_i$.

\paragraph{Retrieval-trained parameters.}
Context Tokens reuse the Transformer, shared projection, and endpoint adapters
learned during retrieval training. The video VAE remains frozen. Token
extraction adds no separate head or token-level alignment or reconstruction loss.

\begin{figure}[!ht]
  \submissionvspace{-3pt}
  \centering
  \includegraphics[width=\linewidth,trim=0 30bp 0 0,clip]{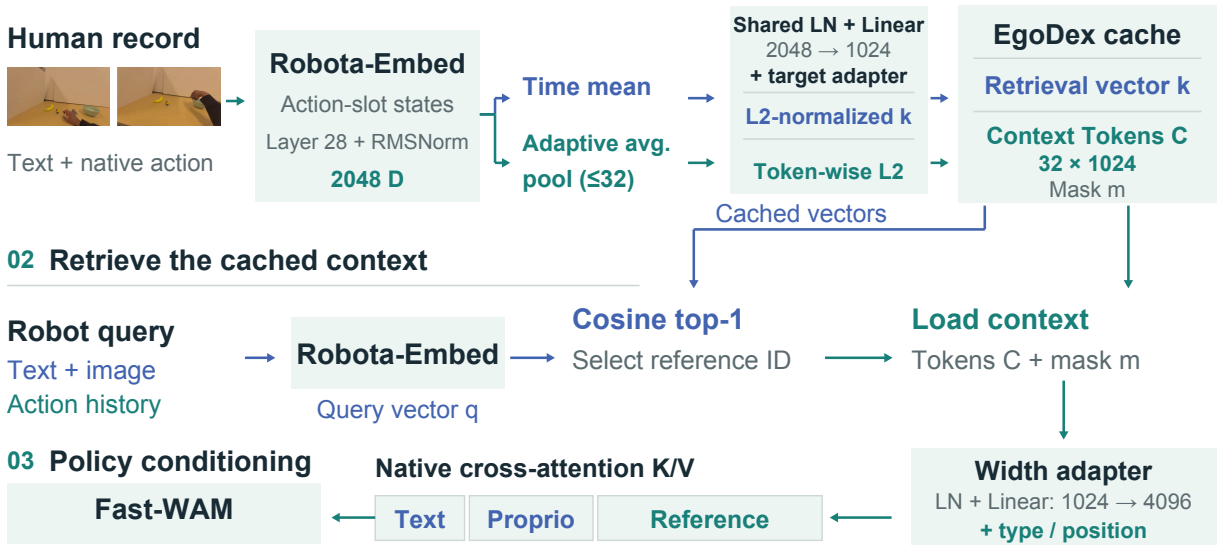}
  \caption{\textbf{Training-based Embed2ICL.}
  Offline encoding caches each candidate's retrieval vector, Context Tokens,
  and context mask. Online cosine top-1 search selects a reference and loads
  its tokens and mask. The prefilling adapter projects the tokens to Fast-WAM's
  context width and appends them to the native text and proprioceptive
  conditioning for cross-attention.}
  \label{fig:fastwam-injection}
  \submissionvspace{-4pt}
\end{figure}

\ifdefined\EXPANDEDREPORT\else\begin{samepage}\fi
\submissionvspace{-3pt}
\paragraph{Offline cache and online lookup.}
Offline, we cache each candidate's retrieval vector, Context Tokens, and
context mask $m_i$. Online, we encode the robot query as $q$, find
$i^*=\arg\max_i q^\top k_i$, and load $(C_{i^*},m_{i^*})$ without re-encoding
the candidate. Each rollout starts without reference context.
After the first successful retrieval, that reference stays active until the
rollout ends. Task-progress metadata is excluded from conditioning.
\par
\ifdefined\EXPANDEDREPORT\else\end{samepage}\fi

\ifdefined\EXPANDEDREPORT\else\begin{samepage}\fi
\submissionvspace{-3pt}
\paragraph{Prefilling.}
Embed2ICL projects $C=C_{i^*}$ to Fast-WAM's conditioning width:
\begin{equation}
    R
    = \operatorname{Linear}_{1024\rightarrow4096}
      \!\left(\operatorname{LN}(C)\right)
      + E_{\mathrm{type}} + E_{\mathrm{pos}},
    \quad R\in\mathbb{R}^{32\times4096}.
    \label{eq:fastwam-prefill-adapter}
\end{equation}
The projected sequence extends native conditioning
$[H_{\mathrm{text}};h_{\mathrm{proprio}}]$; $E_{\mathrm{type}}$ and
$E_{\mathrm{pos}}$ are learned type and position embeddings. Native cross-attention reads it
in all 30 video-expert and 30 action-expert layers, at every
action-diffusion denoising step, and during video KV-cache prefilling.
\par
\ifdefined\EXPANDEDREPORT\else\end{samepage}\fi

\ifdefined\EXPANDEDREPORT\else\begin{samepage}\fi
\paragraph{Policy-adaptation scope.}
Video and action flow-matching losses update approximately 5.00B video-expert
parameters, 1.02B action-expert parameters, the proprioception encoder, and a
4.34M-parameter projection adapter: approximately 6.02B trainable parameters
in total. The text encoder and VAE remain frozen. The context adapter inserts
cached tokens into native conditioning, with no gradient propagation to the
offline \embeddingmodelname{} encoder.
\par
\ifdefined\EXPANDEDREPORT\else\end{samepage}\fi

\paragraph{LIBERO training schedule.}
The retrieved-context condition starts from the official Fast-WAM LIBERO
checkpoint and trains on 1,712 trajectories comprising 277,713 samples
across the four LIBERO suites. Five epochs yield 21,700 parameter updates
with an effective batch size of 64. We use AdamW with a learning rate of
$10^{-4}$, a 5\% warm-up, and cosine decay to $10^{-6}$.
The random-reference condition evaluates this same policy checkpoint.
A separate no-ICL control uses the same 21,700 training updates
(\detailref{sec:fastwam-libero-details}).

\paragraph{Real-robot evaluation.}
Pick Fruit and Split Cup each use 12 rollouts per task and
condition. Following T-Rex's
partial-credit approach~\citep{niu2026trex}, we report
$100\,\frac{1}{12}\sum_{i=1}^{12}s_i$, where $s_i\in[0,1]$ is a rollout's
normalized task-progress score.

\paragraph{Additive scoring.}
Pick Fruit and Split Cup award 0.25 independently for each completed sub-step,
up to a total of 1. For \emph{Pick Fruit}, the four sub-steps are: the left
hand grasps a fruit from the left side; the left hand places that fruit in
the center basket; the right hand grasps a fruit from the right side; and
the right hand places that fruit in the center basket.
For \emph{Split Cup}, they are: the left hand grasps and stabilizes the cup
stack; the right hand separates exactly one cup; the right hand grasps the
separated cup; and the separated cup is placed upright on the table.

Table~\ref{tab:physis-icl-real-robot-results} reports mean task progress and
gains computed before rounding. Figure~\ref{fig:fastwam-real-robot-storyboard}
pairs retrieved references with illustrative executions for these two
tasks and DreamZero/YAM.
\fi
\subsection{LIBERO Supplementary Controls}
\label{app:fastwam-libero}
\label{sec:fastwam-libero-details}

Table~\ref{tab:fastwam-libero-context-controls} compares retrieved and random
Context Tokens with the official release without ICL.
The retrieved- and random-reference conditions evaluate the same policy
checkpoint, changing only the source of Context Tokens at inference.
For the random-context condition, a fixed seed deterministically selects a
pseudorandom real reference window from the EgoDex gallery. Its precomputed
$32\times1024$ Context Tokens replace the retrieved window's tokens;
the control neither injects noise nor shuffles token order.
Each condition uses 50 binary-success trials per task across the five-task
LIBERO difficult subset. Retrieved tokens reach 92.8\% success, compared with
88.0\% for tokens from random references. This isolates reference
selection at fixed policy weights.

Separately, continuing the official release without ICL for 21,700 updates,
matching Embed2ICL's update count, yields 216/250 successes (86.4\%).
Success remains unchanged from the official release and 6.4 percentage points
below Embed2ICL. Thus, continuation for the same number of updates does not
reproduce the gain from retrieved-context conditioning.
\ifdefined\EXPANDEDREPORT
\else
An earlier 2,000-step partial report also shows no gain on its two tasks:
S4 falls from 46/50 to 45/50 and G9 from 45/50 to 43/50; a five-task aggregate
was not reported for that evaluation.
\fi
\paragraph{Retrieved human context.}
Figure~\ref{fig:fastwam-libero-analysis} pairs the G3 bowl-to-drawer task with
a top-1 EgoDex clip of a human moving an object toward a bowl. Both depict
placement into a receptacle. The selected Embed2ICL rollout completes G3,
whereas the rollout from the official release does not.
\begin{figure}[!htb]
  \submissionvspace{-12pt}
  \centering
  \includegraphics[width=0.90\linewidth]{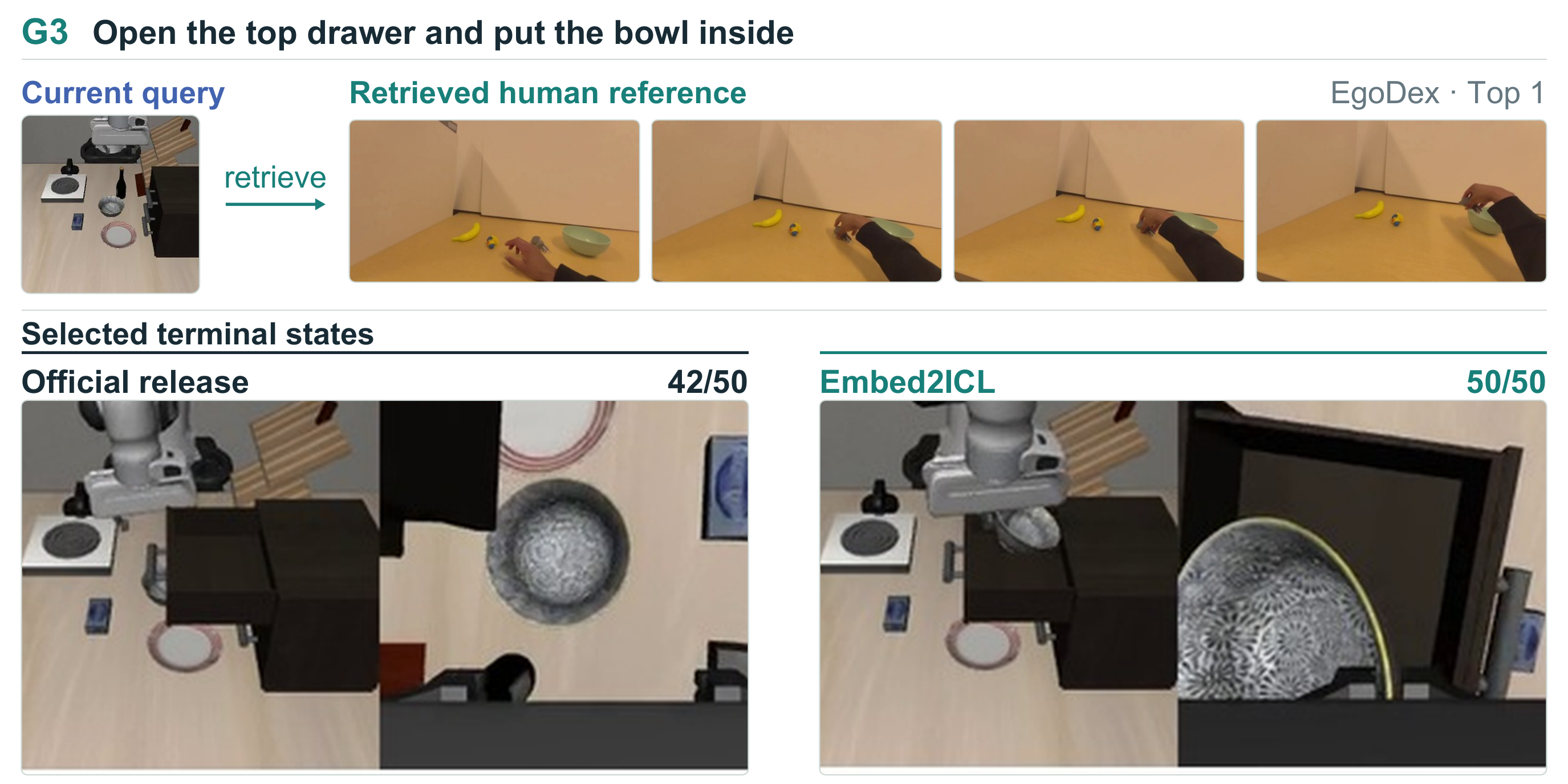}
  \par\vspace{-7pt}
  \ifdefined\EXPANDEDREPORT\captionsetup{skip=13pt}\fi
  \caption{\textbf{LIBERO G3: reference and final states.}
  Successes out of 50 per condition.}
  \label{fig:fastwam-libero-analysis}
  \submissionvspace{-10pt}
\end{figure}
\ifdefined\EXPANDEDREPORT
\else
  \FloatBarrier
\fi
\subsection{Training-Based Integration Ablation}
\label{sec:fastwam-integration-ablation}

We compare two complete integration settings: Embed2ICL's native-context
prefilling with policy fine-tuning, and a frozen-backbone reference side branch.
Both use the \embeddingmodelname{} retriever, EgoDex index, cosine top-1
selection, and complete $32\times1024$ Context Tokens. They differ in both
the conditioning path and trainable scope.

\paragraph{Side-branch setting.}
A shared reference encoder projects the cached tokens to width 512. Each of
the 30 video-expert and 30 action-expert layers receives an eight-head
reference cross-attention block, inserted after self-attention and before
native text/proprioception cross-attention. Channel-wise residual gates are
initialized to zero, preserving the original policy at initialization.
Training updates the reference encoder, added attention blocks, and gates
(approximately 158.74M parameters), while freezing the original Fast-WAM
backbone and proprioception encoder. Standard Embed2ICL instead fine-tunes
approximately 6.02B parameters with its prefilling interface
(\detailref{sec:fastwam-injection-details}).

\ifdefined\EXPANDEDREPORT
\Needspace{20\baselineskip}
\begin{table}[H]
\else
\begin{table}[!ht]
\fi
  \submissionvspace{-3pt}
  \centering
  \caption{\textbf{LIBERO integration ablation.} Successes from 50 trials per
  task and condition; the final row gives the five-task average (\%). Parentheses
  show percentage-point changes from the official release; bold marks the
  better ICL setting.}
  \label{tab:fastwam-integration-libero}
  \begingroup
\small
\setlength{\tabcolsep}{4pt}
\renewcommand{\arraystretch}{1.10}
\begin{tabular*}{\linewidth}{@{\extracolsep{\fill}}lrrr@{}}
  \toprule
  \tablegroup{Suite / ID} & \shortstack{\tablegroup{Official}\\\tablegroup{release}}
    & \tablegroup{Side branch} & \tablegroup{Embed2ICL} \\
  \midrule
  Spatial / S4 & 46 & 46 ($0$) & \tablebest{49} ($+6$) \\
  Goal / G3 & 42 & 44 ($+4$) & \tablebest{50} ($+16$) \\
  Goal / G9 & 45 & 44 ($-2$) & \tablebest{47} ($+4$) \\
  LIBERO-10 / L4 & 43 & 43 ($0$) & \tablebest{44} ($+2$) \\
  LIBERO-10 / L6 & 40 & \tablebest{43} ($+6$) & 42 ($+4$) \\
  \midrule
  \tablegroup{Five-task average (\%)} & 86.4 & 88.0 ($+1.6$)
    & \tablebest{92.8} ($+6.4$) \\
  \bottomrule
\end{tabular*}
\endgroup

  \submissionvspace{-5pt}
\end{table}

\begin{samepage}
\ifdefined\EXPANDEDREPORT
\paragraph{Conclusion: LIBERO.}
\else
\paragraph{LIBERO.}
\fi
The side branch reaches 88.0\% average success, compared with 86.4\% for the
official release and 92.8\% for Embed2ICL
(Table~\ref{tab:fastwam-integration-libero}). Embed2ICL performs better on four
tasks; the side branch performs better on L6 (43 versus 42 successes).
The side branch uses fewer trainable parameters, but Embed2ICL reaches
higher average success.
\par
\end{samepage}

\ifdefined\EXPANDEDREPORT\Needspace{14\baselineskip}\fi
\begin{table}[H]
  \submissionvspace{-3pt}
  \centering
  \caption{\textbf{Real-robot integration ablation.} Mean normalized task
  progress (\%) from 12 rollouts per task and condition. Parentheses show
  percentage-point changes from Fast-WAM without ICL.}
  \label{tab:fastwam-integration-progress}
  \begingroup
\small
\setlength{\tabcolsep}{4pt}
\renewcommand{\arraystretch}{1.10}
\begin{tabular*}{\linewidth}{@{\extracolsep{\fill}}lrrr@{}}
  \toprule
  \tablegroup{Task} & \tablegroup{No ICL}
    & \tablegroup{Side branch} & \tablegroup{Embed2ICL} \\
  \midrule
  Pick Fruit & 75.00 & 77.08 ($+2.08$) & \tablebest{79.17} ($+4.17$) \\
  Split Cup & 66.67 & 75.00 ($+8.33$) & \tablebest{87.50} ($+20.83$) \\
  \bottomrule
\end{tabular*}
\endgroup

  \submissionvspace{-5pt}
\end{table}

\ifdefined\EXPANDEDREPORT
\paragraph{Conclusion: Real robots.}
\else
\paragraph{Real robots.}
\fi
Both settings improve mean normalized task progress over no ICL
(Table~\ref{tab:fastwam-integration-progress}). \mbox{Embed2ICL} reaches 79.17\% on
Pick Fruit and 87.50\% on Split Cup, compared with 77.08\% and 75.00\% for the
side branch. The larger gap occurs on Split Cup.

\ifdefined\EXPANDEDREPORT\Needspace{14\baselineskip}\fi
\begin{table}[H]
  \submissionvspace{-3pt}
  \centering
  \caption{\textbf{Retrieved versus random Context Tokens.}
  Each integration is evaluated on the LIBERO difficult subset with 50 trials
  per task and condition. Results are five-task average success rates (\%).}
  \label{tab:fastwam-integration-context}
  \begingroup
\small
\setlength{\tabcolsep}{4pt}
\renewcommand{\arraystretch}{1.10}
\begin{tabular*}{0.80\linewidth}{@{\extracolsep{\fill}}lrr@{}}
  \toprule
  \tablegroup{Integration} & \tablegroup{Random tokens} & \tablegroup{Retrieved tokens} \\
  \midrule
  Side branch & 85.6 & \tablebest{88.0} \\
  Embed2ICL & 88.0 & \tablebest{92.8} \\
  \bottomrule
\end{tabular*}
\endgroup

  \submissionvspace{-5pt}
\end{table}

\ifdefined\EXPANDEDREPORT
\paragraph{Conclusion: Retrieved versus random context.}
\else
\paragraph{Retrieved versus random context.}
\fi
Retrieved tokens yield higher success than random tokens within each setting
(Table~\ref{tab:fastwam-integration-context}): 88.0\% versus 85.6\% for the
side branch and 92.8\% versus 88.0\% for \mbox{Embed2ICL}. In the \mbox{Embed2ICL} comparison,
both reference conditions use the same policy checkpoint. Retrieved context
helps in both the frozen-backbone and fine-tuned settings.
\FloatBarrier
\ifdefined\TECHREPORTAPPENDIX
  \section{Embed2Evolve: Harnesses, Development, and Evaluation}
\else\ifdefined\EXPANDEDREPORT
  \subsection{Embed2Evolve: Harnesses, Development, and Evaluation Details}
\else
  \section{Embed2Evolve: Harnesses, Development, and Evaluation}
\fi\fi
\label{app:physis-evolve}

Embed2Evolve accumulates completed attempts in task-local memory and searches
for a recipe for using that memory during control. Online retrieval ranks
individual memory anchors; development compares entire context configurations.
The two decisions use different criteria. The policy and embedding models
remain fixed throughout.

\subsection{Policy and Memory Interfaces}
\label{sec:evolve-interface}

\paragraph{Matched embedding-model design.}
We compare \embeddingmodelname{} with Omni-Embed-Nemotron-3B~\citep{xu2025omni}
by replacing only the retrieval embedding model and developing each condition
independently. Both use the same GPT-6 Astra policy, prompt, initial memory,
development and test seeds, candidate-generation rules, and stopping criteria.
Retrieved memories supply the same RGB, action, and feedback fields to the
policy. Both evolved systems use the full memory interface; the random-history
baseline supplies only RGB.

At each decision, GPT-6 Astra receives all current RGB views, robot state,
execution feedback, recent action/feedback history, and external context when
available. It proposes end-effector target poses and gripper commands,
which are validated and executed by a fixed IK/controller.
Table~\ref{tab:physis-evolve-runtime} gives the runtime and budgets.
The four single-arm tasks use \texttt{base\_camera}, \texttt{vision\_side},
\texttt{vision\_top}, and \texttt{vision\_wrist}; TwoRobotStackCube uses base,
side, top, and separate left/right wrist cameras. The policy observes all
four or five views, respectively, at every decision.

\begin{table}[!ht]
  \centering
  \caption{\textbf{Embed2Evolve runtime and observation interface.}
  Simulation time is distinct from client execution time, which includes
  API waiting.}
  \label{tab:physis-evolve-runtime}
\begingroup
\small
\setlength{\tabcolsep}{4pt}
\renewcommand{\arraystretch}{1.12}
\begin{tabular}{@{}p{0.24\linewidth}p{0.71\linewidth}@{}}
\toprule
\tablegroup{Component} & \tablegroup{Configuration} \\
\midrule
Policy & GPT-6 Astra, reasoning effort \texttt{low} \\
Simulator & ManiSkill 3.0.1; SAPIEN 3.0.3 \\
Current RGB & One $256\times256$ image per camera per decision \\
Cameras & Four for single-arm tasks; five for TwoRobotStackCube \\
Control rate & 20\,Hz; two ticks per primitive (0.1\,s simulation time) \\
Decision budget & At most eight primitives per GPT decision \\
Rollout budget & 100 decisions, 300 primitives, and 600 policy control ticks \\
Target-change limits & Per tick: 0.026\,m translation and 0.1\,rad rotation \\
\bottomrule
\end{tabular}
\endgroup

\end{table}

\paragraph{Action execution and observation constraints.}
Each primitive specifies one arm's world-frame end-effector position in metres,
unit XYZW quaternion, and gripper open/close command. Targets are transformed
to that robot's base frame and executed by the native IK/PD controller.
In the two-arm task, the other arm holds its measured pose at the start of the
primitive. The shared prompt requires explicit targets, without hidden tool calls
or automatic execution until a target is reached. It restricts observations
to RGB, the robot's own state, the instruction, and public actions and feedback;
object poses, private simulator state, and task code are excluded. A gripper
command does not establish grasp success, nor does an IK target establish
arrival. The \texttt{joint\_eff} field estimates the robot's own PD drive effort,
not contact ground truth or measured torque. Native environment feedback
determines task success.

\paragraph{Historical-context prompt.}
Both conditions retain the Direct system prompt except that
``This condition is Direct: no memory or retrieval.'' is replaced by
``A historical RGB video excerpt may accompany the current observations.''
Both use this fixed preamble for historical video:
\begin{quote}
This reference is a uniformly random four-second excerpt from a retrieved
historical video. The images are historical, not current observations.
\end{quote}
Here, ``uniformly random'' is inherited prompt wording: both conditions use
the clip associated with the retrieved anchor, not a randomly drawn excerpt
from the selected video.

\paragraph{Anchor index and query.}
For \embeddingmodelname{}, the query combines the instruction, current main-camera RGB,
and an endpoint-error action proxy derived from the most recent acknowledged
active-arm command, rather than achieved displacement. Applying the LIBERO
action codec to ManiSkill is an out-of-distribution adaptation.

Omni-Embed-Nemotron-3B runs in BF16 with eager attention. Its query uses the
\texttt{query:} prefix, task text, and current RGB, without native actions;
its target uses \texttt{passage:}, task text, and historical video.
Retrieved context supplied to GPT includes the clip's public actions,
execution feedback, relative timestamps, and robot state after command
acknowledgement. These policy-context fields are distinct from the
embedding model's inputs.

Memory indexes completed development episodes, including successes and
failures. Each key represents a current-state anchor linked to at most four
seconds of recorded RGB, commands, robot state, feedback, and timestamps.
Retrieval matches the task and active arm, excludes episodes with the current
seed, and selects anchors from distinct source episodes.

\begin{samepage}
\paragraph{Ranking and refresh.}
For candidate anchor $j$, let $S_j\in\{0,1\}$ be its source episode's native
success label and $L_j$ its total control ticks, with budget 600.
For normalized query and anchor vectors $q,z_j$ from the chosen embedding model,
the score uses success/efficiency bonuses $\alpha,\beta$:
\begin{equation}
  R(q,j)=\operatorname{cos}(q,z_j)+\alpha S_j
    +\beta S_j\left(1-\frac{L_j}{600}\right).
  \label{eq:evolve-ranking}
\end{equation}
\end{samepage}
Candidates are ranked in descending order, with at most one anchor retained
per source episode. The efficiency bonus rewards successful episodes that use
fewer control ticks; tokens, cost, and policy decision counts do not enter the
score. For example, 300 ticks with
$\alpha=0.10,\beta=0.03$ adds 0.115 to cosine similarity. The episode-level
label records the outcome of the whole attempt, so it does not establish that
every local action was correct.
Decision zero uses no retrieved context. Retrieval starts at decision one,
refreshes at the selected cadence and upon an active-arm change, and caches
context between refreshes.

\paragraph{Failed attempts.}
Ordinary task failures and budget-exhausted episodes count as unsuccessful
trials and can enter memory. With $S_j=0$, their anchors receive neither a
bonus nor a penalty and remain retrievable by similarity. Exhausted repairs
of malformed policy output also count as valid failures, but the memory
builder does not admit these episodes. Audited infrastructure interruptions
are excluded from the valid-trial denominator and memory; replacement runs
complete the trial budget. The policy receives failed attempts as execution
records under the fixed prompt; these records are not rewritten as prompt advice.

\ifdefined\EXPANDEDREPORT
  \Needspace{10\baselineskip}
\fi
\subsection{Development and Selected Harnesses}
\label{sec:evolve-harnesses}

\newcommand{\evolveharnessfigure}{%
\ifdefined\TECHREPORTPROMOTEEVOLVE
\else
\begin{figure}[!ht]
  \submissionvspace{-6pt}
  \centering
  \ifdefined\EXPANDEDREPORT
    \includegraphics[width=\linewidth]{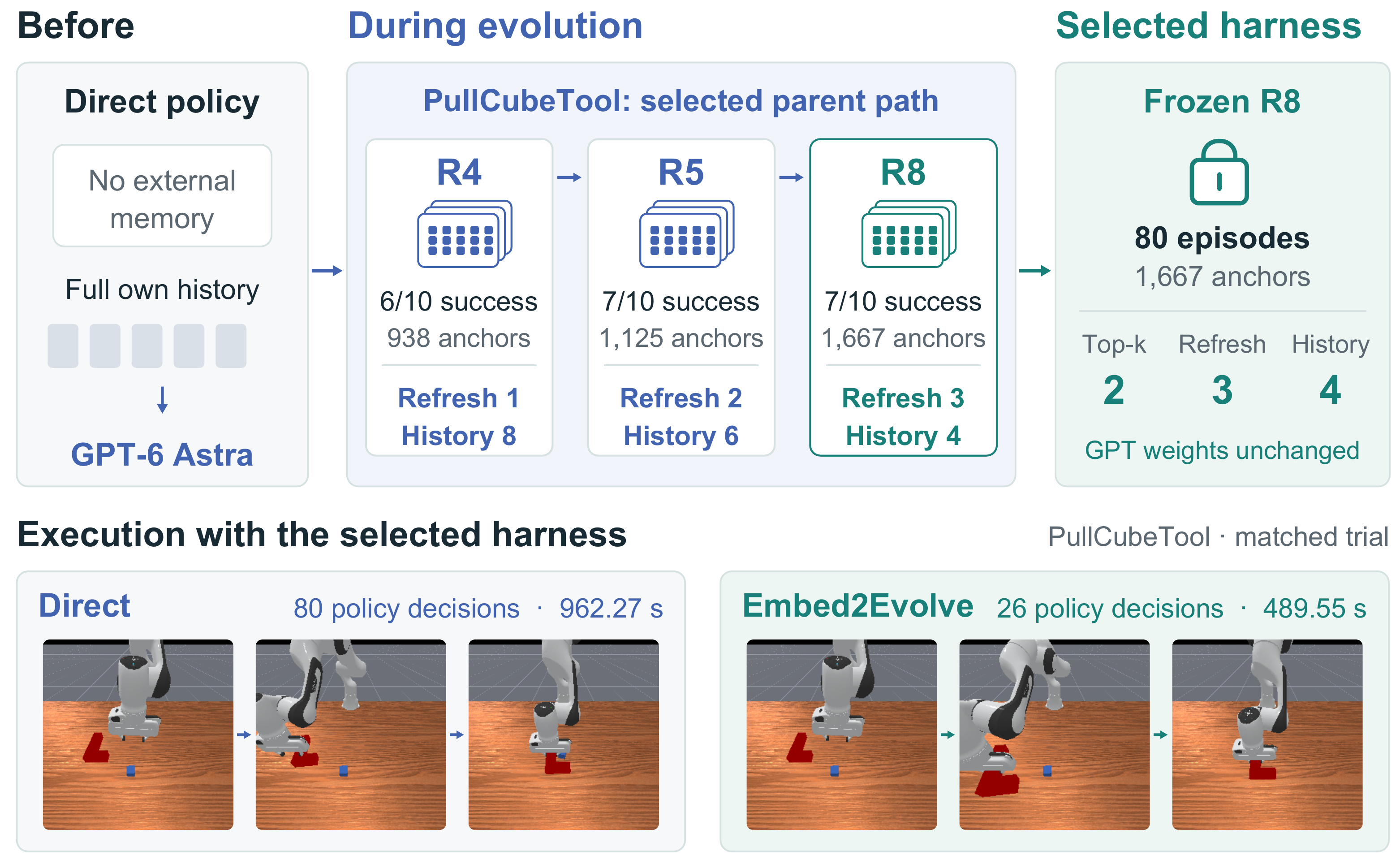}
  \else
    \includegraphics[width=0.87\linewidth]{figures/physis-evolve-harness-evolution.pdf}
  \fi
  \par\vspace{-3pt}
  \caption{\textbf{From Direct control to a selected harness.} PullCubeTool's
  R4 $\rightarrow$ R5 $\rightarrow$ R8 parent--child sequence grows memory
  while reducing retained history and retrieval frequency. The bottom
  filmstrips show the same fresh-test initialization completed successfully
  by both systems. Task execution time includes API waits.}
  \label{fig:physis-evolve-harness}
  \submissionvspace{-4pt}
\end{figure}
\fi
}

\newcommand{\evolvememorygrowthparagraph}{%
Memory expands between rounds, so development changes both available
experience and its retrieval recipe. Search round $r$ need not equal
the numbered report round $\ell$: earlier transport-version rounds can
contribute memory but are excluded from current-version configuration selection.%
}

Search varies retrieval count $k$, refresh cadence $\delta$, history pairs
$H$, and success/efficiency bonuses $\alpha,\beta$. Models, prompt prose,
controller, and episode budgets stay fixed. Every candidate uses development
seeds 0--9 and retains historical images. The history window keeps the most
recent $2H$ messages, corresponding to $H$ action/feedback pairs;
$\delta$ counts policy decisions. Each task begins with the three presets
in Table~\ref{tab:physis-evolve-presets}.

\begin{table}[!ht]
  \centering
  \caption{\textbf{Initial context configurations.} All presets enable
  historical images and are evaluated on the same ten development seeds.}
  \label{tab:physis-evolve-presets}
\begingroup
\small
\setlength{\tabcolsep}{5pt}
\renewcommand{\arraystretch}{1.12}
\begin{tabular*}{\linewidth}{@{\extracolsep{\fill}}lrrrrr@{}}
\toprule
\tablegroup{Preset} & \tablegroup{$k$} & \tablegroup{Cadence $\delta$}
& \tablegroup{History $H$} & \tablegroup{$\alpha$} & \tablegroup{$\beta$} \\
\midrule
A & 2 & 2 & 12 & 0.10 & 0.03 \\
B & 1 & 4 & 6 & 0.10 & 0.10 \\
C & 2 & 1 & 8 & 0.15 & 0.05 \\
\bottomrule
\end{tabular*}
\endgroup

\end{table}
\ifdefined\EXPANDEDREPORT
  \par\evolvememorygrowthparagraph\par
  \evolveharnessfigure
\fi

\paragraph{Success-first selection.}
Let $\mathcal E$ contain the eligible numbered development rounds. For round
$\ell$, write $N_\ell$ for its number of successful trials, and
$\overline L_\ell,\overline T_\ell$ for mean control ticks and client execution
time over its successful trials. We select lexicographically:
\begin{equation}
  \ell^\star=\mathop{\operatorname{arg\,min}}_{\ell\in\mathcal E}
  \left(-N_\ell,\overline L_\ell,\overline T_\ell,\ell\right).
  \label{eq:evolve-selection}
\end{equation}
We first maximize successes, then break ties by fewer control ticks, shorter
time, and the earlier round. Every round contains ten valid trials. If none
succeeds, both successful-trial means are $+\infty$, so rapid failure confers
no efficiency advantage. Configuration selection thus ranks success and
efficiency in order of priority; online retrieval combines them in a weighted score.

\begin{samepage}
\paragraph{Local proposals.}
After the presets, search round $r$ copies the current winner and makes one
deterministic local proposal; all unlisted settings are inherited:
\begin{itemize}
  \item If $r\bmod 3=1$, set $\delta\leftarrow\min(4,\delta+1)$ and
  $H\leftarrow6$ if $H>6$, or $H\leftarrow4$ otherwise.
  \item If $r\bmod 3=2$, set $\delta\leftarrow\max(1,\delta-1)$,
  $k\leftarrow2$, and $\alpha\leftarrow\min(0.25,\alpha+0.05)$.
  \item If $r\bmod 3=0$, set $k\leftarrow1$ and
  $\beta\leftarrow\min(0.20,\beta+0.05)$.
\end{itemize}
\end{samepage}
\ifdefined\EXPANDEDREPORT\else
\evolvememorygrowthparagraph
\fi

\begin{samepage}
\paragraph{Stopping and freezing.}
Stopping tracks the Pareto frontier of
$(N_\ell,-\overline L_\ell,-\overline T_\ell)$. A point dominates another
when it is no worse in all three coordinates and better in at least one.
A candidate adds a frontier point only if no existing point dominates it
and its coordinates are not already present.
\par
\end{samepage}
After at least three eligible
rounds under the current transport version, two consecutive rounds without
a new frontier point trigger stopping. A separate cap allows at most eight
numbered development rounds per task, including earlier transport versions;
baseline collection and formal testing do not count toward it. Any started
round must complete ten valid trials.

For each embedding model, formal testing freezes the winning configuration
and the memory bank from that round. Test outcomes do not enter selection.
The two conditions develop independently and can produce different banks and
recipes. The comparison measures the embedding model's contribution through
evolution, not a replacement within one final harness.

With \embeddingmodelname{}, StackCube stops by the frontier rule after four rounds and
TwoRobotStackCube after eight. PokeCube, LiftPegUpright, and PullCubeTool each
reach the eight-round cap. The 36 recorded configurations and 360 trials
include historical first rounds that are ineligible for selection.
Table~\ref{tab:physis-evolve-harnesses} lists the selected harnesses, and the
trajectories below show eligible candidates.

\begin{table}[!ht]
  \centering
  \caption{\textbf{Selected Embed2Evolve harnesses.} Bank sizes count
  development episodes and indexed anchors. Cadence counts GPT decisions;
  $H$ counts retained action/feedback pairs.}
  \label{tab:physis-evolve-harnesses}
  \begingroup
\small
\setlength{\tabcolsep}{2.5pt}
\renewcommand{\arraystretch}{1.12}
\begin{tabular*}{\linewidth}{@{\extracolsep{\fill}}lrrrrrrr@{}}
\toprule
\tablegroup{Task} & \tablegroup{Round} & \tablegroup{Episodes} & \tablegroup{Anchors}
& \tablegroup{$k$} & \tablegroup{Cadence} & \tablegroup{$H$} & \tablegroup{$\alpha/\beta$} \\
\midrule
Poke Cube & 8 & 80 & 1,610 & 2 & 2 & 6 & .15/.05 \\
Stack Cube & 2 & 20 & 454 & 2 & 2 & 12 & .10/.03 \\
Lift Peg Upright & 6 & 60 & 1,116 & 2 & 1 & 6 & .20/.05 \\
Pull Cube with Tool & 8 & 80 & 1,667 & 2 & 3 & 4 & .15/.05 \\
Two-Robot Stack Cube & 5 & 50 & 1,126 & 2 & 3 & 6 & .10/.03 \\
\bottomrule
\end{tabular*}
\endgroup

\end{table}

\ifdefined\EXPANDEDREPORT\else
  \evolveharnessfigure
\fi

\FloatBarrier
\ifdefined\EXPANDEDREPORT
  \Needspace{0.58\textheight}
\fi
\subsection{Development Trajectories}
\label{sec:evolve-development-trajectories}

\ifdefined\EXPANDEDREPORT
Figure~\ref{fig:physis-evolve-development} traces \embeddingmodelplain{}'s
candidate performance. Decision counts
average each round's successful trials; test-time efficiency uses matched
initializations (\detailref{sec:evolve-test-protocol}). Hollow diamonds mark selected
harnesses, which need not come from the final round.
\else
Figures~\ref{fig:physis-evolve-development} and~\ref{fig:physis-evolve-recipes}
trace \embeddingmodelplain{}'s candidate performance and context recipes. Decision counts
average each round's successful trials; test-time efficiency uses matched
initializations (\detailref{sec:evolve-test-protocol}). Hollow diamonds mark selected
harnesses, which need not come from the final round.
\fi

\ifdefined\EXPANDEDREPORT
  \begin{figure}[H]
\else
  \begin{figure}[!ht]
\fi
  \centering
  \includegraphics[width=\linewidth]{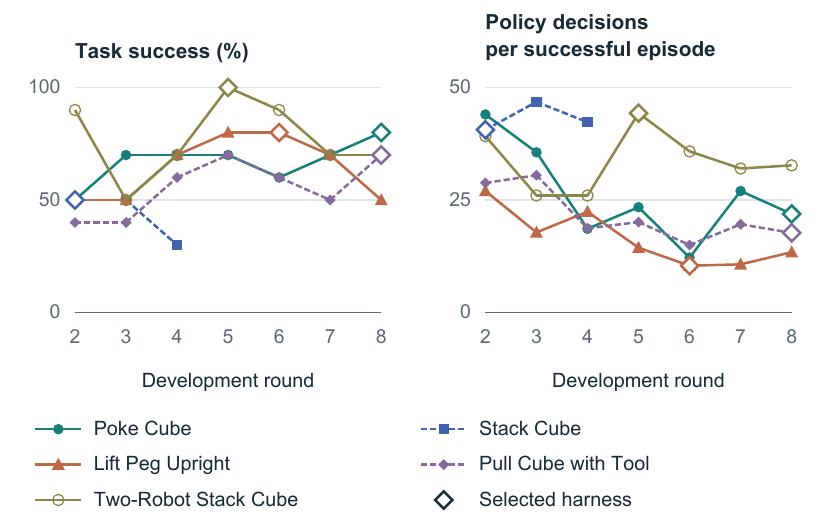}
  \caption{\textbf{Task-wise development trajectories.} Success uses ten
  trials per round; decisions average that round's successful trials.
  Hollow diamonds mark selected harnesses. Lines connect evaluated candidates,
  not necessarily parent--child configurations.}
  \label{fig:physis-evolve-development}
\end{figure}
\ifdefined\EXPANDEDREPORT\else
\ifdefined\EXPANDEDREPORT
  \begin{figure}[!htbp]
\else
  \begin{figure}[!ht]
\fi
  \centering
  \ifdefined\EXPANDEDREPORT
    \relax
  \else
    \relax
  \fi
  \caption{\textbf{Two evolving components: memory and context recipe.}
  Per-task trajectories show memory size, retrieval count, refresh cadence,
  history, and ranking bonuses with fixed model weights. Hollow diamonds
  mark selected rounds; nested diamonds show coincident selections.}
  \label{fig:physis-evolve-recipes}
\end{figure}
\fi

\FloatBarrier
\subsection{Embedding-Model Comparison}
\label{sec:evolve-encoder-control}

Table~\ref{tab:physis-evolve-encoder-development} lists the completed and
selected rounds for each embedding model. Both stay within the eight-round
cap and produce their own memory banks and context configurations.

\ifdefined\EXPANDEDREPORT
  \begin{table}[H]
\else
  \begin{table}[!ht]
\fi
  \centering
  \caption{\textbf{Independent development under matched search rules.}
  Completed rounds count all numbered rounds; selected rounds identify
  the configuration and memory frozen for testing.}
  \label{tab:physis-evolve-encoder-development}
\begingroup
\small
\setlength{\tabcolsep}{4pt}
\renewcommand{\arraystretch}{1.12}
\begin{tabular*}{\linewidth}{@{\extracolsep{\fill}}lrrrr@{}}
\toprule
& \multicolumn{2}{c}{\shortstack{\tablegroup{Omni-Embed-}\\\tablegroup{Nemotron-3B}}}
& \multicolumn{2}{c}{\tablegroup{\embeddingmodelplain{}}} \\
\cmidrule(lr){2-3}\cmidrule(l){4-5}
\tablegroup{Task} & \tablegroup{Completed} & \tablegroup{Selected}
& \tablegroup{Completed} & \tablegroup{Selected} \\
\midrule
Poke Cube & 8 & 6 & 8 & 8 \\
Stack Cube & 6 & 3 & 4 & 2 \\
Lift Peg Upright & 8 & 4 & 8 & 6 \\
Pull Cube with Tool & 3 & 1 & 8 & 8 \\
Two-Robot Stack Cube & 7 & 5 & 8 & 5 \\
\bottomrule
\end{tabular*}
\endgroup

\end{table}

\ifdefined\EXPANDEDREPORT
The selected Omni-Embed-Nemotron-3B releases' frozen banks contain 182 rollout
videos: 49 from the original Direct collection and 133 from that condition's
own development, with 30 successful and 152 failed rollouts. These are the
selected-round snapshots, not a post-hoc merge of all development trajectories.
Video counts differ from indexed-anchor counts; failed rollouts remain
retrievable by similarity under the scoring rule in \detailref{sec:evolve-interface}.

The two context systems evolve independently; the embedding-model replacement
changes both input modalities and retrieval training. It is neither an
action-only ablation nor an encoder swap within
one fixed final harness. Fewer high-level decisions can coexist with more
physical control ticks (Table~\ref{tab:physis-evolve-control-ticks}): these
metrics measure different aspects of execution.
\else
Table~\ref{tab:physis-evolve-omni-frozen-memory} records the rollout videos
in each selected Omni-Embed-Nemotron-3B release's frozen bank. These snapshots
contain 182 videos: 49 from the original Direct collection and
133 from that condition's own development, with 30 successful and 152 failed
rollouts. They are not reconstructed by merging all trajectories after
development ends. Video counts differ from indexed-anchor counts; a failed
rollout remains retrievable by similarity under the scoring
rule in \detailref{sec:evolve-interface}.

\begin{table}[!ht]
  \centering
  \caption{\textbf{Frozen memory banks for Omni-Embed-Nemotron-3B.}
  Counts refer to rollout videos bound to the selected releases. Source
  columns and outcome columns each sum to the video total.}
  \label{tab:physis-evolve-omni-frozen-memory}
  \relax
\end{table}

The harness developed with \embeddingmodelname{} succeeds on 68 of 100 tests,
compared with 33 for Omni-Embed-Nemotron-3B, and improves success on every task
(Table~\ref{tab:physis-evolve-main}). Both systems have access to memory,
but the retrieval representation affects the quality of the resulting harness.
This comparison changes the entire embedding model, including its input modalities
and retrieval training. The benchmark's clip-localization and modality results
suggest an explanation: action-conditioned retrieval may help distinguish
execution stage within the same task.

\paragraph{Context as an action prior.}
Retrieved continuations can provide a soft prior over GPT-6 Astra's next
actions. For tool use, an action--outcome example links tool contact and motion
to their effect on an object; for dual-arm manipulation, it provides a reference
for arm roles and coordination. Stage-relevant context may thus focus the
policy on plausible continuations within the unchanged action interface.
Such guidance offers a possible explanation for higher success with fewer
policy decisions.
Fewer high-level decisions can also coexist with more physical control ticks
(Table~\ref{tab:physis-evolve-control-ticks}); the two measure different aspects
of execution.
\fi

\FloatBarrier
\begin{samepage}
\subsection{Fresh-Test Protocol and Efficiency}
\label{sec:evolve-test-protocol}

The five tasks---PokeCube, StackCube, LiftPegUpright, PullCubeTool, and
TwoRobotStackCube---were selected from earlier development results.
All four conditions use the same 20 test seeds (40--59) per task,
disjoint from development seeds 0--9, yielding 400 valid test episodes.
We measure native task success; budget exhaustion and policy failures
count as failures. Audited infrastructure interruptions are excluded.
\par
\end{samepage}

\paragraph{System comparisons.}
GPT-6 Astra Direct (Direct) uses its original prompt and full within-episode
action/feedback history without external memory. GPT-6 Astra + random history adds one randomly sampled same-task
development clip with up to eight ordered RGB frames. This RGB-only reference
remains fixed within the episode.
The harnesses evolved with Omni-Embed-Nemotron-3B and \embeddingmodelname{}
use their selected memories, bounded histories, ranking, and context refresh.
Table~\ref{tab:physis-evolve-main} compares these four context systems.

Success rates and token-use means include all test trials; decision-count and
execution-time comparisons use shared successes. Table~\ref{tab:physis-evolve-metric-protocols}
specifies each metric's population and aggregation rule.
\begin{table}[!ht]
  \centering
  \caption{\textbf{Embed2Evolve test-metric accounting.} All five tasks
  receive equal weight. Paired comparisons and reference API costs use
  Direct and Embed2Evolve.}
  \label{tab:physis-evolve-metric-protocols}
\begingroup
\small
\setlength{\tabcolsep}{4pt}
\renewcommand{\arraystretch}{1.14}
\begin{tabular}{@{}>{\raggedright\arraybackslash}p{0.22\linewidth}
                  >{\raggedright\arraybackslash}p{0.30\linewidth}
                  >{\raggedright\arraybackslash}p{\dimexpr0.48\linewidth-16pt\relax}@{}}
\toprule
\tablegroup{Metric} & \tablegroup{Test population} & \tablegroup{Five-task aggregation} \\
\midrule
Task success & All 20 trials per task in all four conditions & Mean of task success rates. \\
\addlinespace[3pt]
Policy decisions and execution time & Matched initializations successful under Direct and Embed2Evolve & Mean of task-wise relative reductions; speedup averages Direct/Embed2Evolve time ratios. \\
\addlinespace[3pt]
Policy tokens & All 20 trials per task for Direct, evolution with Omni-Embed-Nemotron-3B, and Embed2Evolve, including failures & Relative reduction between equal-task per-rollout means. \\
\addlinespace[3pt]
Reference API cost & All 20 trials per task for Direct and Embed2Evolve, including failures & Relative reduction between equal-task per-rollout means at fixed input/output rates. \\
\bottomrule
\end{tabular}
\endgroup

\end{table}

\begin{samepage}
\paragraph{Efficiency accounting.}
Direct and Embed2Evolve both succeed on 7, 2, 5, 1, and 4 matched
initializations in task order. Here, Evolve denotes Embed2Evolve.
Per task, we compute the reduction
$100(1-\overline{x}_{\rm Evolve}/\overline{x}_{\rm Direct})$ on this paired set.
Averaging the five task-wise reductions equally gives 64.1\% fewer policy
decisions and 41.4\% less task execution time.
\ifdefined\EXPANDEDREPORT
Observed mean speedup is
\else
Mean speedup is
\fi
$\frac{1}{5}\sum_{j=1}^{5}\overline{T}_{{\rm Direct},j}/\overline{T}_{{\rm Evolve},j}
\approx 1.90$, where each mean uses that task's shared successful trials.
The 1.90$\times$ speedup and 41.4\% reduction average task-wise time ratios
and relative savings, respectively.
\par
\end{samepage}

Task execution time includes context preparation, API waiting, and execution overhead,
but excludes embedding-model cold-start time. It varies with API load, whereas
replay videos show simulation time without API waits.
\ifdefined\EXPANDEDREPORT
During this evaluation, rate-limit recovery changed API caller handling
mid-batch, and Direct was added after the batch began. Concurrency, CPU
placement, and API load were not controlled across conditions. These
differences confound wall-time attribution: the reported time ratios describe
observed execution, not an isolated method-induced speedup.
\fi
Table~\ref{tab:physis-evolve-control-ticks} reports physical control ticks
separately from GPT decisions.

\ifdefined\EXPANDEDREPORT
\begin{table}[!ht]
  \centering
  \caption{\textbf{Efficiency and physical control on shared successes.}
  Direct $\rightarrow$ Embed2Evolve means over the same $n$ matched successful
  initializations for all three metrics. Execution time includes API waiting.
  Tick change is $100(\mathrm{Evolve}/\mathrm{Direct}-1)$; positive values
  indicate more physical control ticks despite fewer GPT decisions.}
  \label{tab:physis-evolve-efficiency}
  \label{tab:physis-evolve-control-ticks}
  \begingroup
\small
\setlength{\tabcolsep}{2.5pt}
\renewcommand{\arraystretch}{1.10}
\begin{tabular*}{\linewidth}{@{\extracolsep{\fill}}lrrrrr@{}}
\toprule
\tablegroup{Task} & \shortstack[r]{\tablegroup{Paired}\\\tablegroup{$n$}}
& \shortstack[r]{\tablegroup{Policy}\\\tablegroup{decisions}}
& \shortstack[r]{\tablegroup{Execution}\\\tablegroup{time (s)}}
& \shortstack[r]{\tablegroup{Control}\\\tablegroup{ticks}}
& \shortstack[r]{\tablegroup{Tick change}\\\tablegroup{(\%)}} \\
\midrule
Poke Cube & 7 & 34.86 $\rightarrow$ \tablebest{22.71} & 419.30 $\rightarrow$ \tablebest{382.16} & 133.0 $\rightarrow$ 181.4 & $+36.4$ \\
Stack Cube & 2 & 45.00 $\rightarrow$ \tablebest{10.00} & 499.03 $\rightarrow$ \tablebest{186.35} & 90.0 $\rightarrow$ 65.0 & $-27.8$ \\
Lift Peg Upright & 5 & 57.20 $\rightarrow$ \tablebest{12.40} & 556.06 $\rightarrow$ \tablebest{234.46} & 152.0 $\rightarrow$ 149.6 & $-1.6$ \\
Pull Cube with Tool & 1 & 80.00 $\rightarrow$ \tablebest{26.00} & 962.27 $\rightarrow$ \tablebest{489.55} & 347.0 $\rightarrow$ 337.0 & $-2.9$ \\
Two-Robot Stack Cube & 4 & 87.25 $\rightarrow$ \tablebest{33.00} & 1054.67 $\rightarrow$ \tablebest{751.59} & 206.0 $\rightarrow$ 312.8 & $+51.8$ \\
\bottomrule
\end{tabular*}
\endgroup

\end{table}
\else
\begin{table}[!ht]
  \centering
  \caption{\textbf{Efficiency on shared successes.} Direct $\rightarrow$
  Embed2Evolve means over $n$ matched successful initializations.
  Execution time includes API waiting.}
  \label{tab:physis-evolve-efficiency}
  \relax
  \submissionvspace{-6pt}
\end{table}

\begin{table}[!ht]
  \centering
  \caption{\textbf{Physical control on shared successes.} Mean executed
  ticks on the same paired trials as Table~\ref{tab:physis-evolve-efficiency}.
  Change is $100(\mathrm{Evolve}/\mathrm{Direct}-1)$; positive values mean
  more physical control ticks despite fewer GPT decisions.}
  \label{tab:physis-evolve-control-ticks}
  \relax
\end{table}
\fi

\FloatBarrier
\ifdefined\EXPANDEDREPORT
  \Needspace{15\baselineskip}
\fi
\subsection{Token Usage and Reference Cost}
\label{sec:evolve-token-cost}

Table~\ref{tab:physis-evolve-token-cost} reports mean policy input, output,
and total tokens across all 20 test rollouts per task and condition,
including successes and failures, for GPT-6 Astra Direct and Embed2Evolve.
We average the five task means equally before comparing methods:
total tokens fall from 920,979.76 to 518,216.15 per rollout, a 43.7\%
reduction. Reference API cost falls from \$9.9641 to \$5.7980, or 41.8\%.
Both reductions use $100(1-\overline{x}_{\rm Evolve}/\overline{x}_{\rm Direct})$
on these five-task means. They measure resource use per attempt, including
failures; timing in \detailref{sec:evolve-test-protocol} uses initializations
on which both Direct and Embed2Evolve succeed.

Reference costs use fixed rates of \$10 per million input tokens and \$50
per million output tokens, giving $(10I+50O)/10^6$ USD for input count $I$
and output count $O$. This fixed-rate estimate of test-time policy cost
excludes cache discounts, retrieval, simulation, and memory/harness development.

\ifdefined\EXPANDEDREPORT
  \begin{table}[!htbp]
\else
  \begin{table}[p]
\fi
  \centering
  \ifdefined\EXPANDEDREPORT
    \vspace{6pt}
  \fi
  \caption{\textbf{Policy token usage and reference cost per rollout.}
  Means include all 20 test rollouts per task and condition, regardless of
  outcome. Direct denotes GPT-6 Astra Direct; the final group averages
  task means equally. USD estimates use the fixed input/output rates above.}
  \label{tab:physis-evolve-token-cost}
\begingroup
\small
\setlength{\tabcolsep}{3pt}
\renewcommand{\arraystretch}{1.10}
\begin{tabular*}{\linewidth}{@{\extracolsep{\fill}}lrrrr@{}}
\toprule
\tablegroup{Method} & \tablegroup{Input tokens} & \tablegroup{Output tokens}
& \tablegroup{Total tokens} & \shortstack[r]{\tablegroup{Ref. cost}\\\tablegroup{(USD)}} \\
\midrule
\multicolumn{5}{l}{\tablegroup{Poke Cube}} \\
Direct & 693,733.05 & 16,052.50 & 709,785.55 & 7.7400 \\
Embed2Evolve & 375,759.60 & 12,903.40 & 388,663.00 & 4.4028 \\
\addlinespace[2pt]
\multicolumn{5}{l}{\tablegroup{Stack Cube}} \\
Direct & 935,086.30 & 18,401.05 & 953,487.35 & 10.2709 \\
Embed2Evolve & 656,130.60 & 19,000.10 & 675,130.70 & 7.5113 \\
\addlinespace[2pt]
\multicolumn{5}{l}{\tablegroup{Lift Peg Upright}} \\
Direct & 839,705.25 & 16,225.95 & 855,931.20 & 9.2083 \\
Embed2Evolve & 380,229.95 & 13,087.65 & 393,317.60 & 4.4567 \\
\addlinespace[2pt]
\multicolumn{5}{l}{\tablegroup{Pull Cube with Tool}} \\
Direct & 955,384.05 & 20,302.90 & 975,686.95 & 10.5690 \\
Embed2Evolve & 399,519.65 & 15,041.75 & 414,561.40 & 4.7473 \\
\addlinespace[2pt]
\multicolumn{5}{l}{\tablegroup{Two-Robot Stack Cube}} \\
Direct & 1,086,699.35 & 23,308.40 & 1,110,007.75 & 12.0324 \\
Embed2Evolve & 702,461.85 & 16,946.20 & 719,408.05 & 7.8719 \\
\midrule
\multicolumn{5}{l}{\tablegroup{Five-task average}} \\
Direct & 902,121.60 & 18,858.16 & 920,979.76 & 9.9641 \\
Embed2Evolve & 502,820.33 & 15,395.82 & 518,216.15 & 5.7980 \\
\bottomrule
\end{tabular*}
\endgroup

\end{table}
\ifdefined\EXPANDEDREPORT
  \FloatBarrier
\fi

Total token use decreases on every task, most on Pull Cube with Tool (57.5\%).
Input tokens account for most of the savings. Even on Stack Cube, where
output usage increases, the reduction in input tokens lowers total consumption.

\ifdefined\EXPANDEDREPORT
  \Needspace{5\baselineskip}
\paragraph{Token use after matched context evolution.}
The Omni-Embed-Nemotron-3B comparison in Table~\ref{tab:physis-evolve-main}
uses all final test rollouts of the independently evolved, frozen harnesses,
including failures, and excludes development consumption. Its aggregate
total-token comparison does not establish a wall-time or monetary-cost ratio:
neither an Omni-Embed-Nemotron-3B input/output split nor matched-success timing
is available.
\else
\paragraph{Token use after matched context evolution.}
The harness independently evolved with Omni-Embed-Nemotron-3B averages
approximately 2,531,000 policy tokens per test rollout, including both
successes and failures.
Embed2Evolve averages 518,216.15 under the same test protocol: approximately
79.5\% fewer tokens alongside 35 percentage points higher success.
These means describe the selected harnesses at test time and exclude development
consumption. The timing results above compare Direct and Embed2Evolve on their
shared successful trials.
\fi
\FloatBarrier
\ifdefined\TECHREPORTAPPENDIX
  \Needspace{0.48\textheight}
  \section{Related Retrieval Protocols}
\else\ifdefined\EXPANDEDREPORT
  \subsection{Related Retrieval Protocols}
\else
  \clearpage
  \section{Related Retrieval Protocols}
\fi\fi
\label{app:related-protocols}

Table~\ref{tab:related-protocols} compares retrieval inputs, the role of action,
gallery scope, and policy dependence. Action can enter the retrieval key,
a retrieved record, or the consuming policy's output.
\emph{Policy coupling} specifies how the retrieval interface depends on that
policy; \emph{lifetime} specifies the scope over which retrieved experience is
stored or reused. In Zeva, both phase keys encode recurrent interaction history,
including executed actions and observed effects. RADAR's row follows its
observation--language retrieval formulation. Embed2ICL and Embed2Evolve share
\embeddingmodelname{}'s retrieval space.
\begin{table}[!ht]
  \submissionvspace{-5pt}
  \centering
  \caption{\tablegroup{Representative physical-experience retrieval interfaces.}
  V/L/A/S/P denote vision, language, native action, state, and proprioception;
  3D denotes point clouds. Training-only action use is listed under action
  role; cross-source galleries span datasets or interfaces.}
  \label{tab:related-protocols}
  \ifdefined\TECHREPORTAPPENDIX
    \footnotesize
    \setlength{\tabcolsep}{2pt}
    \renewcommand{\arraystretch}{1.18}
    \begin{tabularx}{\linewidth}{@{}
      >{\raggedright\arraybackslash}p{0.16\linewidth}
      >{\raggedright\arraybackslash}p{0.11\linewidth}
      >{\raggedright\arraybackslash}X
      >{\raggedright\arraybackslash}p{0.14\linewidth}
      >{\centering\arraybackslash}p{0.07\linewidth}
      >{\raggedright\arraybackslash}p{0.12\linewidth}
      >{\raggedright\arraybackslash}p{0.12\linewidth}@{}}
  \else
    \scriptsize
    \setlength{\tabcolsep}{1.7pt}
    \renewcommand{\arraystretch}{1.12}
    \begin{tabular*}{\linewidth}{@{\extracolsep{\fill}}lllllll@{}}
  \fi
    \toprule
    \tablegroup{Method} & \tablegroup{Modalities} & \tablegroup{Query $\rightarrow$ candidate} & \tablegroup{Action role} & \tablegroup{Cross-source} & \tablegroup{Policy coupling} & \tablegroup{Lifetime} \\
    \midrule
    Deep Multimodal & 3D/L/A & 3D+L $\rightarrow$ trajectory & Candidate key & No & Independent & Offline corpus \\
    CLASP & L/S/A & L $\rightarrow$ state+action & Candidate key & No & Independent & Offline corpus \\
    VLA-R & V/L/A & scene V/L $\rightarrow$ action token & Candidate key & No & Direct control & Action vocab. \\
    \midrule
    ROSER & P & prototype $\rightarrow$ window & None & No & Independent & Offline corpus \\
    RAEA & V/L & V/L $\rightarrow$ V/L record & Returned only & Yes & Coupled & Persistent bank \\
    REGENT & S & state $\rightarrow$ demo state & Returned+blend & No & Coupled & Task buffer \\
    RICL & V & image $\rightarrow$ demo frame & Returned+blend & No & Coupled & Task buffer \\
    Zeva & V/A & task phase $\rightarrow$ stored phase & Recurrent keys & No & Coupled & Within episode \\
    RADAR & V/L & eval V/L $\rightarrow$ train V/L & Policy prior & No & Analysis only & Offline corpus \\
    \midrule
    \tablegroup{\embeddingmodelname{}} & \tablegroup{V/L/A} & \tablegroup{current VLA $\rightarrow$ future VLA} & \tablegroup{Query + candidate} & \tablegroup{Yes} & \tablegroup{Independent} & \tablegroup{Persistent index} \\
    \bottomrule
  \ifdefined\TECHREPORTAPPENDIX
    \end{tabularx}
  \else
    \end{tabular*}
  \fi
  \submissionvspace{-6pt}
\end{table}

\FloatBarrier

\end{document}